\pdfoutput=1
\documentclass[11pt]{article}

\usepackage[preprint]{acl}
\usepackage{booktabs}
\usepackage{float}
\usepackage{placeins}
\usepackage{longtable}
\usepackage{pdflscape}
\usepackage{xltabular}
\usepackage[normalem]{ulem}

\newcommand{\conferencetablestyle}{%
  \setlength{\abovecaptionskip}{4pt}%
  \setlength{\belowcaptionskip}{0pt}%
  \renewcommand{\arraystretch}{1.08}%
}
\usepackage{times}
\usepackage{latexsym}

\usepackage{multirow}
\usepackage{makecell}

\usepackage{amsmath}

\usepackage{colortbl}

\definecolor{bestshade}{RGB}{217,239,217}
\definecolor{worstshade}{RGB}{255,224,178}

\makeatletter
\newcommand{\HU}[1]{%
  \@tempdima=#1pt\relax
  \ifdim\@tempdima<0.45pt\cellcolor{orange!55}#1%
  \else\ifdim\@tempdima<0.55pt\cellcolor{orange!35}#1%
  \else\ifdim\@tempdima<0.65pt\cellcolor{orange!20}#1%
  \else\ifdim\@tempdima<0.75pt\cellcolor{yellow!30}#1%
  \else\ifdim\@tempdima<0.85pt\cellcolor{yellow!15}#1%
  \else#1\fi\fi\fi\fi\fi}

\newcommand{\HD}[1]{%
  \@tempdima=#1pt\relax
  \ifdim\@tempdima>0.40pt\cellcolor{orange!55}#1%
  \else\ifdim\@tempdima>0.25pt\cellcolor{orange!35}#1%
  \else\ifdim\@tempdima>0.15pt\cellcolor{orange!20}#1%
  \else\ifdim\@tempdima>0.08pt\cellcolor{yellow!30}#1%
  \else\ifdim\@tempdima>0.03pt\cellcolor{yellow!15}#1%
  \else#1\fi\fi\fi\fi\fi}

\newcommand{\HUb}[1]{%
  \@tempdima=#1pt\relax
  \ifdim\@tempdima<0.45pt\cellcolor{orange!55}\textbf{#1}%
  \else\ifdim\@tempdima<0.55pt\cellcolor{orange!35}\textbf{#1}%
  \else\ifdim\@tempdima<0.65pt\cellcolor{orange!20}\textbf{#1}%
  \else\ifdim\@tempdima<0.75pt\cellcolor{yellow!30}\textbf{#1}%
  \else\ifdim\@tempdima<0.85pt\cellcolor{yellow!15}\textbf{#1}%
  \else\textbf{#1}\fi\fi\fi\fi\fi}

\newcommand{\HDb}[1]{%
  \@tempdima=#1pt\relax
  \ifdim\@tempdima>0.40pt\cellcolor{orange!55}\textbf{#1}%
  \else\ifdim\@tempdima>0.25pt\cellcolor{orange!35}\textbf{#1}%
  \else\ifdim\@tempdima>0.15pt\cellcolor{orange!20}\textbf{#1}%
  \else\ifdim\@tempdima>0.08pt\cellcolor{yellow!30}\textbf{#1}%
  \else\ifdim\@tempdima>0.03pt\cellcolor{yellow!15}\textbf{#1}%
  \else\textbf{#1}\fi\fi\fi\fi\fi}
\makeatother

\usepackage[most]{tcolorbox}

\newtcolorbox{promptbox}{
  colback=blue!3!white,     
  colframe=blue!18!white,   
  boxrule=0.6pt,            
  arc=0pt,                  
  left=2mm,                 
  right=2mm,                
  top=2mm,                  
  bottom=2mm,               
  before skip=0.75em,       
  after skip=1.0em,         
  breakable                 
}

\usepackage[T1]{fontenc}

\usepackage[utf8]{inputenc}

\usepackage{microtype}

\usepackage{inconsolata}

\usepackage{graphicx}
\setkeys{Gin}{draft=false}

\title{From Causal Plausibility to Causal Reliability: Evaluating LLMs as Calibrated Direct Causal-Edge Classifiers}

\author{
Amit Kumar\textsuperscript{1} \quad
Elnur Adl Zarabi\textsuperscript{1} \quad
Suranjana Trivedy\textsuperscript{2} \quad
Zhiqian Chen\textsuperscript{3} \\
Lei Zhang\textsuperscript{4} \quad
Kaiqun Fu\textsuperscript{5} \quad
Taoran Ji\textsuperscript{1} \\
\textsuperscript{1}Texas A\&M University-Corpus Christi, USA \\
\textsuperscript{2}BITS Pilani Goa, India \\
\textsuperscript{3}Mississippi State University, USA \\
\textsuperscript{4}Northern Illinois University, USA \\
\textsuperscript{5}Texas Christian University, USA
}

\begin{document}
\maketitle

\begin{abstract}

Large language models (LLMs) are increasingly used to provide prior causal knowledge for structural causal discovery, yet whether their direct-edge judgments and associated confidence can be trusted remains unclear. We systematically evaluate 12 instruction-tuned open-weight models across six benchmark causal graphs, five prompting strategies, and four confidence sources: verbalized, logit-based, cross-prompt agreement, and cross-model agreement. Our evaluation yields three key findings. (i) LLM-based causal judgments are strongly recall-dominant. Models tend to predict overly dense graphs with many false-positive edges, while prompting primarily shifts the precision--recall trade-off rather than consistently resolving overprediction. Improvements with model scale also diminish on the largest graphs and do not eliminate miscalibration. (ii) LLMs often capture causal relatedness without reliably identifying directness or orientation. Relative to the published reference graphs, models incorrectly classify 40.0\% of indirect and 36.0\% of reversed non-edges as direct causal edges, compared with 28.2\% of other non-edges. Moreover, 80.8\% and 84.6\% of these false positives receive verbalized confidence of at least 80\%, revealing substantial overconfidence in structurally incorrect predictions. (iii) Conventional confidence estimates are unreliable, whereas agreement provides a more promising signal. Logit-based confidence frequently collapses near 1.0 regardless of correctness, while cross-prompt and cross-model agreement achieve better mean calibration and discrimination, although their advantages are not statistically significant after Holm correction. A benchmark-familiarity audit additionally identifies potential familiarity in five model--dataset pairs, all involving AsiaM. Overall, our results suggest that LLMs are better viewed as sources of externally validated soft causal priors than as direct evidence of causal structure. Replication materials are available on \href{https://github.com/aamitssharma07/calibrated-llm-causal-discovery}{GitHub}.

\end{abstract}
\section{Introduction}\label{sec:intro}

Causal discovery (CD) aims to infer directed causal relationships among variables, typically represented as a causal graph, to support explanation, intervention analysis, and scientific understanding \citep{pearl2009causality}. For example, consider a simplified medical graph \(\textit{Viral Exposure} \rightarrow \textit{Infection} \rightarrow \textit{Positive Test}\). Within this graph, adjacent variables are connected by direct edges, whereas \textit{Viral Exposure} influences \textit{Positive Test} only indirectly through \textit{Infection}. Traditional approaches, including constraint-based methods such as PC and FCI \citep{spirtes2000causation}, score-based methods such as GES \citep{chickering2002optimal} and NOTEARS \citep{zheng2018dags}, and neural formulations such as GraN-DAG \citep{lachapelle2020gradient}, operate on observational data under assumptions about the data-generating process. However, observational data may be limited or insufficient to distinguish Markov-equivalent structures \citep{glymour2019review}, motivating the use of domain knowledge to guide or orient candidate graphs \citep{vashishtha2023causal}. Such knowledge is often costly to elicit or unavailable in novel domains.

\begin{figure}[H]
    \centering
    \includegraphics[width=\columnwidth,keepaspectratio]{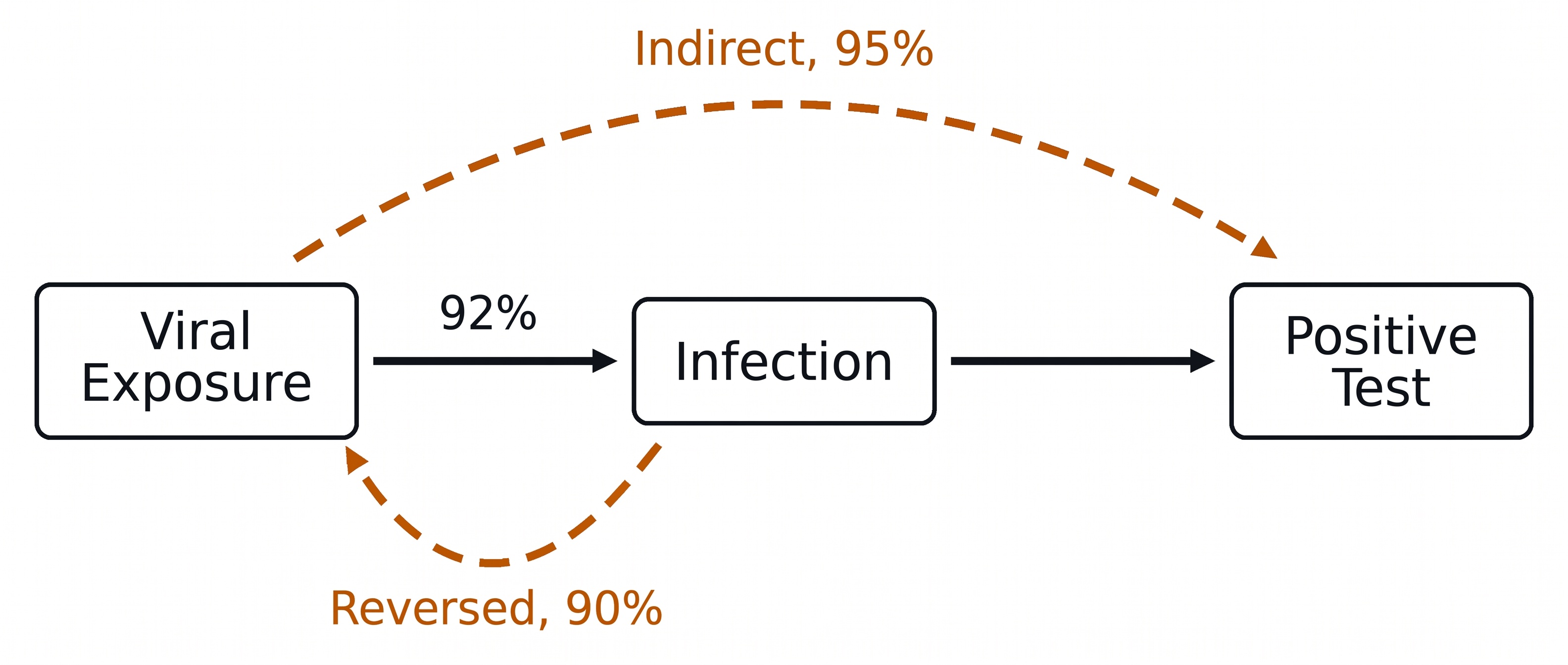}
    \caption{Illustration of unreliable pairwise direct-edge judgments. Solid arrows form the reference chain; dashed arrows show an indirect relation predicted as direct and a reversed relation. Confidence values are illustrative.}
    \label{fig:causal_reliability_problem}
\end{figure}

Large language models (LLMs) offer a complementary source of causal knowledge for CD, with recent work exploring direct causal reasoning \citep{kiciman2024causal}, prior-guided structure learning \citep{vashishtha2023causal,ban2023from,kampani2024llm}, and graph orientation and refinement \citep{long2023imperfect,ban2023causal}. Although LLMs demonstrate non-trivial performance on pairwise causal tasks using textual metadata \citep{kiciman2024causal}, their predictions may depend on the frequency of causal relations in pre-training corpora and vary under contextual changes \citep{feng-etal-2025-reliability}. Prompt style and response agreement can also affect confidence calibration \citep{xia2025influences}. Existing work shows that pairwise LLM-based edge classification can yield poor graph recovery \citep{babakov2025causalgraphbench}, but does not systematically identify the structural sources of these errors or whether confidence reflects their correctness. Because such judgments are used as structural constraints, optimization priors, and causal-order information in CD pipelines \citep{ban2023from,kampani2024llm,vashishtha2023causal}, their reliability remains important even when pairwise classification is insufficient as a standalone CD method.

This gap reflects a distinction between causal plausibility and causal reliability. For queried variables \(A\) and \(B\), a reliable direct-edge judgment requires distinguishing \(A \rightarrow B\) from reverse directionality, indirect influence, and no direct edge. A model may recognize causal relatedness while misidentifying directness or orientation, producing many spurious edges in sparse graphs. Confidence reliability is therefore as important as classification performance when LLM judgments inform graph learning. Figure~\ref{fig:causal_reliability_problem} illustrates this reliability gap.

Accordingly, our aim is to characterize the reliability of edge judgments on which LLM-assisted CD pipelines may depend, rather than to propose a new CD method. Our contributions are as follows:
\begin{itemize}
    \item We formulate pairwise direct causal-edge classification as a calibrated evaluation task, jointly assessing edge prediction quality, graph reconstruction, and confidence reliability under structural ambiguity and severe class imbalance.

    \item We develop an evaluation framework spanning six benchmark causal graphs, five prompting strategies, 12 instruction-tuned models at two scales, and four confidence sources: verbalized, logit-based, cross-prompt agreement, and cross-model agreement. We also evaluate potential benchmark familiarity across all 72 model--dataset combinations. We will release the code and evaluation artifacts to support reproducibility.

    \item We show that LLM edge judgments exhibit recall-dominant behavior and prompt sensitivity. Larger models achieve higher mean F1 on five of six datasets, but provide only marginal gains on the largest graphs and do not resolve miscalibration. Relative to the published reference graphs, false-positive rates are higher for indirect (40.0\%) and reversed (36.0\%) non-edges than for other non-edges (28.2\%); over 80\% of false positives on indirect and reversed relations receive verbalized confidence of at least 80\%. A three-way task-formulation ablation further shows that these errors are not solely induced by evaluating the two directions through separate binary queries.

    \item We find that agreement-based confidence achieves better mean calibration and discrimination than verbalized and logit-based confidence, although the differences are not statistically significant after Holm correction and remain model- and prompt-dependent.
\end{itemize}

\begin{figure*}[!t]
    \centering
    \includegraphics[draft=false,width=0.88\textwidth]{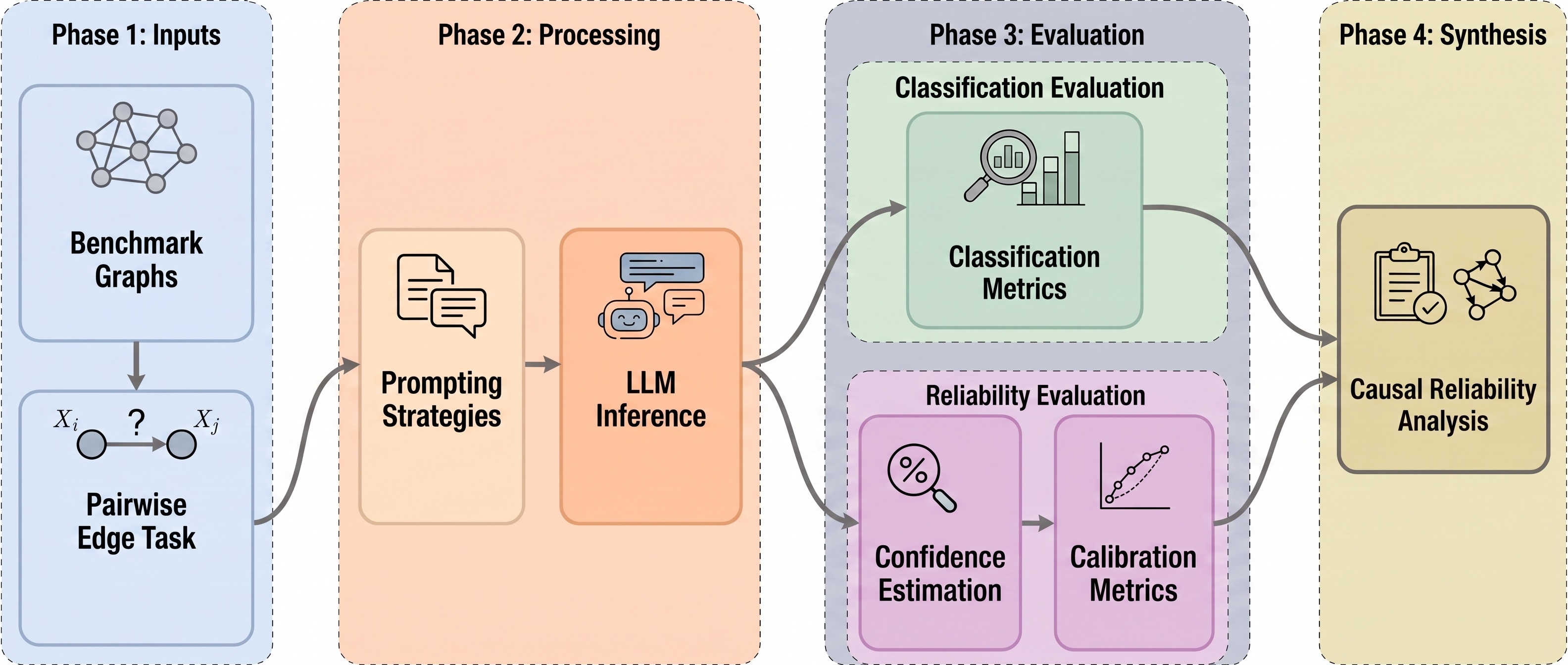}
    \caption{Evaluation pipeline for pairwise direct-edge classification, graph reconstruction, confidence calibration, and structural error analysis.}
    \label{fig:calibration_pipeline}
    \vspace{-0.8em}
\end{figure*}

\section{Evaluation Design}\label{sec:eval}

The framework in Fig.~\ref{fig:calibration_pipeline} treats LLMs as calibrated pairwise direct causal-edge classifiers. Across two model-scale groups, five prompting strategies, and four confidence sources, we ask whether models avoid edge overprediction, distinguish direct edges from reversed, indirect, and other non-edges, and assign confidence that reflects correctness. Pairwise predictions are compared with published reference graphs and aggregated into reconstructed graphs to assess edge classification and graph recovery. Structural error analysis, confidence thresholding, a three-way label-space ablation, and a benchmark-familiarity audit further test these reliability concerns.

\subsection{Prompt Styles}
\label{subsec:prompt_styles}



Prompt formulation affects LLM performance \citep{white2023prompt,chen2023unleashing,qiao2023reasoning}. We adapt pairwise causal prompting from prior work on causal reasoning and LLM-guided structure discovery \citep{kiciman2024causal,vashishtha2023causal} to direct-edge classification with explicit direct-causation instructions and verbal confidence elicitation \citep{lin2022teaching}.

We evaluate five prompt styles that provide progressively richer guidance. \textbf{Name-only} supplies the dataset context and variable names, testing whether the model can infer a direct edge from names alone. \textbf{Metadata} additionally provides variable definitions to clarify their domain-specific meanings. \textbf{Chain-of-Thought (CoT)} requests a brief rationale before the prediction \citep{wei2022chain}. \textbf{Few-shot (FS)} provides five labeled direct-edge examples that demonstrate the required task and output format \citep{brown2020language}. \textbf{Few-shot + CoT (FS+CoT)} combines these demonstrations with brief reasoning. All styles ask whether \(A\) directly causes \(B\) and require a binary ``Yes''/``No'' judgment with a self-reported confidence score from 0--100. Full templates appear in Appendix~\ref{app:prompts}.

\subsection{Direct Edge Classification and Graph Reconstruction}
\label{subsec:primary_classification}

The primary task is binary direct-edge classification. Given an ordered pair \((A,B)\) and dataset context, the model predicts whether \(A\) directly causes \(B\), with \textsc{Yes} as the positive class and \textsc{No} as the negative class. Responses follow a standardized answer--confidence format and are parsed deterministically, prioritizing labeled fields and then answer-first forms while ignoring prompt echoes. Missing fields trigger one retry; fewer than 0.2\% of responses remain without a valid label and are excluded from classification.

For each model--prompt--dataset setting, all positively predicted ordered pairs are combined to form a reconstructed directed graph. We compare this reconstructed graph with the published reference graph using normalized Structural Hamming Distance (nSHD), defined in Section~\ref{subsec:evaluation_metrics}. Graph reconstruction is induced directly from pairwise predictions and does not enforce global structural constraints such as acyclicity.

\subsection{Confidence Estimation Task}
\label{subsec:confidence_estimation}


This task assesses whether confidence reflects correctness in binary edge classification. Let \(x\) denote an input query, \(\mathcal{Y}=\{\textsc{Yes},\textsc{No}\}\) the label set, and \(\hat{y}\in\mathcal{Y}\) the predicted label. We consider four confidence sources, all normalized to \([0,1]\).

\paragraph{Verbalized confidence.}
The model reports a confidence score \(s_{\mathrm{verb}}\in[0,100]\) with its binary prediction \citep{lin2022teaching,tian2023just}. We define
\begin{equation}
c_{\mathrm{verb}}(\hat{y})=\frac{s_{\mathrm{verb}}}{100}.
\end{equation}

\paragraph{Logit-based confidence.}
We aggregate valid surface forms for each label and normalize their decoder logits over \(\mathcal{Y}\) \citep{jiang2021know,tian2023just}:
\begin{equation}
p(y\mid x)=
\frac{\exp(z_y)}
{\sum_{y'\in\mathcal{Y}}\exp(z_{y'})}.
\end{equation}
The predicted label and its confidence are
\begin{equation}
\hat{y}=\operatorname*{argmax}_{y\in\mathcal{Y}}p(y\mid x),
\qquad
c_{\mathrm{logit}}(\hat{y})=
\max_{y\in\mathcal{Y}}p(y\mid x).
\end{equation}

\paragraph{Cross-prompt agreement.}
For \(K=5\) prompt-specific predictions \(y^{(1)},\ldots,y^{(K)}\), let \(\hat{y}\) be the majority label. Confidence is the fraction of prompts supporting it \citep{wang2023selfconsistency,xiong2024can}:
\begin{equation}
c_{\mathrm{prompt}}(\hat{y})=
\frac{1}{K}\sum_{k=1}^{K}
\mathbf{1}\!\left[y^{(k)}=\hat{y}\right].
\end{equation}

\paragraph{Cross-model agreement.}
For \(M\) predictions \(y_1,\ldots,y_M\) under a fixed prompt, let \(\hat{y}\) be the majority label. Computed separately within each model-scale group, confidence is \citep{xia2025influences,zhang-etal-2024-dont-go}
\begin{equation}
c_{\mathrm{model}}(\hat{y})=
\frac{1}{M}\sum_{m=1}^{M}
\mathbf{1}\!\left[y_m=\hat{y}\right].
\end{equation}

\section{Experiment}\label{sec:expt}


\subsection{Models}\label{subsec:models}

We evaluate 12 instruction-tuned open-weight models from five families: Qwen, Gemma, Llama, Mistral, and Phi \citep{yang2025qwen3,yang2024qwen25,google2026gemma4,
grattafiori2024llama,meta2024llama33,mistral2024ministral8b,
abouelenin2025phi4mini,abdin2024phi4}, accessed through Hugging Face.\footnote{Hugging Face models: \url{https://huggingface.co/models}.} To examine scale effects, we group models by parameter count.

\textbf{Small models (4--14B):} Qwen3-4B-Instruct, Qwen3-8B-Instruct, Gemma-4-E4B-IT, Llama-3.1-8B-Instruct, Ministral-8B-Instruct-2410, Phi-4-Mini-Instruct, and Phi-4.

\textbf{Large models (31--72B):} Qwen3-32B-Instruct, Qwen2.5-72B-Instruct, Gemma-4-31B-IT, Llama-3.1-70B-Instruct, and Llama-3.3-70B-Instruct.

Figures abbreviate Gemma-4-31B-IT as Gemma-31B, Gemma-4-E4B-IT as Gemma-E4B, and Llama-3.1-8B-Instruct as Llama-8B.

\subsection{Datasets}
\label{subsec:dataset}


We use six graphs from CausalGraphBench \citep{babakov2025causalgraphbench}: AsiaM, River Status, COVID, Coal Gasifier, Hepar2, and Munin1. They span four size categories and medical, ecological, public-health, and industrial domains. Dataset statistics, benchmark IDs, and preprocessing details appear in Appendix~\ref{app:datasets}, Table~\ref{tab:dataset_stats}.


This subset provides a controlled size- and domain-stratified evaluation. Pairwise classification scales as $n(n{-}1)$, and the six graphs already require approximately 2.5M queries, making full 35-graph evaluation prohibitive. We acknowledge in the Limitations that results may not capture full benchmark diversity.

Each dataset provides a domain description, variable definitions, and a published reference graph. The pairwise setting is highly imbalanced—only a small fraction of ordered pairs are true edges, worsening with graph size. This motivates our calibration focus, as aggregate accuracy can mask overconfidence on sparse edge classes.

\subsection{Evaluation Metrics}
\label{subsec:evaluation_metrics}

We evaluate performance along two dimensions: direct-edge classification and confidence calibration. For classification, we report Precision, Recall, and F1 over ordered variable pairs \citep{davis2006relationship}. To compare structural recovery across datasets of different sizes, we also report normalized Structural Hamming Distance (nSHD), based on SHD \citep{tsamardinos2006max}. For dataset \(d\), nSHD is defined as:
\begin{equation}
\mathrm{nSHD}_d = \frac{\mathrm{SHD}_d}{n_d(n_d-1)},
\label{eq:nshd}
\end{equation}
where \(n_d\) is the number of variables in dataset \(d\).

For calibration, we report 10-bin squared-gap ECE, based on the binning framework of \citet{guo2017calibration}, Brier Score \citep{brier1950verification}, and AUROC \citep{ulmer2024calibrating}. Lower ECE and Brier indicate better confidence–correctness alignment; higher AUROC indicates stronger discrimination between correct and incorrect predictions.

\subsection{Experimental Settings}
\label{subsec:experimental_settings}

Few-shot prompts use five fixed demonstrations: three positive and two negative examples, held constant across models. We use temperature 1.0, repetition penalty 1.1, and maximum generation lengths of 128 tokens for direct-answer prompts and 384 tokens for reasoning prompts.

\section{Results and Analysis}
\label{sec:res}
\setlength{\textfloatsep}{8pt plus 2pt minus 2pt}
\setlength{\dbltextfloatsep}{8pt plus 2pt minus 2pt}
\setlength{\floatsep}{6pt plus 2pt minus 2pt}
\setlength{\intextsep}{6pt plus 2pt minus 2pt}

\subsection{Primary Edge Classification}
\label{subsec:primary_edge_classification}

We analyze how prompt design, dataset characteristics, model scale, and potential benchmark familiarity affect direct-edge classification. We first examine the general precision--recall behavior and prompt sensitivity, then assess whether performance differences are associated with dataset characteristics and model scale. We finally evaluate potential benchmark familiarity and whether a three-way label space better distinguishes edge direction from the absence of a direct edge.

\FloatBarrier
\subsubsection{Recall-Dominant Behavior and Prompt Sensitivity}

\begin{figure*}[!t]
    \centering
    \includegraphics[width=\textwidth,keepaspectratio]{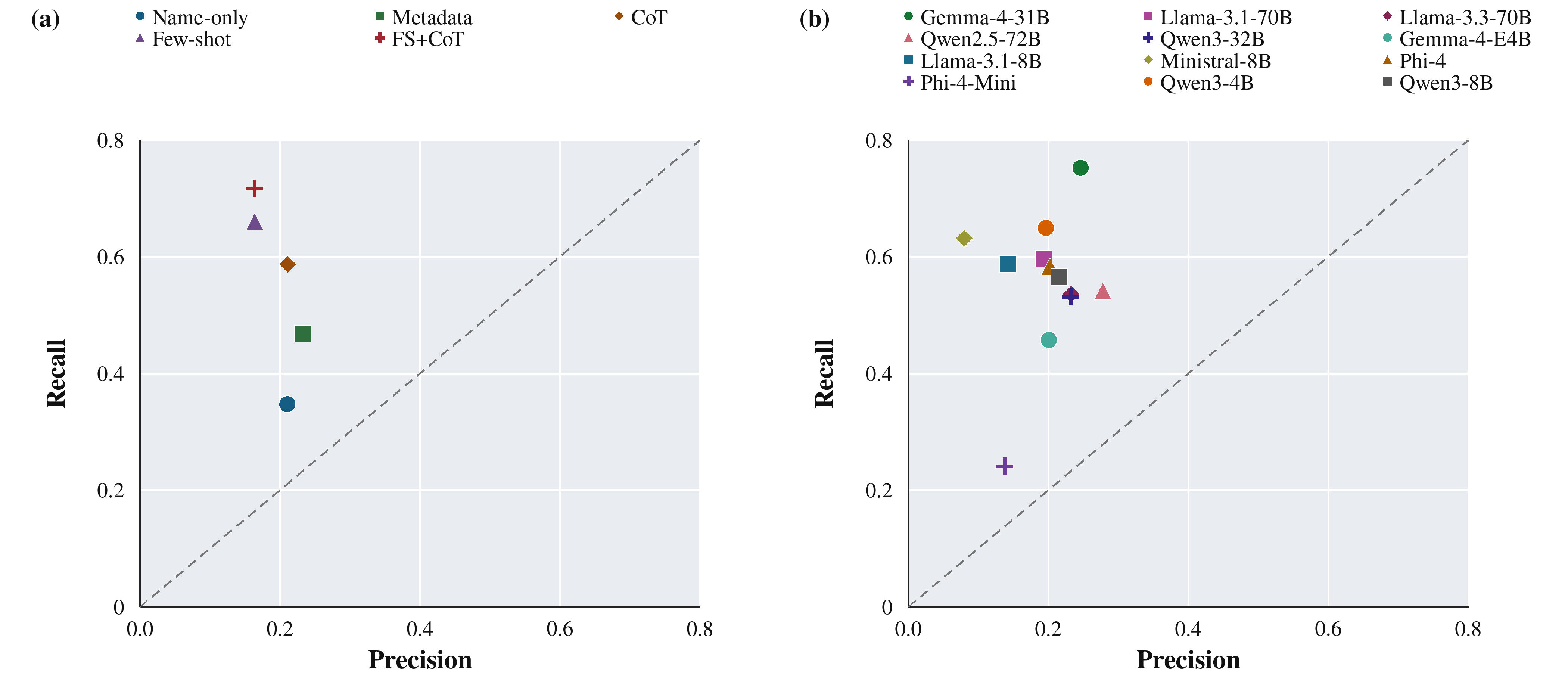}
    \caption{Precision--recall behavior by (a) prompt, averaged across models and datasets, and (b) model, averaged across prompts and datasets. The dashed diagonal denotes equal precision and recall; points above it are recall-dominant.}
    \label{fig:prompt_level_precision_recall_tradeoff}
    \label{fig:model_level_precision_recall_behavior}
\end{figure*}

The results reveal a consistent recall-dominant pattern across both model groups. Recall exceeds precision in 68 of 72 model--dataset combinations (94.4\%) after averaging across prompts, and in 305 of 360 model--dataset--prompt combinations (84.7\%). Figure~\ref{fig:prompt_level_precision_recall_tradeoff} summarizes this pattern at the prompt and model levels, macro-averaged across the remaining dimensions. Nearly all aggregates lie above the equal-precision--recall diagonal.

Although the magnitude of this behavior varies across settings, its direction remains stable. In Table~\ref{tab:main_prf1_nshd}, mean recall exceeds mean precision for every dataset and model-scale group, with recall standard deviations of 0.173--0.316 across model--prompt combinations. Models therefore predict edges for a large fraction of ordered pairs, producing dense graphs with many false positives. This pattern suggests that LLMs often treat causal relatedness as evidence of direct causation.

\begin{table*}[!t]
\caption{Dataset-level edge-classification and graph-reconstruction performance. Values are mean $\pm$ SD across model--prompt combinations within each scale group (S: 7 models; L: 5 models). Higher precision, recall, and F1 and lower nSHD indicate better performance.}
\label{tab:main_prf1_nshd}
\centering
\footnotesize
\setlength{\tabcolsep}{3.5pt}
\renewcommand{\arraystretch}{1.10}

\resizebox{\textwidth}{!}{%
\begin{tabular}{@{}l cc cc cc cc@{}}
\toprule
\multirow{2}{*}{\textbf{Dataset}}
& \multicolumn{2}{c}{\textbf{Precision} $\uparrow$}
& \multicolumn{2}{c}{\textbf{Recall} $\uparrow$}
& \multicolumn{2}{c}{\textbf{F1} $\uparrow$}
& \multicolumn{2}{c}{\textbf{nSHD} $\downarrow$} \\
\cmidrule(lr){2-3}
\cmidrule(lr){4-5}
\cmidrule(lr){6-7}
\cmidrule(lr){8-9}
& \textbf{S} & \textbf{L}
& \textbf{S} & \textbf{L}
& \textbf{S} & \textbf{L}
& \textbf{S} & \textbf{L} \\
\midrule

AsiaM
& $0.415 \pm 0.175$ & $\mathbf{0.608 \pm 0.139}$
& $0.707 \pm 0.264$ & $\mathbf{0.830 \pm 0.210}$
& $0.478 \pm 0.149$ & $\mathbf{0.680 \pm 0.126}$
& $0.281 \pm 0.145$ & $\mathbf{0.141 \pm 0.062}$ \\

River Status
& $0.198 \pm 0.048$ & $\mathbf{0.216 \pm 0.062}$
& $0.592 \pm 0.213$ & $\mathbf{0.597 \pm 0.292}$
& $0.281 \pm 0.050$ & $\mathbf{0.299 \pm 0.107}$
& $0.357 \pm 0.139$ & $\mathbf{0.292 \pm 0.090}$ \\

COVID
& $0.261 \pm 0.181$ & $\mathbf{0.412 \pm 0.171}$
& $0.443 \pm 0.221$ & $\mathbf{0.545 \pm 0.252}$
& $0.280 \pm 0.144$ & $\mathbf{0.403 \pm 0.080}$
& $0.191 \pm 0.158$ & $\mathbf{0.104 \pm 0.039}$ \\

Coal Gasifier
& $0.049 \pm 0.029$ & $\mathbf{0.069 \pm 0.057}$
& $0.428 \pm 0.247$ & $\mathbf{0.505 \pm 0.316}$
& $0.078 \pm 0.045$ & $\mathbf{0.095 \pm 0.053}$
& $0.288 \pm 0.204$ & $\mathbf{0.225 \pm 0.159}$ \\

Hepar2
& $0.062 \pm 0.021$ & $\mathbf{0.086 \pm 0.024}$
& $0.546 \pm 0.173$ & $\mathbf{0.593 \pm 0.218}$
& $0.106 \pm 0.030$ & $\mathbf{0.143 \pm 0.029}$
& $0.265 \pm 0.144$ & $\mathbf{0.191 \pm 0.097}$ \\

Munin1
& $\mathbf{0.018 \pm 0.016}$ & $\mathbf{0.018 \pm 0.008}$
& $0.468 \pm 0.259$ & $\mathbf{0.480 \pm 0.295}$
& $\mathbf{0.031 \pm 0.019}$ & $\mathbf{0.031 \pm 0.014}$
& $0.326 \pm 0.241$ & $\mathbf{0.269 \pm 0.219}$ \\

\bottomrule
\end{tabular}%
}

\end{table*}

We further observe that this behavior persists across prompt styles. FS, CoT, and FS+CoT often shift models toward higher recall without comparable precision gains. Appendix Figure~\ref{fig:promptwise_f1_all_models} also shows that the best-performing prompt varies across model--dataset combinations. Thus, prompt design affects the precision--recall tradeoff, but no prompt style consistently achieves a better balance across models and datasets.

\subsubsection{Effect of Graph Size and Semantic Accessibility}

Having established that prompting does not eliminate recall-dominant behavior, we next examine why the balance between precision and recall varies across datasets. We use F1 to summarize this balance and compare edge-classification performance across graphs. Table~\ref{tab:main_prf1_nshd} shows that F1 generally declines as graph size and class imbalance increase: mean F1 falls from 0.281 and 0.299 on \textsc{River Status} to 0.031 for both model groups on \textsc{Munin1}. Although the SDs indicate variation across model--prompt combinations, the largest graphs consistently yield low F1. Appendix Figure~\ref{fig:dataset_size_avg_f1_by_model} and Table~\ref{tab:dataset_avg_classification_combined} report the complete results.

Graph size alone does not fully explain these differences. We therefore examine semantic accessibility using description length, code-like variable names, and acronym density (Appendix Table~\ref{tab:metadata_readability_indicators}). Considering all 12 models, even the lowest model-level mean F1 on \textsc{River Status} (0.220) exceeds the highest achieved on \textsc{Coal Gasifier} (0.121) and \textsc{Munin1} (0.050), with each model averaged across five prompts. This contrast is consistent with the natural-language descriptions in \textsc{River Status} and the more technical metadata in the latter datasets. \textsc{Munin1} is particularly challenging, combining 186 variables with 100\% code-like names and 27.4\% acronym-containing descriptions.

However, metadata complexity alone is insufficient. \textsc{COVID} achieves comparatively higher F1 despite having 100\% code-like names and 45\% acronym-containing descriptions, as its metadata refers to broadly recognizable public-health concepts. Overall, these exploratory indicators suggest that performance is jointly associated with graph size, class imbalance, and the linguistic accessibility of domain concepts; they do not establish causal effects. Overall, these exploratory indicators suggest that performance is jointly associated with graph size, class imbalance, and the linguistic accessibility of domain concepts; they do not establish causal effects. This variation also motivates assessing whether greater model scale mitigates these constraints.

\subsubsection{Effect of Model Scale}

\begin{figure}[H]
    \centering
    \includegraphics[width=\columnwidth,keepaspectratio,trim=34.2bp 11.4bp 8.5bp 10bp,clip]{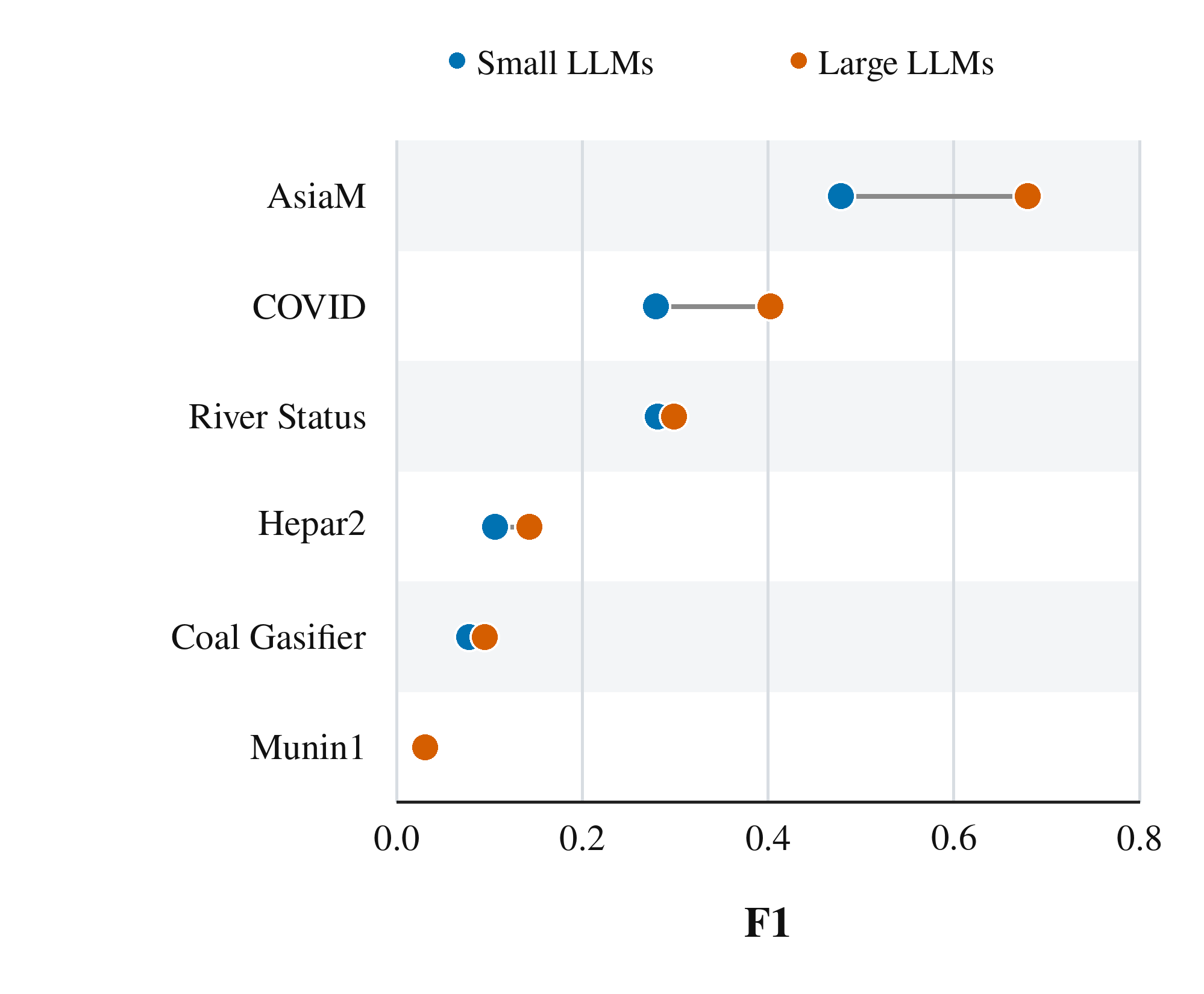}
    \caption{Dataset-level mean F1 for small and large LLMs, averaged across models and prompts within each group. Lines connect group means; rightward shifts favor large LLMs.}
    \label{fig:model_scale_f1_dumbbell}
\end{figure}

 Model scale improves mean F1 on five of six datasets, but the gains are uneven. Figure~\ref{fig:model_scale_f1_dumbbell} compares small and large LLMs across datasets. The largest improvements occur on \textsc{AsiaM} ($0.478$ to $0.680$) and \textsc{COVID} ($0.280$ to $0.403$). In contrast, improvements are modest on \textsc{River Status} ($+0.018$), \textsc{Coal Gasifier} ($+0.017$), and \textsc{Hepar2} ($+0.037$), while performance is effectively unchanged on \textsc{Munin1}.

Scale also does not eliminate recall-dominant behavior. Figure~\ref{fig:model_level_precision_recall_behavior} shows that nearly all models remain above the equal-precision--recall diagonal, while Table~\ref{tab:main_prf1_nshd} shows recall exceeding precision for both scale groups on every dataset. Thus, larger models can improve edge recovery, particularly on semantically accessible graphs, but provide limited gains on the most difficult datasets and do not resolve systematic edge overprediction. The especially large improvement on \textsc{AsiaM} warrants qualification through the benchmark-familiarity audit presented next.

\subsubsection{Potential Benchmark Familiarity Audit}
\label{subsec:contamination_analysis}

The benchmark-familiarity audit flags five model--dataset pairs as high risk, all involving \textsc{AsiaM}; no model is flagged on the other five datasets. Recall-dominant overprediction and low performance on larger graphs persist across these unflagged datasets, indicating that the principal classification patterns are not driven by \textsc{AsiaM} alone.

Following \citet{babakov2025causalgraphbench}, we apply a two-stage node- and structure-recall test to all 72 model--dataset pairs. Each model first reproduces a graph's nodes using only its source reference and domain description. Generated and reference nodes are aligned through semantic matching using Mixtral-8x7B-Instruct, an independent judge not included among the evaluated models. A pair is flagged when node-count deviation is below 15\% and node recall exceeds 0.85; only flagged pairs proceed to structure recall.
\begin{table}[H]
\caption{Structure recall for model--dataset pairs passing the benchmark-familiarity screen, all on AsiaM (7 nodes, 8 edges). Nodes reports generated/true counts, Dev. is node-count deviation, and SHD is the number of edge edits.}
\label{tab:contamination_flagged}
\centering
\footnotesize
\setlength{\tabcolsep}{3.5pt}
\renewcommand{\arraystretch}{1.08}
\resizebox{\columnwidth}{!}{%
\begin{tabular}{lccccc}
\toprule
\textbf{Model}
& \textbf{Nodes}
& \textbf{Recall}
& \textbf{Dev.}
& \textbf{Edge F1}
& \textbf{SHD$\downarrow$} \\
\midrule
Phi-4          & 8/7 & 0.857 & 14.3\% & 0.500 & 6 \\
Qwen3-32B      & 7/7 & 0.857 &  0.0\% & 0.714 & 4 \\
Qwen2.5-72B    & 8/7 & 1.000 & 14.3\% & 0.857 & 2 \\
Gemma-4-31B    & 7/7 & 1.000 &  0.0\% & 0.800 & 3 \\
Llama-3.3-70B  & 8/7 & 1.000 & 14.3\% & 0.941 & 1 \\
\bottomrule
\end{tabular}%
}
\end{table}
The flagged pairs show varying degrees of structure recovery (Table~\ref{tab:contamination_flagged}), indicating potential familiarity rather than uniform graph memorization. Notably, Gemma-4-31B-IT is flagged, suggesting that familiarity may partly contribute to the strong model-scale gain observed on \textsc{AsiaM}. We therefore interpret results on this dataset cautiously, while the principal classification conclusions remain supported by the five unflagged datasets. Full results appear in Appendix Table~\ref{tab:full_node_contamination}.

\begin{figure*}[!t]
    \centering
    \includegraphics[width=\textwidth]{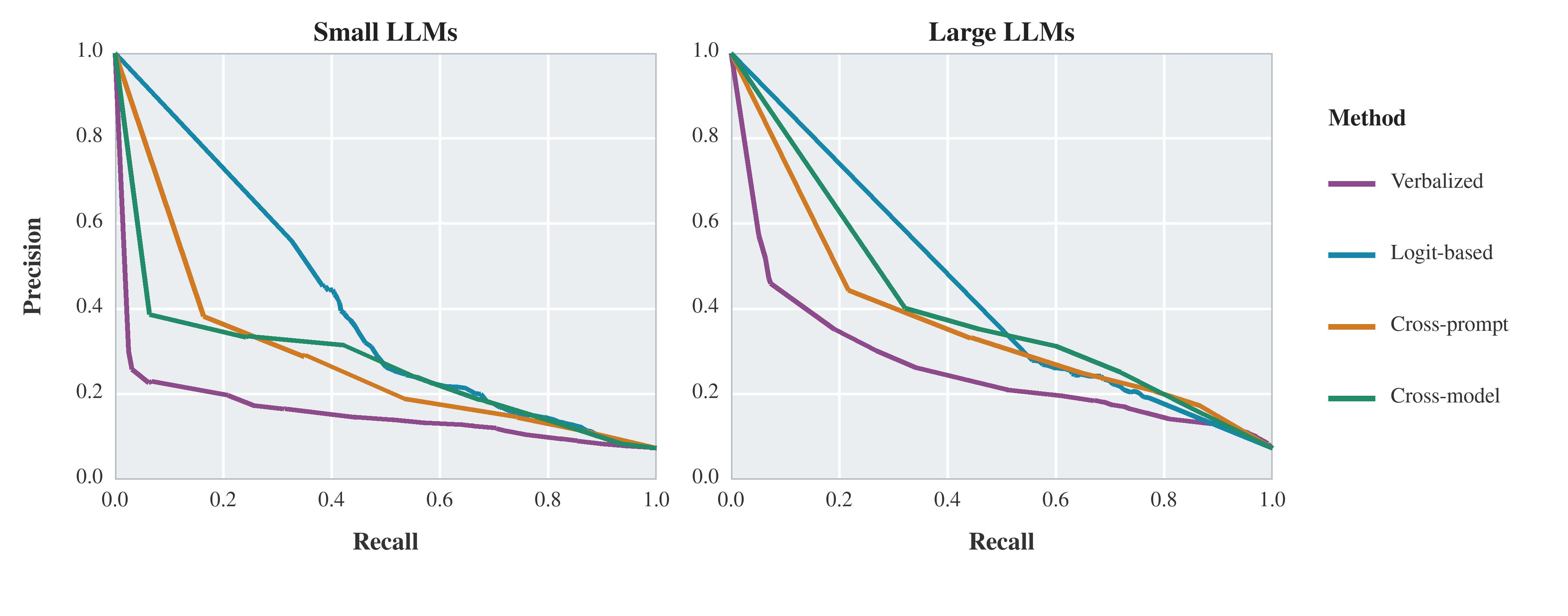}
    \caption{Edge precision--recall curves obtained by thresholding positive-edge confidence for small and large model groups, macro-averaged across six datasets.}
    \label{fig:pr_curves}
    \label{fig:edge_precision_recall_curves}
\end{figure*}

\subsection{Confidence Calibration}
\label{subsec:confidence_calibration}
Having established that LLMs frequently overpredict causal edges, we examine whether confidence can identify which predictions are reliable. This is important when LLM judgments are used as causal priors: informative confidence can support edge weighting, thresholding, and graph sparsification, whereas overconfident errors can reinforce incorrect structure. We compare four confidence sources in terms of calibration and discrimination, examine their sensitivity to prompt style, and test whether post-hoc temperature scaling improves logit-based calibration.

\subsubsection{Agreement Supports Edge Ranking but Does Not Guarantee Instance-Level Calibration}
\label{subsubsec:agreement_based_confidence_auc}
\label{subsubsec:concentrated_confidence_calibration}

Agreement-based confidence provides the strongest aggregate reliability signal, although its advantage varies across datasets and model groups. Table~\ref{tab:calibration_results} shows that cross-model agreement achieves the highest mean AUROC for small models ($0.692\pm0.043$), while cross-prompt agreement performs best for large models ($0.772\pm0.047$). Both methods also achieve lower mean ECE and Brier scores than verbalized and logit-based confidence. However, pairwise two-sided Wilcoxon tests over the six datasets show no significant differences after Holm adjustment (minimum adjusted $p=0.1875$; Appendix Table~\ref{tab:wilcoxon_calibration}). Agreement therefore performs better descriptively, but its statistical superiority across datasets is not established.

Its practical value is clearest when confidence is used to rank or filter candidate edges. Correctness AUROC evaluates whether confidence separates correct from incorrect predictions, whereas edge AUPRC treats the presence of a reference edge as the positive class. We assign positive-edge confidence \(c\) to \textsc{Yes} predictions and \(1-c\) to \textsc{No} predictions; logit-based confidence uses the normalized \textsc{Yes} probability. Figure~\ref{fig:edge_precision_recall_curves} shows that cross-model agreement achieves the highest edge AUPRC for both small ($0.228$) and large ($0.269$) LLMs. For large models, a cross-prompt threshold of $0.8$ reduces graph density by $44.7\%$ and increases precision from $0.250$ to $0.331$, while F1 remains nearly unchanged ($0.314$ to $0.310$). Agreement can therefore support controllable graph sparsification, although increasing precision generally reduces recall.

\begin{table}[H]
\caption{Calibration by model group and confidence source, reported as mean $\pm$ SD across six datasets. ECE denotes squared-gap ECE. Bold and underlining indicate the best and second-best results within each group.}
\label{tab:calibration_results}
\centering
\footnotesize
\setlength{\tabcolsep}{2.5pt}
\renewcommand{\arraystretch}{1.12}
\begin{tabular}{@{}llcccc@{}}
\toprule
\textbf{Group}
& \textbf{Method}
& \textbf{ECE$\downarrow$}
& \textbf{Brier$\downarrow$}
& \textbf{AUROC$\uparrow$}
& \textbf{Acc.$\uparrow$} \\
\midrule

\multirow{4}{*}{Small}
& Cross-model
& \makecell{\textbf{0.019}\\\textbf{$\pm$0.013}}
& \makecell{\textbf{0.149}\\\textbf{$\pm$0.043}}
& \makecell{\textbf{0.692}\\\textbf{$\pm$0.043}}
& \makecell{0.792\\$\pm$0.071} \\

& Cross-prompt
& \makecell{\underline{0.038}\\\underline{$\pm$0.012}}
& \makecell{\underline{0.176}\\\underline{$\pm$0.043}}
& \makecell{\underline{0.661}\\\underline{$\pm$0.047}}
& \makecell{0.755\\$\pm$0.060} \\

& Logit-based
& \makecell{0.226\\$\pm$0.054}
& \makecell{0.354\\$\pm$0.046}
& \makecell{0.611\\$\pm$0.037}
& \makecell{0.549\\$\pm$0.064} \\

& Verbalized
& \makecell{0.229\\$\pm$0.034}
& \makecell{0.364\\$\pm$0.032}
& \makecell{0.358\\$\pm$0.059}
& \makecell{0.709\\$\pm$0.057} \\

\midrule

\multirow{4}{*}{Large}
& Cross-prompt
& \makecell{\textbf{0.016}\\\textbf{$\pm$0.012}}
& \makecell{\textbf{0.127}\\\textbf{$\pm$0.050}}
& \makecell{\textbf{0.772}\\\textbf{$\pm$0.047}}
& \makecell{0.820\\$\pm$0.070} \\

& Cross-model
& \makecell{\underline{0.024}\\\underline{$\pm$0.018}}
& \makecell{\underline{0.139}\\\underline{$\pm$0.056}}
& \makecell{\underline{0.704}\\\underline{$\pm$0.055}}
& \makecell{0.822\\$\pm$0.070} \\

& Verbalized
& \makecell{0.205\\$\pm$0.061}
& \makecell{0.320\\$\pm$0.063}
& \makecell{0.448\\$\pm$0.099}
& \makecell{0.794\\$\pm$0.072} \\

& Logit-based
& \makecell{0.217\\$\pm$0.053}
& \makecell{0.346\\$\pm$0.057}
& \makecell{0.582\\$\pm$0.021}
& \makecell{0.620\\$\pm$0.061} \\

\bottomrule
\end{tabular}
\end{table}

These aggregate gains do not guarantee informative instance-level uncertainty. When confidence is concentrated near aggregate accuracy, ECE can be low without separating correct from incorrect predictions. Appendix Figures~\ref{fig:verbalized_best_reliability}--\ref{fig:cross_model_best_reliability} show that verbalized and logit-based confidence can cluster within narrow high-confidence ranges, partly reflecting dominant non-edge behavior. Logit-based confidence is particularly concentrated near $1.0$ for both correct and incorrect predictions (Appendix Figure~\ref{fig:gemma_31b_metadata_confidence_distribution}).

\begin{figure}[t]
    \centering
    \includegraphics[width=\columnwidth,keepaspectratio]{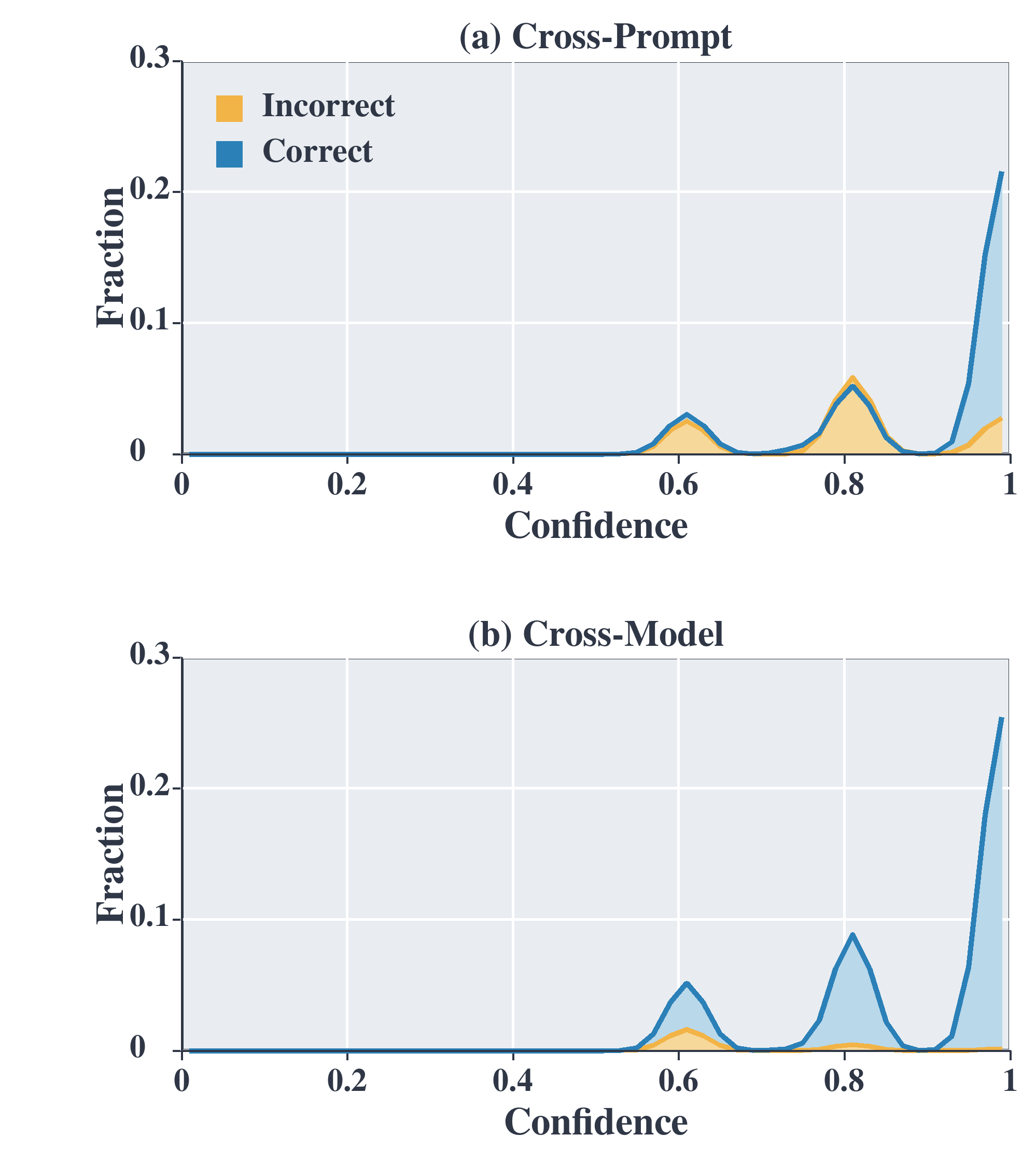}
    \vspace{-0.5em}
    \caption{Agreement-confidence distributions pooled across six datasets: (a) cross-prompt agreement for Gemma-4-31B-IT and (b) cross-model agreement under Metadata prompting. Correct predictions concentrate more strongly near full agreement.}
    \label{fig:gemma_metadata_agreement_confidence}
    \vspace{-0.8em}
\end{figure}
Agreement is less affected because it measures stability across prompts or models. Figure~\ref{fig:gemma_metadata_agreement_confidence} shows that correct predictions concentrate more strongly near full agreement than incorrect predictions. Nevertheless, agreement should be interpreted as an aggregate stability signal rather than a calibrated probability of correctness. Because prompt formulation is one such inference condition, the following subsection examines how prompt style affects calibration.

\subsubsection{Effects of Prompt Style on Calibration}
\label{subsubsec:prompt_style_calibration_effects}

Prompt formulation affects both causal-edge predictions and how confidence is expressed, but no prompt style consistently provides the best calibration across models and datasets. Appendix Tables~\ref{tab:calib_prompt_small_models} and~\ref{tab:calib_prompt_large_models} show substantial variation across prompt styles and confidence sources. For example, CoT reduces verbalized ECE for Gemma-4-E4B-IT on \textsc{COVID}, \textsc{Hepar2}, and \textsc{Munin1}, but the corresponding AUROC does not consistently improve. Prompting can therefore shift confidence closer to aggregate accuracy without making it more informative for distinguishing correct from incorrect predictions.

This instability provides the rationale for cross-prompt agreement, which measures whether an edge judgment remains stable across all five prompt formulations rather than relying on confidence from one prompt. Its reliability nevertheless remains model- and dataset-dependent, so prompt agreement should not be treated as uniformly calibrated. The instability of direct confidence across prompt settings also raises whether its calibration can be improved after inference, which we examine next through post-hoc temperature scaling.

\subsubsection{Post-hoc Temperature Scaling}
\label{subsubsec:post_hoc_temperature_sensitivity}

Post-hoc temperature scaling changes logit-based calibration in a model-specific way but cannot correct errors in the underlying causal-edge decision. We evaluate this without rerunning the LLMs by rescaling the saved Yes/No logits and recomputing normalized binary confidence. The predicted label remains unchanged because positive temperature scaling preserves the ordering of the Yes and No logits. This analysis therefore isolates changes in confidence calibration from changes in edge classification.

As shown in Appendix Fig.~\ref{fig:temperature_sensitivity_munin1}, Gemma-31B benefits from higher-temperature smoothing: its ECE and Brier score decrease as temperature increases, suggesting that the original Yes/No logits are overconfident. Qwen2.5-72B changes less across temperatures and maintains lower ECE, indicating comparatively stable logit-based confidence. AUROC remains nearly unchanged for both models because temperature scaling preserves the ranking induced by the original logit differences.

These results show that temperature scaling can adjust confidence sharpness but cannot correct cases where the model assigns the higher logit to the wrong causal label. To identify the structural distinctions underlying these persistent errors, we next examine whether overconfident false positives concentrate on indirect and reversed relations.

\subsection{Overconfident False Positives on Structurally Difficult Non-Edges}
\label{subsec:overconfident_false_positives}
The analysis reveals a clear structural pattern: false positives occur more frequently on indirect and reversed relations and often carry high verbal confidence. We analyze valid model-query instances where the reference graph contains no direct edge \(A \rightarrow B\). Non-edges are divided into three mutually exclusive categories: \emph{reversed direct}, where \(B \rightarrow A\) exists; \emph{indirect}, where \(B\) is reachable from \(A\) through a directed path of length at least two; and \emph{other} non-edges. This separation distinguishes orientation errors from cases where a mediated causal relation is incorrectly classified as a direct edge.

\begin{table}[H]
\caption{False-positive behavior by non-edge type. FP Rate is the fraction of valid queries producing false positives; High-Conf. FP Rate is the fraction producing false positives with verbal confidence at least 80.}
\label{tab:error_behavior_nonedge_type}
\centering
\conferencetablestyle
\small
\setlength{\tabcolsep}{4.5pt}
\renewcommand{\arraystretch}{1.9}
\resizebox{\columnwidth}{!}{%
\begin{tabular}{lrrrr}
\toprule
\textbf{Non-edge Type} &
\makecell{\textbf{Valid}\\\textbf{Pairs}} &
\makecell{\textbf{FP}\\\textbf{Count}} &
\makecell{\textbf{FP}\\\textbf{Rate}} &
\makecell{\textbf{High-Conf.}\\\textbf{FP Rate}} \\
\midrule
Reversed direct & 29,559 & 10,644 & 36.0\% & 30.4\% \\
Indirect        & 161,520 & 64,568 & \textbf{40.0\%} & \textbf{32.3\%} \\
Other           & 2,256,303 & 637,205 & 28.2\% & 22.1\% \\
\bottomrule
\end{tabular}%
}
\end{table}

Table~\ref{tab:error_behavior_nonedge_type} shows that models predict a direct edge for 40.0\% of indirect non-edges and 36.0\% of reversed direct non-edges, compared with 28.2\% of other non-edges. High-confidence false positives follow the same pattern: 32.3\% of indirect and 30.4\% of reversed direct queries produce false positives with verbal confidence of at least 80, compared with 22.1\% of other non-edges. Overprediction is therefore especially pronounced when the queried variables are causally related in the reference graph but not through the proposed direct edge. The indirect-versus-other pattern holds across all six datasets, while reversed direct non-edges have a higher false-positive rate than other non-edges on five of six datasets.

\begin{figure}[t]
\centering
\includegraphics[width=\columnwidth,keepaspectratio]{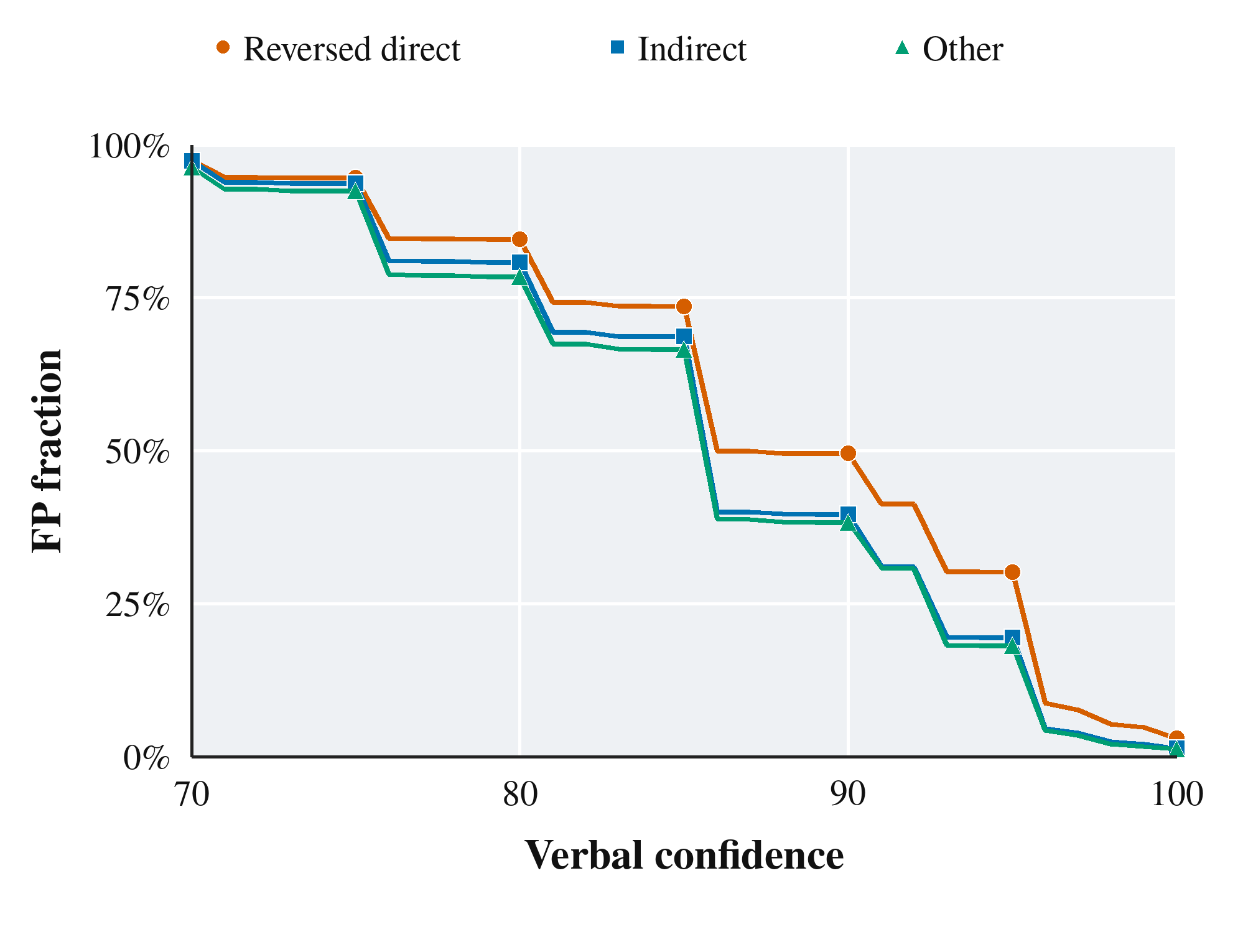}
\caption{Verbal-confidence survival among false positives with parseable scores. Each curve shows the fraction at or above a given threshold; reversed-direct errors remain most concentrated at high confidence.}
\label{fig:false_positive_confidence_survival}
\end{figure}

Figure~\ref{fig:false_positive_confidence_survival} further shows that these errors remain concentrated at high confidence. Among false positives with parseable verbal confidence, 84.6\% of reversed direct, 80.8\% of indirect, and 78.4\% of other false positives fall in the 80--100 confidence range (Appendix Table~\ref{tab:false_positive_confidence_buckets}). Appendix Table~\ref{tab:model_worst_case_high_conf_fp} reports the corresponding worst-case model-level behavior.

These findings should be interpreted relative to the published reference graphs. Because CausalGraphBench provides a single reference structure for each dataset, some absent edges may reflect graph-construction choices, and our analysis cannot establish that every false positive is causally invalid. Nevertheless, the higher false-positive rates on reversed and indirect relations show that the errors are structurally concentrated rather than uniform. Reversed predictions contradict the encoded orientation, while indirect predictions collapse a represented mediated path into a direct edge. Thus, relative to the benchmark specification, LLMs struggle to preserve causal direction and distinguish direct from mediated influence, although some predicted relations absent from the reference graph may remain causally plausible.

Prompting strategy is also associated with this behavior. Appendix Table~\ref{tab:prompt_high_conf_fp} shows that FS+CoT produces the highest high-confidence false-positive rate across all three categories, indicating that reasoning-oriented demonstrations are associated with a greater tendency to infer direct edges between causally related variables. Overall, LLMs appear to capture coarse causal relatedness while frequently misstating directness or orientation with high confidence. This finding motivates the following ablation, which tests whether jointly representing both edge directions and the no-edge class reduces these errors.

\subsection{Three-Way Label-Space Ablation}
\label{subsec:three_way_ablation}

Three-way labeling does not consistently improve graph recovery, indicating that independent binary queries are not the sole cause of edge overprediction. In the original formulation, each unordered variable pair is evaluated through two separate questions: whether \(A \rightarrow B\) exists and whether \(B \rightarrow A\) exists. Because these decisions are made independently, a model can predict both directions or fail to compare a directed edge directly against its reverse.

The ablation replaces these two queries with one three-way decision over \(A \rightarrow B\), \(B \rightarrow A\), and no direct edge. This formulation forces the two directions to compete and prevents reciprocal predictions for the same pair. It also makes directness explicit: if \(A\) influences \(B\) only through an intermediate variable, the correct label is no direct edge. The ablation therefore tests whether jointly deciding edge existence and orientation reduces the indirect and reversed errors identified in Section~\ref{subsec:overconfident_false_positives}.

We evaluate Qwen3-4B-Instruct, Phi-4, Gemma-4-31B-IT, and Llama-3.3-70B-Instruct on \textsc{River Status}, \textsc{COVID}, \textsc{Hepar2}, and \textsc{Munin1} using Metadata prompting. The selected datasets represent the small, medium, large, and very-large categories through River Status, COVID, Hepar2, and Munin1, respectively, while the models represent four families across two scale groups. Binary results are taken from the corresponding Metadata runs in the main experiment. Three-way predictions are converted into directed graphs by adding the selected \(A \rightarrow B\) or \(B \rightarrow A\) edge and adding no edge for the third label. Both formulations are then evaluated using the same edge precision, recall, F1, and nSHD metrics, macro-averaged across the four datasets. All three-way runs cover every unordered pair and produce validly parsed labels. The complete prompt appears in Appendix~\ref{app:three_way_prompt}.

\begin{table}[H]
\caption{Binary versus three-way direct-edge classification under Metadata prompting, macro-averaged across four datasets. Bold indicates the better formulation for each model and metric.}
\label{tab:three_way_ablation_main}
\label{tab:three_way_ablation}
\centering
\footnotesize
\setlength{\tabcolsep}{3.2pt}
\renewcommand{\arraystretch}{1.08}
\resizebox{\columnwidth}{!}{%
\begin{tabular}{llcccc}
\toprule
\textbf{Model} & \textbf{Formulation}
& \textbf{P$\uparrow$}
& \textbf{R$\uparrow$}
& \textbf{F1$\uparrow$}
& \textbf{nSHD$\downarrow$} \\
\midrule
\multirow{2}{*}{Qwen3-4B}
& Binary    & \textbf{0.229} & 0.570 & \textbf{0.261} & \textbf{0.200} \\
& Three-way & 0.124 & \textbf{0.678} & 0.194 & 0.276 \\
\midrule
\multirow{2}{*}{Phi-4}
& Binary    & 0.222 & 0.416 & 0.234 & \textbf{0.132} \\
& Three-way & \textbf{0.224} & \textbf{0.627} & \textbf{0.303} & 0.145 \\
\midrule
\multirow{2}{*}{Gemma-4-31B}
& Binary    & 0.177 & \textbf{0.816} & 0.270 & 0.242 \\
& Three-way & \textbf{0.187} & 0.714 & \textbf{0.275} & \textbf{0.205} \\
\midrule
\multirow{2}{*}{Llama-3.3-70B}
& Binary    & \textbf{0.162} & 0.076 & 0.098 & \textbf{0.068} \\
& Three-way & 0.152 & \textbf{0.795} & \textbf{0.236} & 0.294 \\
\bottomrule
\end{tabular}%
}
\end{table}

Table~\ref{tab:three_way_ablation} shows that three-way labeling improves F1 for Phi-4, Gemma-4-31B-IT, and Llama-3.3-70B-Instruct, but decreases it for Qwen3-4B-Instruct. These F1 changes do not translate into consistent structural improvement. Only Gemma-4-31B-IT improves both F1 and nSHD, and its F1 increase is marginal. For Llama-3.3-70B-Instruct, recall increases from 0.076 to 0.795, but nSHD worsens from 0.068 to 0.294. The model therefore predicts substantially more true edges while adding enough false edges to produce a less accurate reconstructed graph.

The explicit no-edge label also does not eliminate directness errors. We define the three-way non-edge false-positive rate as the fraction of reference no-edge pairs assigned either directed label. On \textsc{Munin1}, this rate ranges from 13.2\% for Phi-4 to 91.0\% for Llama-3.3-70B-Instruct, reaching 62.9\% for Qwen3-4B-Instruct and 45.2\% for Gemma-4-31B-IT. Direction-specific recall is also asymmetric under the fixed unordered-pair ordering, with Qwen3-4B-Instruct and Llama-3.3-70B-Instruct recovering no \(B \rightarrow A\) edges on \textsc{Munin1}. Because the directional classes are imbalanced, this asymmetry indicates sensitivity to pair ordering rather than establishing that one causal direction is intrinsically harder.

Overall, forcing edge existence and orientation into a single mutually exclusive decision does not reliably reduce overprediction or improve orientation recovery. The errors observed in the binary experiment are therefore not solely artifacts of asking about each direction separately; models continue to select directed relations when the reference graph specifies no direct edge. Full dataset-level results are reported in Appendix Table~\ref{tab:three_way_ablation_detailed}.

\section{Conclusion}
\label{sec:concl}

We evaluated whether LLMs provide reliable direct causal-edge judgments and confidence estimates across 12 open-weight models, six benchmark graphs, and five prompting strategies. Models were recall-dominant, producing dense graphs, while prompting shifted the precision--recall tradeoff and gains from model scale diminished on larger graphs. False positives concentrated on indirect and reversed relations and were frequently assigned high confidence, indicating that LLMs often capture causal relatedness without preserving directness or orientation. A mutually exclusive three-way formulation did not consistently improve graph recovery, showing that these errors are not solely artifacts of separate binary queries.

Verbalized and logit-based confidence were unreliable, whereas agreement-based methods achieved better mean calibration and discrimination without establishing statistical superiority across datasets. These patterns persisted on the five graphs not flagged for potential benchmark familiarity. Overall, LLM judgments are better treated as externally validated soft priors than as direct evidence of causal structure.

\paragraph{Future Work.}
Future work should develop graph-aware uncertainty methods that incorporate sparsity, acyclicity, and directional consistency, and evaluate calibrated LLM judgments as soft priors within data-driven CD pipelines. Experimenting with Small Language Models (SLMs) can further assess whether lightweight models offer competitive edge-level reliability at lower computational cost. Robustness should also be tested across paraphrased prompts and broader graph collections.
\section*{Limitations}\label{sec:lim}
Our evaluation covers six of the 35 CausalGraphBench graphs. Although stratified by size and domain, they may not represent the benchmark's full diversity or noisier real-world settings. We treat the published structures as fixed reference graphs, although they may reflect expert choices about causal granularity; alternative annotations and inter-annotator agreement estimates are unavailable. Consequently, not every prediction labeled as a false positive can be established as causally invalid.

Our language-only, pairwise formulation uses neither observational nor interventional data and does not enforce global constraints such as acyclicity. The semantic-accessibility analysis relies on heuristic metadata indicators and is exploratory rather than causal. We compare prompting strategies but do not test robustness to paraphrased wording within each strategy. Results may also depend on the selected open-weight models and decoding settings. Finally, the benchmark-familiarity audit flags five model--dataset pairs involving AsiaM, but its single detection protocol and judge model cannot establish memorization or rule out familiarity with unflagged datasets.


\bibliography{references}

\clearpage
\appendix
\raggedbottom
\twocolumn
\setcounter{table}{0}
\renewcommand{\thetable}{A\arabic{table}}
\setcounter{figure}{0}
\renewcommand{\thefigure}{A\arabic{figure}}

\section{Appendix}

\subsection{Prompt Templates}
\label{app:prompts}

\subsubsection{Name-only Prompt}

\begin{promptbox}
\normalsize
Does A cause B? Answer Yes or No.
Provide your confidence (0-100\%) that A directly causes B.

Context: \texttt{\{context\}}
Variable A: \texttt{\{var\_a\}}
Variable B: \texttt{\{var\_b\}}

Answer (Yes/No, Confidence\%):
\end{promptbox}

\subsubsection{Metadata Prompt}

\begin{promptbox}
\normalsize
Given variable definitions, does A cause B? Answer Yes or No.
Provide your confidence (0-100\%) that A directly causes B.

Context: \texttt{\{context\}}
Variable A: \texttt{\{var\_a\}}
Definition: \texttt{\{def\_a\}}
Variable B: \texttt{\{var\_b\}}
Definition: \texttt{\{def\_b\}}

Answer (Yes/No, Confidence\%):
\end{promptbox}

\subsubsection{Chain-of-Thought Prompt}

\begin{promptbox}
\normalsize
Given variable definitions, does A cause B?
Think step by step before answering.

Context: \texttt{\{context\}}
Variable A: \texttt{\{var\_a\}}
Definition: \texttt{\{def\_a\}}
Variable B: \texttt{\{var\_b\}}
Definition: \texttt{\{def\_b\}}

Answer (Yes/No):
Confidence (0-100\%):
Reasoning:
\end{promptbox}

\subsubsection{Few-shot Prompt}

\begin{promptbox}
\normalsize
Does A cause B? Answer Yes or No.
Provide your confidence (0-100\%) that A directly causes B.

Context: \texttt{\{context\}}

Example 1:
Variable A: Smoking (tobacco intake)
Variable B: Lung Cancer (malignant tumor)
Answer: Yes, 95\%

Example 2:
Variable A: Rain (precipitation)
Variable B: Wet Roads (surface moisture)
Answer: Yes, 98\%

Example 3:
Variable A: Exercise (physical activity)
Variable B: Income (earnings)
Answer: No, 90\%

Example 4:
Variable A: Height (body length)
Variable B: Intelligence (cognitive ability)
Answer: No, 88\%

Example 5:
Variable A: UV Radiation (sun exposure)
Variable B: Skin Cancer (malignant melanoma)
Answer: Yes, 93\%

Now answer:
Variable A: \texttt{\{var\_a\}}
Definition: \texttt{\{def\_a\}}
Variable B: \texttt{\{var\_b\}}
Definition: \texttt{\{def\_b\}}

Answer (Yes/No, Confidence\%):
\end{promptbox}

\subsubsection{Few-shot + Chain-of-Thought Prompt}

\begin{promptbox}
\normalsize
Given variable definitions, does A cause B? Think step by step.
Provide answer, confidence, reasoning.

Context: \texttt{\{context\}}

Example 1:
Variable A: Smoking (tobacco intake)
Variable B: Lung Cancer (malignant tumor)
Answer: Yes, 95\%
Reasoning: Tobacco contains carcinogens that directly damage lung tissue leading to malignant tumor formation.

Example 2:
Variable A: Rain (precipitation)
Variable B: Wet Roads (surface moisture)
Answer: Yes, 98\%
Reasoning: Rain directly causes water to accumulate on road surfaces.

Example 3:
Variable A: Exercise (physical activity)
Variable B: Income (earnings)
Answer: No, 90\%
Reasoning: Physical activity level has no direct causal effect on earnings.

Example 4:
Variable A: Height (body length)
Variable B: Intelligence (cognitive ability)
Answer: No, 88\%
Reasoning: Body height does not causally determine cognitive ability.

Example 5:
Variable A: UV Radiation (sun exposure)
Variable B: Skin Cancer (malignant melanoma)
Answer: Yes, 93\%
Reasoning: UV radiation directly damages DNA in skin cells, triggering mutations that lead to malignant tumor formation.

Now answer:
Variable A: \texttt{\{var\_a\}}
Definition: \texttt{\{def\_a\}}
Variable B: \texttt{\{var\_b\}}
Definition: \texttt{\{def\_b\}}

Answer (Yes/No):
Confidence (0-100\%):
Reasoning:
\end{promptbox}

\subsubsection{Three-Way Ablation Prompt}
\label{app:three_way_prompt}

\begin{promptbox}
\small
\ttfamily
Given the context and variable definitions, determine the direct causal relationship between Variable A and Variable B.

\medskip
Choose exactly one label:

A\_TO\_B: Variable A directly causes Variable B.

B\_TO\_A: Variable B directly causes Variable A.

NO\_EDGE: There is no direct causal edge in either direction.

\medskip
Provide your confidence (0--100\%) in the selected label.

Return only the selected label and confidence. Do not provide reasoning or an explanation.

\medskip
Context: \{context\}

Variable A: \{var\_a\}

Definition: \{def\_a\}

Variable B: \{var\_b\}

Definition: \{def\_b\}

\medskip
Answer (A\_TO\_B/B\_TO\_A/NO\_EDGE, Confidence\%):
\end{promptbox}


\subsection{Confidence Estimation Details}
\label{app:confidence_estimation_details}

Formal definitions are provided in Section~\ref{subsec:confidence_estimation}. Verbalized confidence is extracted from the standardized answer--confidence response after deterministic parsing of labeled fields and answer-first forms. Logit-based confidence aggregates accepted surface forms of \textsc{Yes} and \textsc{No} before renormalization over the binary label set. Cross-prompt agreement uses predictions from the five prompt styles, whereas cross-model agreement is computed separately within the small- and large-model groups under a fixed prompt. Agreement-based predictions use the majority label.

\clearpage
\onecolumn
\subsection{Benchmark Dataset Details}
\label{app:datasets}

\subsubsection{Dataset Statistics}
\label{app:dataset_statistics}

\begin{table}[H]
\caption{Reference-graph statistics. No-edge:edge denotes class imbalance over ordered variable pairs.}
\label{tab:dataset_stats}
\centering
\small
\setlength{\tabcolsep}{4pt}
\renewcommand{\arraystretch}{1.08}
\resizebox{\textwidth}{!}{%
\begin{tabular}{lllrrrr}
\toprule
\textbf{Dataset} & \textbf{Domain} & \textbf{Source} & \textbf{Nodes} & \textbf{Edges} & \textbf{Pairs} & \textbf{No-edge:Edge} \\
\midrule
\multicolumn{7}{l}{\textit{Small networks} ($n<20$)} \\
\addlinespace[1pt]
AsiaM & Respiratory diagnosis & bnlearn; \citep{lauritzen1988asia} & 7 & 8 & 42 & 4.25:1 \\
River Status & Ecological quality & \citep{molina2020river} & 15 & 25 & 210 & 7.40:1 \\
\midrule
\multicolumn{7}{l}{\textit{Medium networks} ($20 \leq n \leq 50$)} \\
\addlinespace[1pt]
COVID & Vaccine risk-benefit & BayesFusion; \citep{mayfield2022covid} & 20 & 26 & 380 & 13.62:1 \\
Coal Gasifier & Industrial risk & \citep{liu2022coal} & 39 & 39 & 1,482 & 37.00:1 \\
\midrule
\multicolumn{7}{l}{\textit{Large networks} ($51 \leq n \leq 100$)} \\
\addlinespace[1pt]
Hepar2 & Hepatic diagnosis & bnlearn; \citep{onisko2003hepar2} & 70 & 123 & 4,830 & 38.27:1 \\
\midrule
\multicolumn{7}{l}{\textit{Very large networks} ($n>100$)} \\
\addlinespace[1pt]
Munin1 & Electromyography & bnlearn; \citep{andreassen1989munin} & 186 & 273 & 34,410 & 125.04:1 \\
\bottomrule
\end{tabular}%
}
\end{table}

\begin{table}[H]
\centering
\conferencetablestyle
\footnotesize
\setlength{\tabcolsep}{7pt}
\begin{tabular}{lrrrr}
\toprule
\textbf{Dataset} &
\makecell{\textbf{Number of}\\\textbf{variables}} &
\makecell{\textbf{Mean words}\\\textbf{per description}} &
\makecell{\textbf{Code-like variable}\\\textbf{names (\%)}} &
\makecell{\textbf{Descriptions with}\\\textbf{acronyms (\%)}} \\
\midrule
AsiaM         &   7 & 5.3 &   0.0 &  0.0 \\
River Status  &  15 & 3.7 &   6.7 &  0.0 \\
COVID         &  20 & 6.8 & 100.0 & 45.0 \\
Coal Gasifier &  39 & 3.9 & 100.0 &  5.1 \\
Hepar2        &  70 & 3.4 &  15.7 &  0.0 \\
Munin1        & 186 & 7.6 & 100.0 & 27.4 \\
\bottomrule
\end{tabular}
\caption{Metadata-level indicators of variable-description complexity across datasets. The table reports simple descriptive measures of the node metadata used in the classification prompts. Lower values for the number of variables, code-like variable names, and acronym-containing descriptions generally indicate easier metadata conditions. Mean words per description is included as a descriptive measure of label length, but is not assumed to have a strictly monotonic relationship with performance because longer descriptions may either clarify or complicate variable meaning.}
\label{tab:metadata_readability_indicators}
\end{table}

\subsubsection{Dataset Preprocessing}
\label{app:dataset_preprocessing}

To avoid leaking structural cues to the LLM, we remove explicit references to ``Bayesian Network'' from the dataset-level context before prompting. Since this term may implicitly suggest that directed edges exist among the variables, we replace it with the neutral term ``system,'' while preserving the original domain meaning of each dataset.

\subsection{Model Classification Performance Ranking}
\label{app:model_classification_performance_ranking}

\begin{figure}[H]
    \centering
    \includegraphics[width=\columnwidth,keepaspectratio]{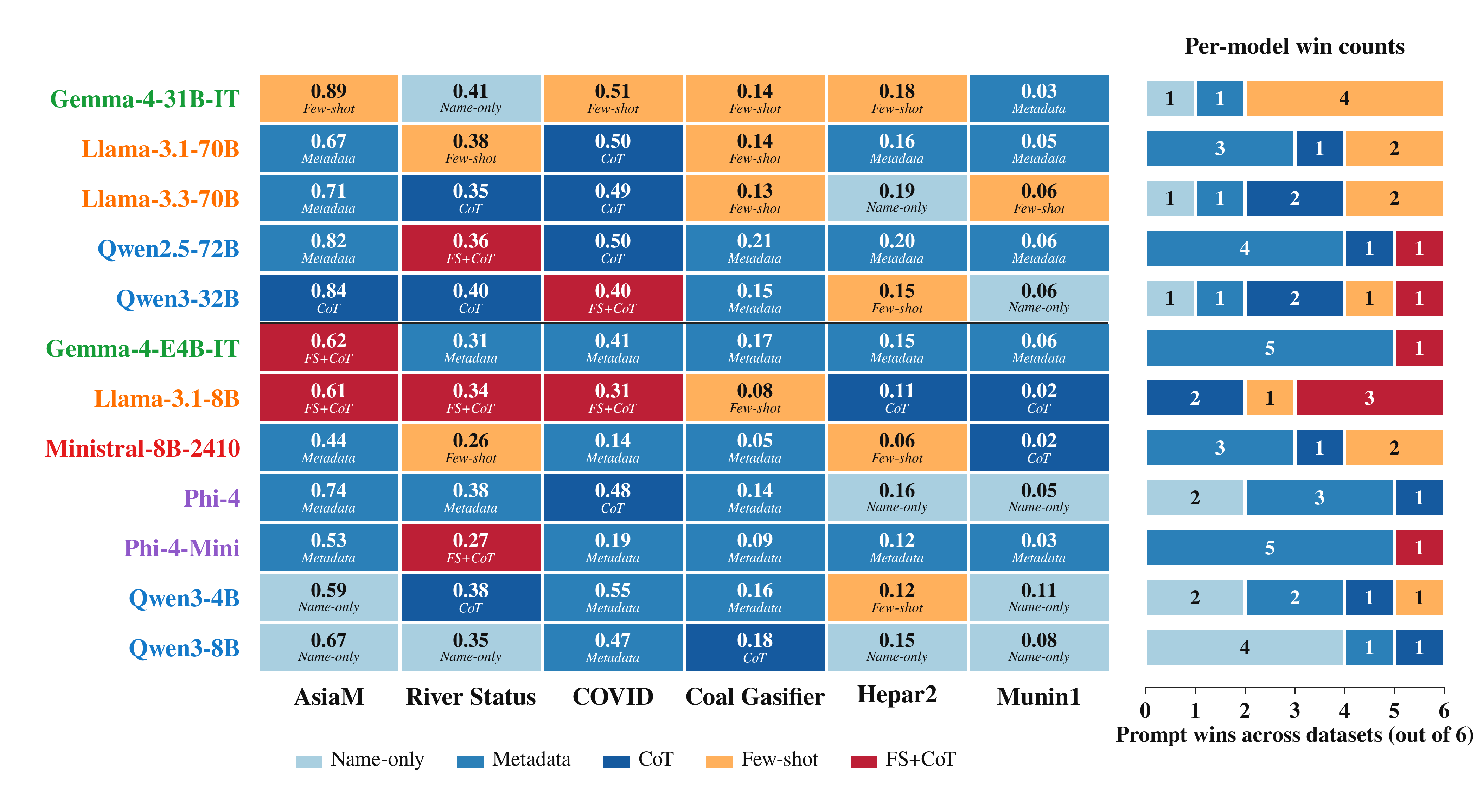}
    \caption{Prompt-level F1 performance across models and datasets. Each cell shows the best F1 score and prompt, with right-side bars summarizing prompt wins per model. The horizontal line separates large and small models.}
    \label{fig:promptwise_f1_all_models}
\end{figure}

\subsection{Complete Results Tables}
\label{app:complete_results_tables}

\subsubsection{Primary Classification Tables}
\label{app:classification_results}

\begingroup
\vfuzz=1000pt
\hfuzz=1000pt
\begin{table}[H]
\centering
\scriptsize
\setlength{\tabcolsep}{2.6pt}
\resizebox{\textwidth}{!}{%
\begin{tabular}{llrrrrrrrrrrrrrrrrrrrr}
\toprule
\multirow{2}{*}{Model} & \multirow{2}{*}{Dataset} & \multicolumn{4}{c}{Name-only} & \multicolumn{4}{c}{Metadata} & \multicolumn{4}{c}{CoT} & \multicolumn{4}{c}{Few-shot} & \multicolumn{4}{c}{Few-shot+CoT} \\
\cmidrule(lr){3-6} \cmidrule(lr){7-10} \cmidrule(lr){11-14} \cmidrule(lr){15-18} \cmidrule(lr){19-22}
 & & P$\uparrow$ & R$\uparrow$ & F1$\uparrow$ & nSHD$\downarrow$ & P$\uparrow$ & R$\uparrow$ & F1$\uparrow$ & nSHD$\downarrow$ & P$\uparrow$ & R$\uparrow$ & F1$\uparrow$ & nSHD$\downarrow$ & P$\uparrow$ & R$\uparrow$ & F1$\uparrow$ & nSHD$\downarrow$ & P$\uparrow$ & R$\uparrow$ & F1$\uparrow$ & nSHD$\downarrow$ \\
\midrule
\multirow{6}{*}{\rotatebox[origin=c]{90}{\parbox{1.8cm}{\centering\texttt{Gemma-4-\\E4B-IT}}}} & \texttt{asiam} & \cellcolor{yellow!55}0.500 & \cellcolor{orange!50}0.375 & \cellcolor{yellow!35}0.429 & \cellcolor{orange!20}0.190 & \cellcolor{yellow!55}0.417 & \cellcolor{orange!20}0.625 & \cellcolor{yellow!55}0.500 & \cellcolor{orange!20}0.214 & \cellcolor{yellow!55}1.000 & \cellcolor{orange!65}0.125 & \cellcolor{orange!20}0.222 & \cellcolor{orange!20}0.167 & \cellcolor{yellow!55}0.400 & \cellcolor{orange!35}0.500 & \cellcolor{yellow!55}0.444 & \cellcolor{orange!20}0.190 & \cellcolor{yellow!55}0.625 & \cellcolor{orange!20}0.625 & \cellcolor{yellow!55}0.625 & \cellcolor{yellow!35}0.143 \\
 & \texttt{river} & \cellcolor{orange!20}0.200 & \cellcolor{orange!50}0.240 & \cellcolor{orange!20}0.218 & \cellcolor{orange!20}0.205 & \cellcolor{yellow!35}0.239 & \cellcolor{orange!35}0.440 & \cellcolor{yellow!35}0.310 & \cellcolor{orange!20}0.233 & \cellcolor{yellow!35}0.212 & \cellcolor{orange!35}0.560 & \cellcolor{yellow!35}0.308 & \cellcolor{orange!35}0.300 & \cellcolor{orange!20}0.167 & \cellcolor{yellow!55}0.920 & \cellcolor{orange!20}0.282 & \cellcolor{orange!65}0.557 & \cellcolor{orange!20}0.169 & \cellcolor{yellow!55}0.880 & \cellcolor{orange!20}0.284 & \cellcolor{orange!65}0.529 \\
 & \texttt{covid} & \cellcolor{orange!35}0.071 & \cellcolor{orange!65}0.038 & \cellcolor{orange!50}0.050 & \cellcolor{yellow!35}0.097 & \cellcolor{yellow!55}0.435 & \cellcolor{orange!35}0.385 & \cellcolor{yellow!35}0.408 & \cellcolor{yellow!55}0.074 & \cellcolor{yellow!55}0.421 & \cellcolor{orange!50}0.308 & \cellcolor{yellow!35}0.356 & \cellcolor{yellow!55}0.076 & \cellcolor{yellow!35}0.202 & \cellcolor{orange!20}0.654 & \cellcolor{yellow!35}0.309 & \cellcolor{orange!20}0.197 & \cellcolor{orange!20}0.162 & \cellcolor{orange!20}0.654 & \cellcolor{orange!20}0.260 & \cellcolor{orange!35}0.255 \\
 & \texttt{coal} & \cellcolor{orange!50}0.042 & \cellcolor{orange!65}0.026 & \cellcolor{orange!65}0.032 & \cellcolor{yellow!55}0.041 & \cellcolor{orange!20}0.125 & \cellcolor{orange!50}0.256 & \cellcolor{orange!20}0.168 & \cellcolor{yellow!55}0.067 & \cellcolor{orange!20}0.096 & \cellcolor{orange!50}0.308 & \cellcolor{orange!20}0.146 & \cellcolor{yellow!55}0.093 & \cellcolor{orange!50}0.038 & \cellcolor{yellow!55}0.846 & \cellcolor{orange!50}0.073 & \cellcolor{orange!65}0.562 & \cellcolor{orange!50}0.043 & \cellcolor{yellow!35}0.718 & \cellcolor{orange!35}0.082 & \cellcolor{orange!50}0.423 \\
 & \texttt{hepar2} & \cellcolor{orange!20}0.096 & \cellcolor{orange!50}0.325 & \cellcolor{orange!20}0.148 & \cellcolor{yellow!55}0.094 & \cellcolor{orange!20}0.093 & \cellcolor{orange!35}0.415 & \cellcolor{orange!20}0.152 & \cellcolor{yellow!35}0.115 & \cellcolor{orange!20}0.096 & \cellcolor{orange!65}0.179 & \cellcolor{orange!35}0.125 & \cellcolor{yellow!55}0.062 & \cellcolor{orange!35}0.046 & \cellcolor{orange!20}0.626 & \cellcolor{orange!35}0.086 & \cellcolor{orange!35}0.335 & \cellcolor{orange!35}0.049 & \cellcolor{orange!20}0.561 & \cellcolor{orange!35}0.091 & \cellcolor{orange!35}0.281 \\
 & \texttt{munin1} & \cellcolor{orange!65}0.014 & \cellcolor{orange!65}0.051 & \cellcolor{orange!65}0.022 & \cellcolor{yellow!55}0.037 & \cellcolor{orange!50}0.033 & \cellcolor{orange!65}0.194 & \cellcolor{orange!50}0.056 & \cellcolor{yellow!55}0.051 & \cellcolor{orange!65}0.018 & \cellcolor{orange!50}0.242 & \cellcolor{orange!50}0.034 & \cellcolor{yellow!35}0.109 & \cellcolor{orange!65}0.010 & \cellcolor{yellow!55}0.850 & \cellcolor{orange!65}0.019 & \cellcolor{orange!65}0.687 & \cellcolor{orange!65}0.009 & \cellcolor{yellow!55}0.802 & \cellcolor{orange!65}0.018 & \cellcolor{orange!65}0.675 \\
\specialrule{0.10em}{0.10em}{0.10em}
\multirow{6}{*}{\rotatebox[origin=c]{90}{\parbox{1.8cm}{\centering\texttt{Llama-3.1-8B-\\Instruct}}}} & \texttt{asiam} & \cellcolor{yellow!35}0.333 & \cellcolor{yellow!55}0.875 & \cellcolor{yellow!55}0.483 & \cellcolor{orange!35}0.357 & \cellcolor{yellow!55}0.375 & \cellcolor{yellow!35}0.750 & \cellcolor{yellow!55}0.500 & \cellcolor{orange!35}0.262 & \cellcolor{yellow!55}0.556 & \cellcolor{orange!20}0.625 & \cellcolor{yellow!55}0.588 & \cellcolor{yellow!35}0.143 & \cellcolor{yellow!35}0.304 & \cellcolor{yellow!55}0.875 & \cellcolor{yellow!55}0.452 & \cellcolor{orange!50}0.405 & \cellcolor{yellow!55}0.467 & \cellcolor{yellow!55}0.875 & \cellcolor{yellow!55}0.609 & \cellcolor{orange!20}0.214 \\
 & \texttt{river} & \cellcolor{orange!20}0.172 & \cellcolor{yellow!35}0.680 & \cellcolor{orange!20}0.274 & \cellcolor{orange!50}0.414 & \cellcolor{orange!20}0.192 & \cellcolor{orange!35}0.560 & \cellcolor{orange!20}0.286 & \cellcolor{orange!35}0.310 & \cellcolor{yellow!35}0.245 & \cellcolor{orange!35}0.480 & \cellcolor{yellow!35}0.324 & \cellcolor{orange!20}0.224 & \cellcolor{orange!20}0.153 & \cellcolor{yellow!55}0.840 & \cellcolor{orange!20}0.259 & \cellcolor{orange!65}0.567 & \cellcolor{yellow!35}0.222 & \cellcolor{yellow!35}0.720 & \cellcolor{yellow!35}0.340 & \cellcolor{orange!35}0.329 \\
 & \texttt{covid} & \cellcolor{orange!20}0.136 & \cellcolor{orange!35}0.462 & \cellcolor{orange!20}0.211 & \cellcolor{orange!20}0.226 & \cellcolor{orange!20}0.200 & \cellcolor{orange!35}0.462 & \cellcolor{orange!20}0.279 & \cellcolor{yellow!35}0.158 & \cellcolor{orange!20}0.095 & \cellcolor{orange!65}0.077 & \cellcolor{orange!35}0.085 & \cellcolor{yellow!35}0.111 & \cellcolor{orange!20}0.134 & \cellcolor{yellow!35}0.692 & \cellcolor{orange!20}0.225 & \cellcolor{orange!35}0.324 & \cellcolor{orange!20}0.198 & \cellcolor{yellow!35}0.692 & \cellcolor{yellow!35}0.308 & \cellcolor{orange!20}0.203 \\
 & \texttt{coal} & \cellcolor{orange!50}0.029 & \cellcolor{orange!35}0.513 & \cellcolor{orange!50}0.055 & \cellcolor{orange!50}0.457 & \cellcolor{orange!50}0.029 & \cellcolor{orange!50}0.359 & \cellcolor{orange!50}0.053 & \cellcolor{orange!35}0.329 & \cellcolor{orange!65}0.018 & \cellcolor{orange!65}0.154 & \cellcolor{orange!65}0.033 & \cellcolor{orange!20}0.233 & \cellcolor{orange!50}0.042 & \cellcolor{orange!20}0.667 & \cellcolor{orange!50}0.078 & \cellcolor{orange!50}0.411 & \cellcolor{orange!50}0.037 & \cellcolor{orange!35}0.410 & \cellcolor{orange!50}0.069 & \cellcolor{orange!35}0.287 \\
 & \texttt{hepar2} & \cellcolor{orange!50}0.044 & \cellcolor{orange!20}0.561 & \cellcolor{orange!35}0.081 & \cellcolor{orange!35}0.320 & \cellcolor{orange!35}0.049 & \cellcolor{orange!35}0.504 & \cellcolor{orange!35}0.090 & \cellcolor{orange!35}0.257 & \cellcolor{orange!35}0.068 & \cellcolor{orange!50}0.366 & \cellcolor{orange!35}0.115 & \cellcolor{yellow!35}0.140 & \cellcolor{orange!35}0.046 & \cellcolor{yellow!35}0.740 & \cellcolor{orange!35}0.087 & \cellcolor{orange!50}0.391 & \cellcolor{orange!35}0.060 & \cellcolor{yellow!35}0.683 & \cellcolor{orange!35}0.110 & \cellcolor{orange!35}0.278 \\
 & \texttt{munin1} & \cellcolor{orange!65}0.009 & \cellcolor{orange!20}0.571 & \cellcolor{orange!65}0.018 & \cellcolor{orange!50}0.488 & \cellcolor{orange!65}0.012 & \cellcolor{orange!35}0.509 & \cellcolor{orange!65}0.023 & \cellcolor{orange!35}0.337 & \cellcolor{orange!65}0.013 & \cellcolor{orange!35}0.462 & \cellcolor{orange!65}0.025 & \cellcolor{orange!35}0.288 & \cellcolor{orange!65}0.010 & \cellcolor{yellow!55}0.835 & \cellcolor{orange!65}0.020 & \cellcolor{orange!65}0.649 & \cellcolor{orange!65}0.010 & \cellcolor{orange!20}0.615 & \cellcolor{orange!65}0.019 & \cellcolor{orange!50}0.506 \\
\specialrule{0.10em}{0.10em}{0.10em}
\multirow{6}{*}{\rotatebox[origin=c]{90}{\parbox{1.8cm}{\centering\texttt{Ministral-8B-\\Instruct-2410}}}} & \texttt{asiam} & \cellcolor{orange!20}0.182 & \cellcolor{yellow!35}0.750 & \cellcolor{orange!20}0.293 & \cellcolor{orange!65}0.667 & \cellcolor{yellow!35}0.292 & \cellcolor{yellow!55}0.875 & \cellcolor{yellow!55}0.438 & \cellcolor{orange!50}0.405 & \cellcolor{orange!20}0.171 & \cellcolor{yellow!35}0.750 & \cellcolor{orange!20}0.279 & \cellcolor{orange!65}0.690 & \cellcolor{yellow!35}0.214 & \cellcolor{yellow!35}0.750 & \cellcolor{yellow!35}0.333 & \cellcolor{orange!50}0.524 & \cellcolor{yellow!35}0.241 & \cellcolor{yellow!55}0.875 & \cellcolor{yellow!35}0.378 & \cellcolor{orange!50}0.524 \\
 & \texttt{river} & \cellcolor{orange!20}0.128 & \cellcolor{orange!20}0.640 & \cellcolor{orange!20}0.213 & \cellcolor{orange!65}0.529 & \cellcolor{orange!20}0.152 & \cellcolor{yellow!35}0.680 & \cellcolor{orange!20}0.248 & \cellcolor{orange!50}0.467 & \cellcolor{orange!20}0.113 & \cellcolor{orange!20}0.640 & \cellcolor{orange!20}0.193 & \cellcolor{orange!65}0.614 & \cellcolor{orange!20}0.154 & \cellcolor{yellow!55}0.800 & \cellcolor{orange!20}0.258 & \cellcolor{orange!65}0.533 & \cellcolor{orange!20}0.124 & \cellcolor{yellow!35}0.760 & \cellcolor{orange!20}0.213 & \cellcolor{orange!65}0.643 \\
 & \texttt{covid} & \cellcolor{orange!35}0.059 & \cellcolor{orange!35}0.500 & \cellcolor{orange!35}0.106 & \cellcolor{orange!65}0.555 & \cellcolor{orange!35}0.079 & \cellcolor{orange!20}0.577 & \cellcolor{orange!35}0.139 & \cellcolor{orange!50}0.476 & \cellcolor{orange!35}0.053 & \cellcolor{orange!35}0.400 & \cellcolor{orange!35}0.093 & \cellcolor{orange!50}0.495 & \cellcolor{orange!35}0.057 & \cellcolor{orange!35}0.500 & \cellcolor{orange!35}0.102 & \cellcolor{orange!65}0.582 & \cellcolor{orange!35}0.058 & \cellcolor{orange!35}0.500 & \cellcolor{orange!35}0.104 & \cellcolor{orange!65}0.568 \\
 & \texttt{coal} & \cellcolor{orange!50}0.024 & \cellcolor{orange!20}0.564 & \cellcolor{orange!50}0.047 & \cellcolor{orange!65}0.594 & \cellcolor{orange!50}0.027 & \cellcolor{orange!20}0.564 & \cellcolor{orange!50}0.052 & \cellcolor{orange!65}0.539 & \cellcolor{orange!50}0.026 & \cellcolor{orange!20}0.615 & \cellcolor{orange!50}0.051 & \cellcolor{orange!65}0.601 & \cellcolor{orange!50}0.024 & \cellcolor{orange!20}0.564 & \cellcolor{orange!50}0.046 & \cellcolor{orange!65}0.611 & \cellcolor{orange!50}0.024 & \cellcolor{orange!20}0.641 & \cellcolor{orange!50}0.047 & \cellcolor{orange!65}0.681 \\
 & \texttt{hepar2} & \cellcolor{orange!50}0.026 & \cellcolor{orange!20}0.602 & \cellcolor{orange!50}0.049 & \cellcolor{orange!65}0.587 & \cellcolor{orange!50}0.026 & \cellcolor{orange!35}0.553 & \cellcolor{orange!50}0.050 & \cellcolor{orange!65}0.529 & \cellcolor{orange!50}0.027 & \cellcolor{orange!20}0.577 & \cellcolor{orange!50}0.052 & \cellcolor{orange!65}0.532 & \cellcolor{orange!50}0.034 & \cellcolor{orange!20}0.642 & \cellcolor{orange!50}0.064 & \cellcolor{orange!50}0.470 & \cellcolor{orange!50}0.030 & \cellcolor{orange!20}0.610 & \cellcolor{orange!50}0.058 & \cellcolor{orange!50}0.499 \\
 & \texttt{munin1} & \cellcolor{orange!65}0.008 & \cellcolor{orange!20}0.575 & \cellcolor{orange!65}0.015 & \cellcolor{orange!65}0.590 & \cellcolor{orange!65}0.008 & \cellcolor{orange!20}0.601 & \cellcolor{orange!65}0.017 & \cellcolor{orange!65}0.559 & \cellcolor{orange!65}0.009 & \cellcolor{orange!20}0.637 & \cellcolor{orange!65}0.018 & \cellcolor{orange!65}0.539 & \cellcolor{orange!65}0.008 & \cellcolor{orange!35}0.560 & \cellcolor{orange!65}0.016 & \cellcolor{orange!65}0.559 & \cellcolor{orange!65}0.009 & \cellcolor{orange!20}0.641 & \cellcolor{orange!65}0.017 & \cellcolor{orange!65}0.571 \\
\specialrule{0.10em}{0.10em}{0.10em}
\multirow{6}{*}{\rotatebox[origin=c]{90}{\texttt{Phi-4}}} & \texttt{asiam} & \cellcolor{yellow!55}0.417 & \cellcolor{orange!20}0.625 & \cellcolor{yellow!55}0.500 & \cellcolor{orange!20}0.214 & \cellcolor{yellow!55}0.636 & \cellcolor{yellow!55}0.875 & \cellcolor{yellow!55}0.737 & \cellcolor{yellow!35}0.119 & \cellcolor{yellow!55}0.500 & \cellcolor{yellow!55}0.875 & \cellcolor{yellow!55}0.636 & \cellcolor{orange!20}0.190 & \cellcolor{yellow!55}0.421 & \cellcolor{yellow!55}1.000 & \cellcolor{yellow!55}0.593 & \cellcolor{orange!35}0.262 & \cellcolor{yellow!55}0.375 & \cellcolor{yellow!35}0.750 & \cellcolor{yellow!55}0.500 & \cellcolor{orange!35}0.262 \\
 & \texttt{river} & \cellcolor{yellow!35}0.238 & \cellcolor{orange!35}0.400 & \cellcolor{yellow!35}0.299 & \cellcolor{orange!20}0.224 & \cellcolor{yellow!35}0.333 & \cellcolor{orange!35}0.440 & \cellcolor{yellow!35}0.379 & \cellcolor{orange!20}0.171 & \cellcolor{yellow!35}0.210 & \cellcolor{yellow!35}0.680 & \cellcolor{yellow!35}0.321 & \cellcolor{orange!35}0.343 & \cellcolor{yellow!35}0.220 & \cellcolor{orange!35}0.520 & \cellcolor{yellow!35}0.310 & \cellcolor{orange!35}0.276 & \cellcolor{orange!20}0.179 & \cellcolor{yellow!35}0.680 & \cellcolor{orange!20}0.283 & \cellcolor{orange!50}0.405 \\
 & \texttt{covid} & \cellcolor{yellow!55}0.450 & \cellcolor{orange!50}0.346 & \cellcolor{yellow!35}0.391 & \cellcolor{yellow!55}0.071 & \cellcolor{yellow!55}0.444 & \cellcolor{orange!50}0.308 & \cellcolor{yellow!35}0.364 & \cellcolor{yellow!55}0.074 & \cellcolor{yellow!55}0.464 & \cellcolor{orange!35}0.500 & \cellcolor{yellow!55}0.481 & \cellcolor{yellow!55}0.071 & \cellcolor{yellow!35}0.296 & \cellcolor{orange!20}0.615 & \cellcolor{yellow!35}0.400 & \cellcolor{yellow!35}0.126 & \cellcolor{orange!20}0.200 & \cellcolor{orange!35}0.560 & \cellcolor{yellow!35}0.295 & \cellcolor{orange!20}0.176 \\
 & \texttt{coal} & \cellcolor{orange!65}0.000 & \cellcolor{orange!65}0.000 & \cellcolor{orange!65}0.000 & \cellcolor{yellow!55}0.074 & \cellcolor{orange!35}0.080 & \cellcolor{orange!20}0.667 & \cellcolor{orange!20}0.143 & \cellcolor{orange!20}0.210 & \cellcolor{orange!35}0.051 & \cellcolor{yellow!35}0.744 & \cellcolor{orange!35}0.096 & \cellcolor{orange!50}0.364 & \cellcolor{orange!35}0.050 & \cellcolor{orange!20}0.641 & \cellcolor{orange!35}0.093 & \cellcolor{orange!35}0.327 & \cellcolor{orange!50}0.041 & \cellcolor{yellow!35}0.744 & \cellcolor{orange!50}0.079 & \cellcolor{orange!50}0.455 \\
 & \texttt{hepar2} & \cellcolor{orange!20}0.095 & \cellcolor{orange!35}0.537 & \cellcolor{orange!20}0.161 & \cellcolor{yellow!35}0.141 & \cellcolor{orange!35}0.088 & \cellcolor{orange!20}0.593 & \cellcolor{orange!20}0.153 & \cellcolor{orange!20}0.165 & \cellcolor{orange!35}0.077 & \cellcolor{yellow!35}0.772 & \cellcolor{orange!35}0.141 & \cellcolor{orange!35}0.239 & \cellcolor{orange!35}0.057 & \cellcolor{yellow!35}0.732 & \cellcolor{orange!35}0.106 & \cellcolor{orange!35}0.313 & \cellcolor{orange!35}0.055 & \cellcolor{yellow!35}0.777 & \cellcolor{orange!35}0.103 & \cellcolor{orange!35}0.337 \\
 & \texttt{munin1} & \cellcolor{orange!50}0.029 & \cellcolor{orange!65}0.110 & \cellcolor{orange!50}0.047 & \cellcolor{yellow!55}0.035 & \cellcolor{orange!50}0.022 & \cellcolor{orange!50}0.322 & \cellcolor{orange!50}0.041 & \cellcolor{yellow!35}0.119 & \cellcolor{orange!65}0.018 & \cellcolor{orange!35}0.480 & \cellcolor{orange!50}0.034 & \cellcolor{orange!20}0.214 & \cellcolor{orange!65}0.015 & \cellcolor{orange!35}0.527 & \cellcolor{orange!65}0.028 & \cellcolor{orange!35}0.287 & \cellcolor{orange!65}0.011 & \cellcolor{yellow!35}0.681 & \cellcolor{orange!65}0.021 & \cellcolor{orange!50}0.495 \\
\specialrule{0.10em}{0.10em}{0.10em}
\multirow{6}{*}{\rotatebox[origin=c]{90}{\parbox{1.8cm}{\centering\texttt{Phi-4-Mini-\\Instruct}}}} & \texttt{asiam} & \cellcolor{yellow!55}0.375 & \cellcolor{orange!50}0.375 & \cellcolor{yellow!35}0.375 & \cellcolor{orange!20}0.190 & \cellcolor{yellow!55}0.571 & \cellcolor{orange!35}0.500 & \cellcolor{yellow!55}0.533 & \cellcolor{orange!20}0.167 & \cellcolor{yellow!55}0.500 & \cellcolor{orange!50}0.250 & \cellcolor{yellow!35}0.333 & \cellcolor{orange!20}0.167 & \cellcolor{orange!65}0.000 & \cellcolor{orange!65}0.000 & \cellcolor{orange!65}0.000 & \cellcolor{orange!35}0.262 & \cellcolor{yellow!35}0.333 & \cellcolor{orange!50}0.250 & \cellcolor{orange!20}0.286 & \cellcolor{orange!20}0.167 \\
 & \texttt{river} & \cellcolor{orange!20}0.200 & \cellcolor{orange!65}0.208 & \cellcolor{orange!20}0.204 & \cellcolor{orange!20}0.176 & \cellcolor{yellow!35}0.227 & \cellcolor{orange!65}0.208 & \cellcolor{orange!20}0.217 & \cellcolor{orange!20}0.171 & \cellcolor{orange!20}0.175 & \cellcolor{orange!50}0.280 & \cellcolor{orange!20}0.215 & \cellcolor{orange!20}0.229 & \cellcolor{orange!20}0.152 & \cellcolor{orange!35}0.400 & \cellcolor{orange!20}0.220 & \cellcolor{orange!35}0.329 & \cellcolor{yellow!35}0.208 & \cellcolor{orange!35}0.400 & \cellcolor{orange!20}0.274 & \cellcolor{orange!35}0.252 \\
 & \texttt{covid} & \cellcolor{orange!20}0.158 & \cellcolor{orange!65}0.115 & \cellcolor{orange!35}0.133 & \cellcolor{yellow!35}0.097 & \cellcolor{orange!20}0.192 & \cellcolor{orange!65}0.192 & \cellcolor{orange!20}0.192 & \cellcolor{yellow!35}0.108 & \cellcolor{orange!20}0.154 & \cellcolor{orange!65}0.154 & \cellcolor{orange!20}0.154 & \cellcolor{yellow!35}0.108 & \cellcolor{orange!20}0.105 & \cellcolor{orange!65}0.160 & \cellcolor{orange!35}0.127 & \cellcolor{yellow!35}0.139 & \cellcolor{orange!35}0.080 & \cellcolor{orange!65}0.077 & \cellcolor{orange!50}0.078 & \cellcolor{yellow!35}0.121 \\
 & \texttt{coal} & \cellcolor{orange!35}0.055 & \cellcolor{orange!65}0.105 & \cellcolor{orange!50}0.072 & \cellcolor{yellow!55}0.067 & \cellcolor{orange!35}0.064 & \cellcolor{orange!65}0.179 & \cellcolor{orange!35}0.094 & \cellcolor{yellow!55}0.089 & \cellcolor{orange!35}0.060 & \cellcolor{orange!65}0.205 & \cellcolor{orange!35}0.092 & \cellcolor{yellow!35}0.103 & \cellcolor{orange!50}0.045 & \cellcolor{orange!50}0.282 & \cellcolor{orange!50}0.078 & \cellcolor{orange!20}0.171 & \cellcolor{orange!50}0.044 & \cellcolor{orange!50}0.231 & \cellcolor{orange!50}0.074 & \cellcolor{yellow!35}0.148 \\
 & \texttt{hepar2} & \cellcolor{orange!35}0.071 & \cellcolor{orange!65}0.228 & \cellcolor{orange!35}0.109 & \cellcolor{yellow!55}0.092 & \cellcolor{orange!35}0.084 & \cellcolor{orange!65}0.203 & \cellcolor{orange!35}0.119 & \cellcolor{yellow!55}0.074 & \cellcolor{orange!35}0.075 & \cellcolor{orange!50}0.260 & \cellcolor{orange!35}0.117 & \cellcolor{yellow!35}0.098 & \cellcolor{orange!35}0.063 & \cellcolor{orange!50}0.369 & \cellcolor{orange!35}0.107 & \cellcolor{yellow!35}0.152 & \cellcolor{orange!35}0.065 & \cellcolor{orange!50}0.382 & \cellcolor{orange!35}0.111 & \cellcolor{yellow!35}0.152 \\
 & \texttt{munin1} & \cellcolor{orange!65}0.012 & \cellcolor{orange!65}0.185 & \cellcolor{orange!65}0.022 & \cellcolor{yellow!35}0.129 & \cellcolor{orange!65}0.016 & \cellcolor{orange!50}0.275 & \cellcolor{orange!65}0.031 & \cellcolor{yellow!35}0.137 & \cellcolor{orange!65}0.011 & \cellcolor{orange!65}0.198 & \cellcolor{orange!65}0.021 & \cellcolor{yellow!35}0.142 & \cellcolor{orange!65}0.013 & \cellcolor{orange!50}0.292 & \cellcolor{orange!65}0.024 & \cellcolor{orange!20}0.182 & \cellcolor{orange!65}0.013 & \cellcolor{orange!50}0.260 & \cellcolor{orange!65}0.024 & \cellcolor{orange!20}0.164 \\
\specialrule{0.10em}{0.10em}{0.10em}
\multirow{6}{*}{\rotatebox[origin=c]{90}{\parbox{1.8cm}{\centering\texttt{Qwen3-4B-\\Instruct}}}} & \texttt{asiam} & \cellcolor{yellow!55}0.421 & \cellcolor{yellow!55}1.000 & \cellcolor{yellow!55}0.593 & \cellcolor{orange!35}0.262 & \cellcolor{yellow!55}0.438 & \cellcolor{yellow!55}0.875 & \cellcolor{yellow!55}0.583 & \cellcolor{orange!35}0.238 & \cellcolor{yellow!35}0.333 & \cellcolor{yellow!55}0.875 & \cellcolor{yellow!55}0.483 & \cellcolor{orange!35}0.357 & \cellcolor{yellow!55}0.381 & \cellcolor{yellow!55}1.000 & \cellcolor{yellow!55}0.552 & \cellcolor{orange!35}0.310 & \cellcolor{yellow!35}0.320 & \cellcolor{yellow!55}1.000 & \cellcolor{yellow!55}0.485 & \cellcolor{orange!50}0.405 \\
 & \texttt{river} & \cellcolor{yellow!35}0.222 & \cellcolor{orange!20}0.640 & \cellcolor{yellow!35}0.330 & \cellcolor{orange!35}0.310 & \cellcolor{yellow!35}0.213 & \cellcolor{yellow!35}0.760 & \cellcolor{yellow!35}0.333 & \cellcolor{orange!50}0.362 & \cellcolor{yellow!35}0.244 & \cellcolor{yellow!55}0.833 & \cellcolor{yellow!35}0.377 & \cellcolor{orange!35}0.319 & \cellcolor{yellow!35}0.207 & \cellcolor{yellow!35}0.760 & \cellcolor{yellow!35}0.325 & \cellcolor{orange!50}0.376 & \cellcolor{yellow!35}0.211 & \cellcolor{yellow!55}0.920 & \cellcolor{yellow!35}0.343 & \cellcolor{orange!50}0.414 \\
 & \texttt{covid} & \cellcolor{yellow!55}0.615 & \cellcolor{orange!50}0.333 & \cellcolor{yellow!55}0.432 & \cellcolor{yellow!55}0.061 & \cellcolor{yellow!55}0.619 & \cellcolor{orange!35}0.500 & \cellcolor{yellow!55}0.553 & \cellcolor{yellow!55}0.053 & \cellcolor{yellow!55}0.436 & \cellcolor{orange!20}0.654 & \cellcolor{yellow!55}0.523 & \cellcolor{yellow!55}0.082 & \cellcolor{yellow!35}0.321 & \cellcolor{yellow!35}0.692 & \cellcolor{yellow!55}0.439 & \cellcolor{yellow!35}0.121 & \cellcolor{orange!20}0.198 & \cellcolor{yellow!35}0.769 & \cellcolor{yellow!35}0.315 & \cellcolor{orange!20}0.229 \\
 & \texttt{coal} & \cellcolor{orange!65}0.000 & \cellcolor{orange!65}0.000 & \cellcolor{orange!65}0.000 & \cellcolor{yellow!55}0.028 & \cellcolor{orange!20}0.097 & \cellcolor{orange!35}0.474 & \cellcolor{orange!20}0.161 & \cellcolor{yellow!35}0.126 & \cellcolor{orange!35}0.067 & \cellcolor{orange!35}0.513 & \cellcolor{orange!35}0.119 & \cellcolor{orange!20}0.199 & \cellcolor{orange!35}0.060 & \cellcolor{orange!35}0.410 & \cellcolor{orange!35}0.105 & \cellcolor{orange!20}0.184 & \cellcolor{orange!50}0.043 & \cellcolor{orange!20}0.667 & \cellcolor{orange!50}0.080 & \cellcolor{orange!50}0.402 \\
 & \texttt{hepar2} & \cellcolor{orange!35}0.064 & \cellcolor{orange!35}0.516 & \cellcolor{orange!35}0.114 & \cellcolor{orange!20}0.202 & \cellcolor{orange!35}0.058 & \cellcolor{orange!20}0.642 & \cellcolor{orange!35}0.107 & \cellcolor{orange!35}0.271 & \cellcolor{orange!35}0.054 & \cellcolor{yellow!35}0.733 & \cellcolor{orange!35}0.101 & \cellcolor{orange!35}0.324 & \cellcolor{orange!35}0.066 & \cellcolor{orange!20}0.650 & \cellcolor{orange!35}0.120 & \cellcolor{orange!35}0.241 & \cellcolor{orange!35}0.052 & \cellcolor{yellow!35}0.724 & \cellcolor{orange!35}0.097 & \cellcolor{orange!35}0.340 \\
 & \texttt{munin1} & \cellcolor{orange!35}0.089 & \cellcolor{orange!65}0.136 & \cellcolor{orange!35}0.107 & \cellcolor{yellow!55}0.017 & \cellcolor{orange!50}0.027 & \cellcolor{orange!50}0.377 & \cellcolor{orange!50}0.050 & \cellcolor{yellow!35}0.113 & \cellcolor{orange!50}0.020 & \cellcolor{orange!35}0.509 & \cellcolor{orange!50}0.038 & \cellcolor{orange!20}0.206 & \cellcolor{orange!65}0.017 & \cellcolor{yellow!35}0.692 & \cellcolor{orange!65}0.032 & \cellcolor{orange!35}0.329 & \cellcolor{orange!65}0.011 & \cellcolor{yellow!55}0.832 & \cellcolor{orange!65}0.021 & \cellcolor{orange!65}0.609 \\
\specialrule{0.10em}{0.10em}{0.10em}
\multirow{6}{*}{\rotatebox[origin=c]{90}{\parbox{1.8cm}{\centering\texttt{Qwen3-8B-\\Instruct}}}} & \texttt{asiam} & \cellcolor{yellow!55}0.714 & \cellcolor{orange!20}0.625 & \cellcolor{yellow!55}0.667 & \cellcolor{yellow!35}0.119 & \cellcolor{yellow!55}0.538 & \cellcolor{yellow!55}0.875 & \cellcolor{yellow!55}0.667 & \cellcolor{orange!20}0.167 & \cellcolor{yellow!55}0.467 & \cellcolor{yellow!55}0.875 & \cellcolor{yellow!55}0.609 & \cellcolor{orange!20}0.214 & \cellcolor{yellow!35}0.333 & \cellcolor{yellow!55}1.000 & \cellcolor{yellow!55}0.500 & \cellcolor{orange!50}0.381 & \cellcolor{yellow!55}0.389 & \cellcolor{yellow!55}0.875 & \cellcolor{yellow!55}0.538 & \cellcolor{orange!35}0.286 \\
 & \texttt{river} & \cellcolor{yellow!35}0.333 & \cellcolor{orange!50}0.360 & \cellcolor{yellow!35}0.346 & \cellcolor{orange!20}0.162 & \cellcolor{orange!20}0.200 & \cellcolor{orange!50}0.320 & \cellcolor{orange!20}0.246 & \cellcolor{orange!20}0.233 & \cellcolor{orange!20}0.183 & \cellcolor{orange!35}0.440 & \cellcolor{orange!20}0.259 & \cellcolor{orange!35}0.300 & \cellcolor{orange!20}0.162 & \cellcolor{yellow!55}0.840 & \cellcolor{orange!20}0.271 & \cellcolor{orange!50}0.519 & \cellcolor{orange!20}0.168 & \cellcolor{yellow!55}0.800 & \cellcolor{orange!20}0.278 & \cellcolor{orange!50}0.476 \\
 & \texttt{covid} & \cellcolor{yellow!55}0.500 & \cellcolor{orange!50}0.231 & \cellcolor{yellow!35}0.316 & \cellcolor{yellow!55}0.068 & \cellcolor{yellow!55}0.524 & \cellcolor{orange!35}0.423 & \cellcolor{yellow!55}0.468 & \cellcolor{yellow!55}0.066 & \cellcolor{yellow!55}0.600 & \cellcolor{orange!50}0.346 & \cellcolor{yellow!55}0.439 & \cellcolor{yellow!55}0.061 & \cellcolor{orange!20}0.194 & \cellcolor{yellow!35}0.769 & \cellcolor{yellow!35}0.310 & \cellcolor{orange!35}0.234 & \cellcolor{yellow!35}0.212 & \cellcolor{yellow!55}0.846 & \cellcolor{yellow!35}0.338 & \cellcolor{orange!20}0.226 \\
 & \texttt{coal} & -- & \cellcolor{orange!65}0.000 & \cellcolor{orange!65}0.000 & \cellcolor{yellow!55}0.026 & \cellcolor{orange!35}0.082 & \cellcolor{orange!50}0.231 & \cellcolor{orange!35}0.121 & \cellcolor{yellow!55}0.088 & \cellcolor{orange!20}0.119 & \cellcolor{orange!35}0.385 & \cellcolor{orange!20}0.182 & \cellcolor{yellow!55}0.091 & \cellcolor{orange!50}0.031 & \cellcolor{orange!20}0.667 & \cellcolor{orange!50}0.060 & \cellcolor{orange!65}0.549 & \cellcolor{orange!50}0.035 & \cellcolor{orange!20}0.615 & \cellcolor{orange!50}0.067 & \cellcolor{orange!50}0.454 \\
 & \texttt{hepar2} & \cellcolor{orange!35}0.087 & \cellcolor{orange!35}0.472 & \cellcolor{orange!20}0.147 & \cellcolor{yellow!35}0.138 & \cellcolor{orange!35}0.084 & \cellcolor{orange!35}0.553 & \cellcolor{orange!20}0.145 & \cellcolor{orange!20}0.164 & \cellcolor{orange!35}0.071 & \cellcolor{orange!35}0.480 & \cellcolor{orange!35}0.124 & \cellcolor{orange!20}0.170 & \cellcolor{orange!35}0.048 & \cellcolor{yellow!35}0.780 & \cellcolor{orange!35}0.090 & \cellcolor{orange!50}0.400 & \cellcolor{orange!35}0.051 & \cellcolor{yellow!35}0.772 & \cellcolor{orange!35}0.095 & \cellcolor{orange!50}0.372 \\
 & \texttt{munin1} & \cellcolor{orange!35}0.064 & \cellcolor{orange!65}0.117 & \cellcolor{orange!35}0.083 & \cellcolor{yellow!55}0.020 & \cellcolor{orange!50}0.026 & \cellcolor{orange!65}0.190 & \cellcolor{orange!50}0.046 & \cellcolor{yellow!55}0.061 & \cellcolor{orange!65}0.017 & \cellcolor{orange!65}0.205 & \cellcolor{orange!65}0.031 & \cellcolor{yellow!35}0.101 & \cellcolor{orange!65}0.011 & \cellcolor{yellow!55}0.941 & \cellcolor{orange!65}0.021 & \cellcolor{orange!65}0.701 & \cellcolor{orange!65}0.010 & \cellcolor{yellow!55}0.916 & \cellcolor{orange!65}0.020 & \cellcolor{orange!65}0.714 \\
\bottomrule
\end{tabular}%
}
\caption{Primary causal edge classification performance for small language models across six benchmark datasets and five prompt styles. Cell colors encode performance percentile within the full table (yellow~=~good, orange~=~bad), with intensity scaled per metric direction (P/R/F1 higher is better; nSHD lower is better). A dash denotes undefined precision because the model made no positive predictions.}
\label{tab:primary_classification_full}
\end{table}
\endgroup

\begin{table}[H]
\centering
\small
\setlength{\tabcolsep}{4pt}
\renewcommand{\arraystretch}{1.25}

\begin{minipage}[t]{0.49\textwidth}\vspace{0pt}
\centering
\resizebox{\linewidth}{!}{%
\begin{tabular}{llcccc}
\toprule
\multicolumn{6}{c}{\textbf{Small LLMs}} \\
\midrule
\textbf{Model} & \textbf{Prompt} & \textbf{P} $\uparrow$ & \textbf{R} $\uparrow$ & \textbf{F1} $\uparrow$ & \textbf{Avg. nSHD} $\downarrow$ \\
\midrule
\multirow{5}{*}{\rotatebox[origin=c]{90}{\parbox{1.8cm}{\centering Gemma-4-\\E4B-IT}}} & Name-only & 0.154 & 0.176 & 0.150 & \textbf{0.111} \\
& Metadata & 0.224 & 0.386 & \textbf{0.266} & 0.126 \\
& CoT & \textbf{0.307} & 0.287 & 0.199 & 0.135 \\
& Few-shot & 0.144 & \textbf{0.733} & 0.202 & 0.421 \\
& Few-shot + CoT & 0.176 & 0.707 & 0.227 & 0.384 \\
\midrule
\multirow{5}{*}{\rotatebox[origin=c]{90}{\parbox{1.8cm}{\centering Llama-3.1-8B-\\Instruct}}} & Name-only & 0.121 & 0.610 & 0.187 & 0.377 \\
& Metadata & 0.143 & 0.524 & 0.205 & 0.275 \\
& CoT & \textbf{0.166} & 0.361 & 0.195 & \textbf{0.190} \\
& Few-shot & 0.115 & \textbf{0.775} & 0.187 & 0.458 \\
& Few-shot + CoT & 0.166 & 0.666 & \textbf{0.243} & 0.303 \\
\midrule
\multirow{5}{*}{\rotatebox[origin=c]{90}{\parbox{1.8cm}{\centering Ministral-8B-\\Instruct-2410}}} & Name-only & 0.071 & 0.605 & 0.121 & 0.587 \\
& Metadata & \textbf{0.097} & 0.642 & \textbf{0.157} & \textbf{0.496} \\
& CoT & 0.067 & 0.603 & 0.114 & 0.579 \\
& Few-shot & 0.082 & 0.636 & 0.136 & 0.547 \\
& Few-shot + CoT & 0.081 & \textbf{0.671} & 0.136 & 0.581 \\
\midrule
\multirow{5}{*}{\rotatebox[origin=c]{90}{Phi-4}} & Name-only & 0.205 & 0.336 & 0.233 & \textbf{0.127} \\
& Metadata & \textbf{0.267} & 0.534 & \textbf{0.303} & 0.143 \\
& CoT & 0.220 & 0.675 & 0.285 & 0.237 \\
& Few-shot & 0.176 & 0.672 & 0.255 & 0.265 \\
& Few-shot + CoT & 0.144 & \textbf{0.699} & 0.213 & 0.355 \\
\midrule
\multirow{5}{*}{\rotatebox[origin=c]{90}{\parbox{1.8cm}{\centering Phi-4-Mini-\\Instruct}}} & Name-only & 0.145 & 0.203 & 0.152 & 0.125 \\
& Metadata & \textbf{0.192} & 0.260 & \textbf{0.198} & \textbf{0.124} \\
& CoT & 0.163 & 0.225 & 0.155 & 0.141 \\
& Few-shot & 0.063 & 0.251 & 0.093 & 0.206 \\
& Few-shot + CoT & 0.124 & \textbf{0.267} & 0.141 & 0.167 \\
\midrule
\multirow{5}{*}{\rotatebox[origin=c]{90}{\parbox{1.8cm}{\centering Qwen3-4B-\\Instruct}}} & Name-only & 0.235 & 0.438 & 0.263 & \textbf{0.147} \\
& Metadata & \textbf{0.242} & 0.605 & \textbf{0.298} & 0.194 \\
& CoT & 0.192 & 0.686 & 0.274 & 0.248 \\
& Few-shot & 0.175 & 0.701 & 0.262 & 0.260 \\
& Few-shot + CoT & 0.139 & \textbf{0.819} & 0.224 & 0.400 \\
\midrule
\multirow{5}{*}{\rotatebox[origin=c]{90}{\parbox{1.8cm}{\centering Qwen3-8B-\\Instruct}}} & Name-only & \textbf{0.283} & 0.301 & 0.260 & \textbf{0.089} \\
& Metadata & 0.242 & 0.432 & \textbf{0.282} & 0.130 \\
& CoT & 0.243 & 0.455 & 0.274 & 0.156 \\
& Few-shot & 0.130 & \textbf{0.833} & 0.209 & 0.464 \\
& Few-shot + CoT & 0.144 & 0.804 & 0.223 & 0.421 \\
\bottomrule
\end{tabular}%
}
\end{minipage}\hfill
\begin{minipage}[t]{0.49\textwidth}\vspace{0pt}
\centering
\resizebox{\linewidth}{!}{%
\begin{tabular}{llcccc}
\toprule
\multicolumn{6}{c}{\textbf{Large LLMs}} \\
\midrule
\textbf{Model} & \textbf{Prompt} & \textbf{P} $\uparrow$ & \textbf{R} $\uparrow$ & \textbf{F1} $\uparrow$ & \textbf{Avg. nSHD} $\downarrow$ \\
\midrule
\multirow{5}{*}{\rotatebox[origin=c]{90}{\parbox{1.8cm}{\centering Gemma-4-\\31B-IT}}} & Name-only & 0.236 & 0.524 & 0.278 & \textbf{0.150} \\
& Metadata & 0.241 & \textbf{0.841} & 0.335 & 0.222 \\
& CoT & 0.243 & 0.816 & 0.330 & 0.238 \\
& Few-shot & \textbf{0.276} & 0.738 & \textbf{0.355} & 0.184 \\
& Few-shot + CoT & 0.234 & 0.840 & 0.326 & 0.255 \\
\midrule
\multirow{5}{*}{\rotatebox[origin=c]{90}{\parbox{1.8cm}{\centering Llama-3.1-70B-\\Instruct}}} & Name-only & 0.183 & 0.273 & 0.201 & 0.128 \\
& Metadata & \textbf{0.247} & 0.318 & 0.251 & \textbf{0.096} \\
& CoT & 0.184 & \textbf{0.869} & \textbf{0.272} & 0.372 \\
& Few-shot & 0.176 & 0.695 & 0.261 & 0.243 \\
& Few-shot + CoT & 0.178 & 0.830 & 0.255 & 0.401 \\
\midrule
\multirow{5}{*}{\rotatebox[origin=c]{90}{\parbox{1.8cm}{\centering Llama-3.3-70B-\\Instruct}}} & Name-only & 0.291 & 0.181 & 0.211 & 0.074 \\
& Metadata & \textbf{0.295} & 0.163 & 0.199 & \textbf{0.066} \\
& CoT & 0.204 & \textbf{0.844} & 0.293 & 0.320 \\
& Few-shot & 0.206 & 0.650 & \textbf{0.298} & 0.178 \\
& Few-shot + CoT & 0.168 & 0.842 & 0.251 & 0.403 \\
\midrule
\multirow{5}{*}{\rotatebox[origin=c]{90}{\parbox{1.8cm}{\centering Qwen2.5-72B-\\Instruct}}} & Name-only & \textbf{0.387} & 0.178 & 0.204 & \textbf{0.076} \\
& Metadata & 0.328 & 0.380 & 0.304 & 0.084 \\
& CoT & 0.267 & 0.684 & \textbf{0.320} & 0.198 \\
& Few-shot & 0.213 & 0.658 & 0.285 & 0.213 \\
& Few-shot + CoT & 0.214 & \textbf{0.805} & 0.297 & 0.293 \\
\midrule
\multirow{5}{*}{\rotatebox[origin=c]{90}{\parbox{1.8cm}{\centering Qwen3-32B-\\Instruct}}} & Name-only & 0.207 & 0.344 & 0.208 & 0.144 \\
& Metadata & 0.265 & 0.535 & 0.279 & 0.172 \\
& CoT & \textbf{0.278} & 0.545 & \textbf{0.314} & \textbf{0.143} \\
& Few-shot & 0.212 & 0.578 & 0.276 & 0.196 \\
& Few-shot + CoT & 0.196 & \textbf{0.655} & 0.275 & 0.241 \\
\bottomrule
\end{tabular}%
}
\end{minipage}

\caption{Macro-averaged primary causal edge classification performance for small (left) and large (right) LLMs across all datasets. Precision (P), recall (R), and F1 are macro-averaged over the six benchmark datasets for each model--prompt pair. Avg. nSHD denotes the macro-average of normalized structural Hamming distance, computed for each dataset as $\mathrm{SHD}/(n_d(n_d-1))$, where $n_d$ is the number of variables in dataset $d$. Higher P, R, and F1 are better, while lower Avg. nSHD is better. Bold values indicate the best prompt for each model and metric.}
\label{tab:macro_classification_combined}
\end{table}

\begin{table}[H]
\centering
\conferencetablestyle
\footnotesize
\setlength{\tabcolsep}{3pt}
\resizebox{\textwidth}{!}{%
\begin{tabular}{llcccclcccc}
\toprule
 & \multicolumn{5}{c}{\textbf{Small LLMs}} & \multicolumn{5}{c}{\textbf{Large LLMs}} \\
\cmidrule(lr){2-6} \cmidrule(lr){7-11}
\textbf{Dataset} & \textbf{Model} & \textbf{Avg. P} $\uparrow$ & \textbf{Avg. R} $\uparrow$ & \textbf{Avg. F1} $\uparrow$ & \textbf{Avg. nSHD} $\downarrow$ & \textbf{Model} & \textbf{Avg. P} $\uparrow$ & \textbf{Avg. R} $\uparrow$ & \textbf{Avg. F1} $\uparrow$ & \textbf{Avg. nSHD} $\downarrow$ \\
\midrule
\multirow{7}{*}{\rotatebox[origin=c]{90}{\texttt{asiam}}}
 & Gemma-4-E4B-IT & \textbf{0.588} & 0.450 & 0.444 & \textbf{0.181} & Qwen3-32B-Instruct & 0.603 & 0.825 & 0.694 & 0.124 \\
 & Llama-3.1-8B-Instruct & 0.407 & 0.800 & 0.526 & 0.276 & Gemma-4-31B-IT & \textbf{0.705} & \textbf{0.900} & \textbf{0.780} & \textbf{0.090} \\
 & Ministral-8B-Instruct-2410 & 0.220 & 0.800 & 0.344 & 0.562 & Qwen2.5-72B-Instruct & 0.680 & 0.850 & 0.733 & 0.114 \\
 & Phi-4 & 0.470 & 0.825 & 0.593 & 0.210 & Llama-3.3-70B-Instruct & 0.612 & 0.800 & 0.638 & 0.167 \\
 & Phi-4-Mini-Instruct & 0.356 & 0.275 & 0.305 & 0.190 & Llama-3.1-70B-Instruct & 0.438 & 0.775 & 0.552 & 0.210 \\
 & Qwen3-4B-Instruct & 0.379 & \textbf{0.950} & 0.539 & 0.314 & & & & & \\
 & Qwen3-8B-Instruct & 0.488 & 0.850 & \textbf{0.596} & 0.233 & & & & & \\
\midrule
\multirow{7}{*}{\rotatebox[origin=c]{90}{\texttt{river}}}
 & Gemma-4-E4B-IT & 0.197 & 0.608 & 0.280 & 0.365 & Qwen3-32B-Instruct & 0.223 & 0.616 & 0.326 & 0.303 \\
 & Llama-3.1-8B-Instruct & 0.197 & 0.656 & 0.297 & 0.369 & Gemma-4-31B-IT & \textbf{0.258} & \textbf{0.810} & \textbf{0.391} & 0.296 \\
 & Ministral-8B-Instruct-2410 & 0.134 & 0.704 & 0.225 & 0.557 & Qwen2.5-72B-Instruct & 0.160 & 0.456 & 0.220 & \textbf{0.271} \\
 & Phi-4 & \textbf{0.236} & 0.544 & 0.318 & 0.284 & Llama-3.3-70B-Instruct & 0.228 & 0.496 & 0.261 & 0.272 \\
 & Phi-4-Mini-Instruct & 0.192 & 0.299 & 0.226 & \textbf{0.231} & Llama-3.1-70B-Instruct & 0.213 & 0.608 & 0.295 & 0.315 \\
 & Qwen3-4B-Instruct & 0.219 & \textbf{0.783} & \textbf{0.342} & 0.356 & & & & & \\
 & Qwen3-8B-Instruct & 0.209 & 0.552 & 0.280 & 0.338 & & & & & \\
\midrule
\multirow{7}{*}{\rotatebox[origin=c]{90}{\texttt{covid}}}
 & Gemma-4-E4B-IT & 0.258 & 0.408 & 0.276 & 0.140 & Qwen3-32B-Instruct & 0.418 & 0.362 & 0.341 & 0.089 \\
 & Llama-3.1-8B-Instruct & 0.153 & 0.477 & 0.221 & 0.204 & Gemma-4-31B-IT & 0.352 & \textbf{0.800} & \textbf{0.486} & 0.116 \\
 & Ministral-8B-Instruct-2410 & 0.061 & 0.495 & 0.109 & 0.535 & Qwen2.5-72B-Instruct & \textbf{0.581} & 0.438 & 0.424 & \textbf{0.086} \\
 & Phi-4 & 0.371 & 0.466 & 0.386 & \textbf{0.104} & Llama-3.3-70B-Instruct & 0.359 & 0.531 & 0.370 & 0.108 \\
 & Phi-4-Mini-Instruct & 0.138 & 0.140 & 0.137 & 0.115 & Llama-3.1-70B-Instruct & 0.352 & 0.592 & 0.394 & 0.122 \\
 & Qwen3-4B-Instruct & \textbf{0.438} & \textbf{0.590} & \textbf{0.453} & 0.109 & & & & & \\
 & Qwen3-8B-Instruct & 0.406 & 0.523 & 0.374 & 0.131 & & & & & \\
\midrule
\multirow{7}{*}{\rotatebox[origin=c]{90}{\texttt{coal}}}
 & Gemma-4-E4B-IT & \textbf{0.069} & 0.431 & \textbf{0.100} & 0.237 & Qwen3-32B-Instruct & 0.052 & 0.501 & 0.093 & 0.215 \\
 & Llama-3.1-8B-Instruct & 0.031 & 0.421 & 0.058 & 0.344 & Gemma-4-31B-IT & 0.052 & \textbf{0.608} & 0.096 & 0.241 \\
 & Ministral-8B-Instruct-2410 & 0.025 & \textbf{0.590} & 0.048 & 0.605 & Qwen2.5-72B-Instruct & \textbf{0.091} & 0.513 & \textbf{0.121} & \textbf{0.178} \\
 & Phi-4 & 0.045 & 0.559 & 0.082 & 0.286 & Llama-3.3-70B-Instruct & 0.088 & 0.436 & 0.073 & 0.252 \\
 & Phi-4-Mini-Instruct & 0.053 & 0.201 & 0.082 & \textbf{0.116} & Llama-3.1-70B-Instruct & 0.069 & 0.467 & 0.093 & 0.241 \\
 & Qwen3-4B-Instruct & 0.053 & 0.413 & 0.093 & 0.188 & & & & & \\
 & Qwen3-8B-Instruct & 0.067 & 0.379 & 0.086 & 0.242 & & & & & \\
\midrule
\multirow{7}{*}{\rotatebox[origin=c]{90}{\texttt{hepar2}}}
 & Gemma-4-E4B-IT & \textbf{0.076} & 0.421 & 0.120 & 0.177 & Qwen3-32B-Instruct & 0.074 & 0.481 & 0.128 & 0.164 \\
 & Llama-3.1-8B-Instruct & 0.053 & 0.571 & 0.097 & 0.277 & Gemma-4-31B-IT & 0.093 & \textbf{0.769} & \textbf{0.165} & 0.196 \\
 & Ministral-8B-Instruct-2410 & 0.029 & 0.597 & 0.055 & 0.523 & Qwen2.5-72B-Instruct & \textbf{0.102} & 0.541 & 0.162 & \textbf{0.151} \\
 & Phi-4 & 0.074 & \textbf{0.682} & \textbf{0.133} & 0.239 & Llama-3.3-70B-Instruct & 0.091 & 0.553 & 0.136 & 0.194 \\
 & Phi-4-Mini-Instruct & 0.072 & 0.288 & 0.113 & \textbf{0.113} & Llama-3.1-70B-Instruct & 0.072 & 0.620 & 0.123 & 0.249 \\
 & Qwen3-4B-Instruct & 0.059 & 0.653 & 0.108 & 0.275 & & & & & \\
 & Qwen3-8B-Instruct & 0.068 & 0.611 & 0.120 & 0.249 & & & & & \\
\midrule
\multirow{7}{*}{\rotatebox[origin=c]{90}{\texttt{munin1}}}
 & Gemma-4-E4B-IT & 0.017 & 0.428 & 0.030 & 0.312 & Qwen3-32B-Instruct & \textbf{0.021} & 0.405 & \textbf{0.039} & \textbf{0.180} \\
 & Llama-3.1-8B-Instruct & 0.011 & 0.599 & 0.021 & 0.453 & Gemma-4-31B-IT & 0.015 & \textbf{0.625} & 0.030 & 0.319 \\
 & Ministral-8B-Instruct-2410 & 0.008 & \textbf{0.603} & 0.017 & 0.564 & Qwen2.5-72B-Instruct & 0.017 & 0.447 & 0.032 & 0.238 \\
 & Phi-4 & 0.019 & 0.424 & 0.034 & 0.230 & Llama-3.3-70B-Instruct & 0.019 & 0.401 & 0.024 & 0.257 \\
 & Phi-4-Mini-Instruct & 0.013 & 0.242 & 0.024 & \textbf{0.151} & Llama-3.1-70B-Instruct & 0.017 & 0.519 & 0.030 & 0.352 \\
 & Qwen3-4B-Instruct & \textbf{0.032} & 0.509 & \textbf{0.050} & 0.255 & & & & & \\
 & Qwen3-8B-Instruct & 0.026 & 0.474 & 0.040 & 0.320 & & & & & \\
\bottomrule
\end{tabular}%
}
\caption{Dataset-wise prompt-averaged primary causal edge classification performance for small and large LLMs on the six benchmark datasets. For each model--dataset pair, precision (P), recall (R), F1, and normalized structural Hamming distance (nSHD) are averaged across the five prompt styles. nSHD is computed as $\mathrm{SHD}/(n_d(n_d-1))$, where $n_d$ is the number of variables in dataset $d$. Higher P, R, and F1 are better, while lower Avg. nSHD is better. Bold values indicate the best model within each dataset and metric, with small and large LLMs evaluated independently.}
\label{tab:dataset_avg_classification_combined}
\end{table}

\begingroup
\vfuzz=1000pt
\hfuzz=1000pt
\begin{table}[H]
\centering
\scriptsize
\setlength{\tabcolsep}{2.2pt}
\renewcommand{\arraystretch}{1.05}
\resizebox{\textwidth}{!}{%
\begin{tabular}{llrrrrrrrrrrrrrrrrrrrr}
\toprule
\multirow{2}{*}{Model} & \multirow{2}{*}{Dataset} & \multicolumn{4}{c}{Name-only} & \multicolumn{4}{c}{Metadata} & \multicolumn{4}{c}{CoT} & \multicolumn{4}{c}{Few-shot} & \multicolumn{4}{c}{Few-shot+CoT} \\
\cmidrule(lr){3-6} \cmidrule(lr){7-10} \cmidrule(lr){11-14} \cmidrule(lr){15-18} \cmidrule(lr){19-22}
 & & P$\uparrow$ & R$\uparrow$ & F1$\uparrow$ & nSHD$\downarrow$ & P$\uparrow$ & R$\uparrow$ & F1$\uparrow$ & nSHD$\downarrow$ & P$\uparrow$ & R$\uparrow$ & F1$\uparrow$ & nSHD$\downarrow$ & P$\uparrow$ & R$\uparrow$ & F1$\uparrow$ & nSHD$\downarrow$ & P$\uparrow$ & R$\uparrow$ & F1$\uparrow$ & nSHD$\downarrow$ \\
\midrule
\multirow{6}{*}{\rotatebox[origin=c]{90}{\parbox{1.8cm}{\centering\texttt{Qwen3-32B-\\Instruct}}}} & \texttt{asiam} & \cellcolor{yellow!35}0.500 & \cellcolor{orange!35}0.625 & \cellcolor{yellow!55}0.556 & \cellcolor{orange!35}0.190 & \cellcolor{yellow!55}0.667 & \cellcolor{orange!20}0.750 & \cellcolor{yellow!55}0.706 & \cellcolor{yellow!35}0.095 & \cellcolor{yellow!55}0.727 & \cellcolor{yellow!55}1.000 & \cellcolor{yellow!55}0.842 & \cellcolor{yellow!35}0.071 & \cellcolor{yellow!55}0.583 & \cellcolor{yellow!55}0.875 & \cellcolor{yellow!55}0.700 & \cellcolor{orange!20}0.119 & \cellcolor{yellow!55}0.538 & \cellcolor{yellow!55}0.875 & \cellcolor{yellow!55}0.667 & \cellcolor{orange!20}0.143 \\
 & \texttt{river} & \cellcolor{orange!20}0.203 & \cellcolor{orange!35}0.560 & \cellcolor{orange!20}0.298 & \cellcolor{orange!50}0.314 & \cellcolor{orange!20}0.217 & \cellcolor{orange!35}0.520 & \cellcolor{orange!20}0.306 & \cellcolor{orange!50}0.281 & \cellcolor{orange!20}0.273 & \cellcolor{orange!20}0.720 & \cellcolor{yellow!35}0.396 & \cellcolor{orange!35}0.262 & \cellcolor{orange!20}0.209 & \cellcolor{orange!35}0.560 & \cellcolor{orange!20}0.304 & \cellcolor{orange!50}0.305 & \cellcolor{orange!20}0.212 & \cellcolor{orange!20}0.720 & \cellcolor{orange!20}0.327 & \cellcolor{orange!50}0.352 \\
 & \texttt{covid} & \cellcolor{yellow!35}0.444 & \cellcolor{orange!65}0.154 & \cellcolor{orange!20}0.229 & \cellcolor{yellow!35}0.071 & \cellcolor{yellow!55}0.538 & \cellcolor{orange!50}0.269 & \cellcolor{orange!20}0.359 & \cellcolor{yellow!35}0.066 & \cellcolor{yellow!35}0.500 & \cellcolor{orange!50}0.269 & \cellcolor{orange!20}0.350 & \cellcolor{yellow!35}0.068 & \cellcolor{yellow!35}0.324 & \cellcolor{orange!50}0.423 & \cellcolor{orange!20}0.367 & \cellcolor{yellow!35}0.100 & \cellcolor{orange!20}0.281 & \cellcolor{orange!20}0.692 & \cellcolor{yellow!35}0.400 & \cellcolor{orange!20}0.139 \\
 & \texttt{coal} & \cellcolor{orange!65}0.000 & \cellcolor{orange!65}0.000 & \cellcolor{orange!65}0.000 & \cellcolor{yellow!55}0.028 & \cellcolor{orange!35}0.082 & \cellcolor{orange!35}0.684 & \cellcolor{orange!35}0.146 & \cellcolor{orange!35}0.206 & \cellcolor{orange!35}0.076 & \cellcolor{orange!35}0.513 & \cellcolor{orange!35}0.132 & \cellcolor{orange!35}0.176 & \cellcolor{orange!50}0.055 & \cellcolor{orange!35}0.667 & \cellcolor{orange!50}0.101 & \cellcolor{orange!50}0.310 & \cellcolor{orange!50}0.046 & \cellcolor{orange!35}0.641 & \cellcolor{orange!50}0.087 & \cellcolor{orange!65}0.355 \\
 & \texttt{hepar2} & \cellcolor{orange!50}0.060 & \cellcolor{orange!35}0.472 & \cellcolor{orange!35}0.107 & \cellcolor{orange!35}0.196 & \cellcolor{orange!35}0.067 & \cellcolor{orange!35}0.463 & \cellcolor{orange!35}0.117 & \cellcolor{orange!35}0.173 & \cellcolor{orange!35}0.072 & \cellcolor{orange!35}0.463 & \cellcolor{orange!35}0.124 & \cellcolor{orange!20}0.161 & \cellcolor{orange!35}0.086 & \cellcolor{orange!35}0.528 & \cellcolor{orange!35}0.148 & \cellcolor{orange!20}0.150 & \cellcolor{orange!35}0.086 & \cellcolor{orange!35}0.480 & \cellcolor{orange!35}0.146 & \cellcolor{orange!20}0.139 \\
 & \texttt{munin1} & \cellcolor{orange!50}0.032 & \cellcolor{orange!50}0.256 & \cellcolor{orange!50}0.057 & \cellcolor{yellow!35}0.067 & \cellcolor{orange!50}0.020 & \cellcolor{orange!35}0.524 & \cellcolor{orange!50}0.038 & \cellcolor{orange!35}0.209 & \cellcolor{orange!50}0.021 & \cellcolor{orange!50}0.304 & \cellcolor{orange!50}0.039 & \cellcolor{orange!20}0.117 & \cellcolor{orange!65}0.017 & \cellcolor{orange!50}0.418 & \cellcolor{orange!50}0.034 & \cellcolor{orange!35}0.189 & \cellcolor{orange!65}0.013 & \cellcolor{orange!35}0.524 & \cellcolor{orange!65}0.025 & \cellcolor{orange!50}0.319 \\
\specialrule{0.10em}{0.10em}{0.10em}
\multirow{6}{*}{\rotatebox[origin=c]{90}{\parbox{1.8cm}{\centering\texttt{Gemma-4-\\31B-IT}}}} & \texttt{asiam} & \cellcolor{yellow!55}0.667 & \cellcolor{orange!35}0.500 & \cellcolor{yellow!55}0.571 & \cellcolor{orange!20}0.143 & \cellcolor{yellow!55}0.667 & \cellcolor{yellow!55}1.000 & \cellcolor{yellow!55}0.800 & \cellcolor{yellow!35}0.095 & \cellcolor{yellow!55}0.727 & \cellcolor{yellow!55}1.000 & \cellcolor{yellow!55}0.842 & \cellcolor{yellow!35}0.071 & \cellcolor{yellow!55}0.800 & \cellcolor{yellow!55}1.000 & \cellcolor{yellow!55}0.889 & \cellcolor{yellow!55}0.048 & \cellcolor{yellow!55}0.667 & \cellcolor{yellow!55}1.000 & \cellcolor{yellow!55}0.800 & \cellcolor{yellow!35}0.095 \\
 & \texttt{river} & \cellcolor{orange!20}0.279 & \cellcolor{orange!20}0.773 & \cellcolor{yellow!35}0.410 & \cellcolor{orange!35}0.248 & \cellcolor{orange!20}0.266 & \cellcolor{yellow!55}0.875 & \cellcolor{yellow!35}0.408 & \cellcolor{orange!50}0.295 & \cellcolor{orange!20}0.247 & \cellcolor{yellow!35}0.840 & \cellcolor{yellow!35}0.382 & \cellcolor{orange!50}0.324 & \cellcolor{orange!20}0.257 & \cellcolor{orange!20}0.720 & \cellcolor{yellow!35}0.379 & \cellcolor{orange!50}0.281 & \cellcolor{orange!20}0.241 & \cellcolor{yellow!35}0.840 & \cellcolor{orange!20}0.375 & \cellcolor{orange!50}0.333 \\
 & \texttt{covid} & \cellcolor{yellow!35}0.364 & \cellcolor{orange!20}0.769 & \cellcolor{yellow!35}0.494 & \cellcolor{yellow!35}0.108 & \cellcolor{yellow!35}0.333 & \cellcolor{yellow!35}0.808 & \cellcolor{yellow!35}0.472 & \cellcolor{orange!20}0.124 & \cellcolor{yellow!35}0.318 & \cellcolor{yellow!35}0.808 & \cellcolor{yellow!35}0.457 & \cellcolor{orange!20}0.132 & \cellcolor{yellow!35}0.400 & \cellcolor{orange!20}0.692 & \cellcolor{yellow!35}0.507 & \cellcolor{yellow!35}0.092 & \cellcolor{yellow!35}0.343 & \cellcolor{yellow!55}0.923 & \cellcolor{yellow!35}0.500 & \cellcolor{orange!20}0.126 \\
 & \texttt{coal} & \cellcolor{orange!65}0.000 & \cellcolor{orange!65}0.000 & \cellcolor{orange!65}0.000 & \cellcolor{yellow!55}0.033 & \cellcolor{orange!35}0.069 & \cellcolor{yellow!35}0.784 & \cellcolor{orange!35}0.127 & \cellcolor{orange!50}0.267 & \cellcolor{orange!50}0.059 & \cellcolor{orange!20}0.769 & \cellcolor{orange!35}0.110 & \cellcolor{orange!50}0.326 & \cellcolor{orange!35}0.077 & \cellcolor{orange!20}0.692 & \cellcolor{orange!35}0.138 & \cellcolor{orange!35}0.225 & \cellcolor{orange!50}0.056 & \cellcolor{yellow!35}0.795 & \cellcolor{orange!35}0.105 & \cellcolor{orange!65}0.355 \\
 & \texttt{hepar2} & \cellcolor{orange!35}0.092 & \cellcolor{orange!20}0.746 & \cellcolor{orange!35}0.165 & \cellcolor{orange!35}0.190 & \cellcolor{orange!35}0.093 & \cellcolor{yellow!35}0.798 & \cellcolor{orange!35}0.167 & \cellcolor{orange!35}0.194 & \cellcolor{orange!35}0.091 & \cellcolor{yellow!35}0.813 & \cellcolor{orange!35}0.164 & \cellcolor{orange!35}0.210 & \cellcolor{orange!35}0.105 & \cellcolor{orange!20}0.699 & \cellcolor{orange!20}0.183 & \cellcolor{orange!20}0.157 & \cellcolor{orange!35}0.082 & \cellcolor{yellow!35}0.789 & \cellcolor{orange!35}0.148 & \cellcolor{orange!35}0.229 \\
 & \texttt{munin1} & \cellcolor{orange!65}0.016 & \cellcolor{orange!50}0.359 & \cellcolor{orange!65}0.030 & \cellcolor{orange!35}0.181 & \cellcolor{orange!65}0.017 & \cellcolor{yellow!35}0.783 & \cellcolor{orange!65}0.033 & \cellcolor{orange!65}0.356 & \cellcolor{orange!65}0.014 & \cellcolor{orange!35}0.667 & \cellcolor{orange!65}0.028 & \cellcolor{orange!65}0.367 & \cellcolor{orange!65}0.016 & \cellcolor{orange!35}0.626 & \cellcolor{orange!65}0.032 & \cellcolor{orange!50}0.299 & \cellcolor{orange!65}0.014 & \cellcolor{orange!20}0.692 & \cellcolor{orange!65}0.027 & \cellcolor{orange!65}0.394 \\
\specialrule{0.10em}{0.10em}{0.10em}
\multirow{6}{*}{\rotatebox[origin=c]{90}{\parbox{1.8cm}{\centering\texttt{Qwen2.5-72B-\\Instruct}}}} & \texttt{asiam} & \cellcolor{yellow!55}0.800 & \cellcolor{orange!35}0.500 & \cellcolor{yellow!55}0.615 & \cellcolor{orange!20}0.119 & \cellcolor{yellow!55}0.778 & \cellcolor{yellow!55}0.875 & \cellcolor{yellow!55}0.824 & \cellcolor{yellow!35}0.071 & \cellcolor{yellow!55}0.571 & \cellcolor{yellow!55}1.000 & \cellcolor{yellow!55}0.727 & \cellcolor{orange!20}0.143 & \cellcolor{yellow!55}0.583 & \cellcolor{yellow!55}0.875 & \cellcolor{yellow!55}0.700 & \cellcolor{orange!20}0.143 & \cellcolor{yellow!55}0.667 & \cellcolor{yellow!55}1.000 & \cellcolor{yellow!55}0.800 & \cellcolor{yellow!35}0.095 \\
 & \texttt{river} & \cellcolor{orange!65}0.000 & \cellcolor{orange!65}0.000 & \cellcolor{orange!65}0.000 & \cellcolor{orange!20}0.152 & \cellcolor{orange!20}0.154 & \cellcolor{orange!65}0.080 & \cellcolor{orange!35}0.105 & \cellcolor{orange!20}0.162 & \cellcolor{orange!20}0.213 & \cellcolor{orange!20}0.760 & \cellcolor{orange!20}0.333 & \cellcolor{orange!65}0.362 & \cellcolor{orange!20}0.213 & \cellcolor{orange!35}0.520 & \cellcolor{orange!20}0.302 & \cellcolor{orange!50}0.286 & \cellcolor{orange!20}0.221 & \cellcolor{yellow!55}0.920 & \cellcolor{orange!20}0.357 & \cellcolor{orange!65}0.395 \\
 & \texttt{covid} & \cellcolor{yellow!55}1.000 & \cellcolor{orange!50}0.269 & \cellcolor{yellow!35}0.424 & \cellcolor{yellow!55}0.050 & \cellcolor{yellow!55}0.727 & \cellcolor{orange!50}0.308 & \cellcolor{yellow!35}0.432 & \cellcolor{yellow!55}0.055 & \cellcolor{yellow!55}0.611 & \cellcolor{orange!50}0.423 & \cellcolor{yellow!35}0.500 & \cellcolor{yellow!55}0.058 & \cellcolor{orange!20}0.311 & \cellcolor{orange!35}0.538 & \cellcolor{yellow!35}0.394 & \cellcolor{orange!20}0.113 & \cellcolor{orange!20}0.258 & \cellcolor{orange!35}0.654 & \cellcolor{orange!20}0.370 & \cellcolor{orange!20}0.153 \\
 & \texttt{coal} & -- & \cellcolor{orange!65}0.000 & \cellcolor{orange!65}0.000 & \cellcolor{yellow!55}0.026 & \cellcolor{orange!35}0.144 & \cellcolor{orange!50}0.385 & \cellcolor{orange!20}0.210 & \cellcolor{yellow!35}0.074 & \cellcolor{orange!35}0.092 & \cellcolor{orange!20}0.718 & \cellcolor{orange!35}0.163 & \cellcolor{orange!35}0.194 & \cellcolor{orange!35}0.075 & \cellcolor{orange!20}0.718 & \cellcolor{orange!35}0.136 & \cellcolor{orange!35}0.239 & \cellcolor{orange!50}0.053 & \cellcolor{orange!20}0.744 & \cellcolor{orange!50}0.098 & \cellcolor{orange!65}0.358 \\
 & \texttt{hepar2} & \cellcolor{orange!35}0.125 & \cellcolor{orange!50}0.260 & \cellcolor{orange!35}0.168 & \cellcolor{yellow!55}0.064 & \cellcolor{orange!35}0.133 & \cellcolor{orange!50}0.374 & \cellcolor{orange!20}0.196 & \cellcolor{yellow!35}0.077 & \cellcolor{orange!35}0.094 & \cellcolor{orange!35}0.585 & \cellcolor{orange!35}0.162 & \cellcolor{orange!20}0.152 & \cellcolor{orange!35}0.081 & \cellcolor{orange!20}0.724 & \cellcolor{orange!35}0.146 & \cellcolor{orange!35}0.214 & \cellcolor{orange!35}0.075 & \cellcolor{orange!20}0.764 & \cellcolor{orange!35}0.137 & \cellcolor{orange!35}0.246 \\
 & \texttt{munin1} & \cellcolor{orange!65}0.008 & \cellcolor{orange!65}0.040 & \cellcolor{orange!65}0.014 & \cellcolor{yellow!55}0.045 & \cellcolor{orange!50}0.034 & \cellcolor{orange!50}0.256 & \cellcolor{orange!50}0.060 & \cellcolor{yellow!55}0.063 & \cellcolor{orange!65}0.017 & \cellcolor{orange!35}0.615 & \cellcolor{orange!65}0.033 & \cellcolor{orange!50}0.281 & \cellcolor{orange!65}0.016 & \cellcolor{orange!35}0.571 & \cellcolor{orange!65}0.031 & \cellcolor{orange!50}0.286 & \cellcolor{orange!65}0.012 & \cellcolor{orange!20}0.751 & \cellcolor{orange!65}0.023 & \cellcolor{orange!65}0.512 \\
\specialrule{0.10em}{0.10em}{0.10em}
\multirow{6}{*}{\rotatebox[origin=c]{90}{\parbox{1.8cm}{\centering\texttt{Llama-3.3-70B-\\Instruct}}}} & \texttt{asiam} & \cellcolor{yellow!55}0.750 & \cellcolor{orange!50}0.375 & \cellcolor{yellow!35}0.500 & \cellcolor{orange!20}0.143 & \cellcolor{yellow!55}0.833 & \cellcolor{orange!35}0.625 & \cellcolor{yellow!55}0.714 & \cellcolor{yellow!35}0.095 & \cellcolor{yellow!35}0.533 & \cellcolor{yellow!55}1.000 & \cellcolor{yellow!55}0.696 & \cellcolor{orange!20}0.167 & \cellcolor{yellow!35}0.471 & \cellcolor{yellow!55}1.000 & \cellcolor{yellow!55}0.640 & \cellcolor{orange!35}0.214 & \cellcolor{yellow!35}0.471 & \cellcolor{yellow!55}1.000 & \cellcolor{yellow!55}0.640 & \cellcolor{orange!35}0.214 \\
 & \texttt{river} & \cellcolor{yellow!35}0.357 & \cellcolor{orange!65}0.200 & \cellcolor{orange!20}0.256 & \cellcolor{orange!20}0.138 & \cellcolor{orange!35}0.143 & \cellcolor{orange!65}0.040 & \cellcolor{orange!50}0.062 & \cellcolor{orange!20}0.143 & \cellcolor{orange!20}0.216 & \cellcolor{yellow!55}0.880 & \cellcolor{orange!20}0.346 & \cellcolor{orange!65}0.386 & \cellcolor{orange!20}0.234 & \cellcolor{orange!35}0.600 & \cellcolor{orange!20}0.337 & \cellcolor{orange!50}0.281 & \cellcolor{orange!20}0.190 & \cellcolor{orange!20}0.760 & \cellcolor{orange!20}0.304 & \cellcolor{orange!65}0.414 \\
 & \texttt{covid} & \cellcolor{yellow!35}0.462 & \cellcolor{orange!65}0.231 & \cellcolor{orange!20}0.308 & \cellcolor{yellow!35}0.068 & \cellcolor{yellow!35}0.400 & \cellcolor{orange!65}0.154 & \cellcolor{orange!20}0.222 & \cellcolor{yellow!35}0.071 & \cellcolor{yellow!35}0.357 & \cellcolor{orange!20}0.769 & \cellcolor{yellow!35}0.488 & \cellcolor{orange!20}0.111 & \cellcolor{yellow!35}0.333 & \cellcolor{orange!20}0.769 & \cellcolor{yellow!35}0.465 & \cellcolor{orange!20}0.121 & \cellcolor{orange!20}0.244 & \cellcolor{orange!20}0.731 & \cellcolor{orange!20}0.365 & \cellcolor{orange!20}0.171 \\
 & \texttt{coal} & \cellcolor{orange!65}0.000 & \cellcolor{orange!65}0.000 & \cellcolor{orange!65}0.000 & \cellcolor{yellow!55}0.031 & \cellcolor{orange!20}0.286 & \cellcolor{orange!65}0.051 & \cellcolor{orange!50}0.087 & \cellcolor{yellow!55}0.028 & \cellcolor{orange!50}0.042 & \cellcolor{orange!20}0.769 & \cellcolor{orange!50}0.079 & \cellcolor{orange!65}0.470 & \cellcolor{orange!35}0.073 & \cellcolor{orange!35}0.590 & \cellcolor{orange!35}0.130 & \cellcolor{orange!35}0.207 & \cellcolor{orange!50}0.037 & \cellcolor{orange!20}0.769 & \cellcolor{orange!50}0.072 & \cellcolor{orange!65}0.523 \\
 & \texttt{hepar2} & \cellcolor{orange!20}0.148 & \cellcolor{orange!50}0.276 & \cellcolor{orange!20}0.193 & \cellcolor{yellow!55}0.057 & \cellcolor{orange!35}0.097 & \cellcolor{orange!65}0.106 & \cellcolor{orange!50}0.101 & \cellcolor{yellow!55}0.047 & \cellcolor{orange!50}0.065 & \cellcolor{yellow!35}0.813 & \cellcolor{orange!35}0.120 & \cellcolor{orange!50}0.303 & \cellcolor{orange!35}0.090 & \cellcolor{orange!35}0.691 & \cellcolor{orange!35}0.159 & \cellcolor{orange!35}0.184 & \cellcolor{orange!50}0.056 & \cellcolor{yellow!55}0.878 & \cellcolor{orange!35}0.106 & \cellcolor{orange!65}0.378 \\
 & \texttt{munin1} & \cellcolor{orange!50}0.029 & \cellcolor{orange!65}0.004 & \cellcolor{orange!65}0.007 & \cellcolor{yellow!55}0.009 & \cellcolor{orange!65}0.010 & \cellcolor{orange!65}0.004 & \cellcolor{orange!65}0.005 & \cellcolor{yellow!55}0.011 & \cellcolor{orange!65}0.014 & \cellcolor{yellow!35}0.835 & \cellcolor{orange!65}0.027 & \cellcolor{orange!65}0.485 & \cellcolor{orange!50}0.034 & \cellcolor{orange!65}0.253 & \cellcolor{orange!50}0.060 & \cellcolor{yellow!55}0.062 & \cellcolor{orange!65}0.010 & \cellcolor{yellow!55}0.912 & \cellcolor{orange!65}0.020 & \cellcolor{orange!65}0.717 \\
\specialrule{0.10em}{0.10em}{0.10em}
\multirow{6}{*}{\rotatebox[origin=c]{90}{\parbox{1.8cm}{\centering\texttt{Llama-3.1-70B-\\Instruct}}}} & \texttt{asiam} & \cellcolor{yellow!35}0.333 & \cellcolor{orange!50}0.375 & \cellcolor{orange!20}0.353 & \cellcolor{orange!35}0.238 & \cellcolor{yellow!55}0.538 & \cellcolor{yellow!55}0.875 & \cellcolor{yellow!55}0.667 & \cellcolor{orange!20}0.143 & \cellcolor{yellow!35}0.381 & \cellcolor{yellow!55}1.000 & \cellcolor{yellow!55}0.552 & \cellcolor{orange!50}0.310 & \cellcolor{yellow!35}0.400 & \cellcolor{orange!20}0.750 & \cellcolor{yellow!55}0.522 & \cellcolor{orange!35}0.214 & \cellcolor{yellow!55}0.538 & \cellcolor{yellow!55}0.875 & \cellcolor{yellow!55}0.667 & \cellcolor{orange!20}0.143 \\
 & \texttt{river} & \cellcolor{orange!20}0.214 & \cellcolor{orange!50}0.360 & \cellcolor{orange!20}0.269 & \cellcolor{orange!35}0.233 & \cellcolor{orange!20}0.200 & \cellcolor{orange!65}0.160 & \cellcolor{orange!35}0.178 & \cellcolor{orange!35}0.176 & \cellcolor{orange!20}0.230 & \cellcolor{yellow!55}0.920 & \cellcolor{orange!20}0.368 & \cellcolor{orange!65}0.371 & \cellcolor{orange!20}0.250 & \cellcolor{yellow!35}0.800 & \cellcolor{yellow!35}0.381 & \cellcolor{orange!50}0.310 & \cellcolor{orange!20}0.171 & \cellcolor{yellow!35}0.800 & \cellcolor{orange!20}0.282 & \cellcolor{orange!65}0.486 \\
 & \texttt{covid} & \cellcolor{yellow!35}0.333 & \cellcolor{orange!50}0.346 & \cellcolor{orange!20}0.340 & \cellcolor{yellow!35}0.092 & \cellcolor{yellow!55}0.538 & \cellcolor{orange!50}0.269 & \cellcolor{orange!20}0.359 & \cellcolor{yellow!55}0.063 & \cellcolor{yellow!35}0.380 & \cellcolor{orange!20}0.731 & \cellcolor{yellow!35}0.500 & \cellcolor{yellow!35}0.100 & \cellcolor{orange!20}0.250 & \cellcolor{yellow!35}0.808 & \cellcolor{yellow!35}0.382 & \cellcolor{orange!35}0.179 & \cellcolor{orange!20}0.256 & \cellcolor{yellow!35}0.808 & \cellcolor{yellow!35}0.389 & \cellcolor{orange!35}0.174 \\
 & \texttt{coal} & \cellcolor{orange!35}0.111 & \cellcolor{orange!65}0.051 & \cellcolor{orange!50}0.070 & \cellcolor{yellow!55}0.036 & \cellcolor{orange!35}0.070 & \cellcolor{orange!65}0.128 & \cellcolor{orange!50}0.091 & \cellcolor{yellow!55}0.065 & \cellcolor{orange!50}0.046 & \cellcolor{yellow!35}0.821 & \cellcolor{orange!50}0.086 & \cellcolor{orange!65}0.455 & \cellcolor{orange!35}0.077 & \cellcolor{orange!35}0.564 & \cellcolor{orange!35}0.136 & \cellcolor{orange!35}0.186 & \cellcolor{orange!50}0.042 & \cellcolor{orange!20}0.769 & \cellcolor{orange!50}0.080 & \cellcolor{orange!65}0.462 \\
 & \texttt{hepar2} & \cellcolor{orange!35}0.086 & \cellcolor{orange!50}0.382 & \cellcolor{orange!35}0.140 & \cellcolor{orange!20}0.117 & \cellcolor{orange!35}0.107 & \cellcolor{orange!50}0.341 & \cellcolor{orange!35}0.163 & \cellcolor{yellow!35}0.087 & \cellcolor{orange!50}0.055 & \cellcolor{yellow!35}0.813 & \cellcolor{orange!50}0.103 & \cellcolor{orange!65}0.359 & \cellcolor{orange!50}0.063 & \cellcolor{orange!20}0.740 & \cellcolor{orange!35}0.116 & \cellcolor{orange!50}0.286 & \cellcolor{orange!50}0.051 & \cellcolor{yellow!35}0.821 & \cellcolor{orange!50}0.095 & \cellcolor{orange!65}0.396 \\
 & \texttt{munin1} & \cellcolor{orange!50}0.020 & \cellcolor{orange!65}0.121 & \cellcolor{orange!50}0.034 & \cellcolor{yellow!55}0.054 & \cellcolor{orange!50}0.029 & \cellcolor{orange!65}0.136 & \cellcolor{orange!50}0.048 & \cellcolor{yellow!55}0.042 & \cellcolor{orange!65}0.011 & \cellcolor{yellow!55}0.927 & \cellcolor{orange!65}0.023 & \cellcolor{orange!65}0.638 & \cellcolor{orange!65}0.014 & \cellcolor{orange!35}0.505 & \cellcolor{orange!65}0.028 & \cellcolor{orange!50}0.282 & \cellcolor{orange!65}0.010 & \cellcolor{yellow!55}0.908 & \cellcolor{orange!65}0.019 & \cellcolor{orange!65}0.745 \\
\bottomrule
\end{tabular}%
}
\caption{Primary causal edge classification performance for large language models across six benchmark datasets and five prompt styles. Cell colors encode performance percentile within the full table (yellow~=~good, orange~=~bad), with intensity scaled per metric direction (P/R/F1 higher is better; nSHD lower is better). A dash denotes undefined precision because the model made no positive predictions.}
\label{tab:primary_classification_large_full}
\end{table}
\endgroup

\clearpage
\begin{landscape}
\subsubsection{Calibration Tables}
\begingroup
\scriptsize
\setlength{\tabcolsep}{3pt}
\renewcommand{\arraystretch}{1.05}
\begin{xltabular}{\linewidth}{@{}lll *{15}{>{\centering\arraybackslash}X}@{}}
\toprule
\multirow{2}{*}{\textbf{Model}} & \multirow{2}{*}{\textbf{Dataset}} & \multirow{2}{*}{\textbf{Method}} & \multicolumn{3}{c}{\textbf{Name-only}} & \multicolumn{3}{c}{\textbf{Metadata}} & \multicolumn{3}{c}{\textbf{CoT}} & \multicolumn{3}{c}{\textbf{Few-shot}} & \multicolumn{3}{c}{\textbf{Few-shot + CoT}} \\
\cmidrule(lr){4-6} \cmidrule(lr){7-9} \cmidrule(lr){10-12} \cmidrule(lr){13-15} \cmidrule(lr){16-18}
&  &  & AUROC $\uparrow$ & ECE $\downarrow$ & Brier $\downarrow$ & AUROC $\uparrow$ & ECE $\downarrow$ & Brier $\downarrow$ & AUROC $\uparrow$ & ECE $\downarrow$ & Brier $\downarrow$ & AUROC $\uparrow$ & ECE $\downarrow$ & Brier $\downarrow$ & AUROC $\uparrow$ & ECE $\downarrow$ & Brier $\downarrow$ \\
\midrule
\endfirsthead

\multicolumn{18}{l}{\emph{(continued from previous page)}}\\
\toprule
\multirow{2}{*}{\textbf{Model}} & \multirow{2}{*}{\textbf{Dataset}} & \multirow{2}{*}{\textbf{Method}} & \multicolumn{3}{c}{\textbf{Name-only}} & \multicolumn{3}{c}{\textbf{Metadata}} & \multicolumn{3}{c}{\textbf{CoT}} & \multicolumn{3}{c}{\textbf{Few-shot}} & \multicolumn{3}{c}{\textbf{Few-shot + CoT}} \\
\cmidrule(lr){4-6} \cmidrule(lr){7-9} \cmidrule(lr){10-12} \cmidrule(lr){13-15} \cmidrule(lr){16-18}
&  &  & AUROC $\uparrow$ & ECE $\downarrow$ & Brier $\downarrow$ & AUROC $\uparrow$ & ECE $\downarrow$ & Brier $\downarrow$ & AUROC $\uparrow$ & ECE $\downarrow$ & Brier $\downarrow$ & AUROC $\uparrow$ & ECE $\downarrow$ & Brier $\downarrow$ & AUROC $\uparrow$ & ECE $\downarrow$ & Brier $\downarrow$ \\
\midrule
\endhead

\bottomrule
\multicolumn{18}{r}{\emph{(continued on next page)}}\\
\endfoot

\bottomrule
\addlinespace[1.5em]
\caption{Prompt-specific calibration results for small LLMs on \texttt{asiam}, \texttt{river}, \texttt{covid}, \texttt{coal}, \texttt{hepar2}, and \texttt{munin1}. Cell colors encode performance percentile within the full table (yellow~=~good, orange~=~bad), with intensity scaled separately per metric direction.}
\label{tab:calib_all_smallmodels}\\
\label{tab:calib_prompt_small_models}\\
\endlastfoot

\multirow{12}{*}{\rotatebox[origin=c]{90}{\parbox{1.8cm}{\centering Gemma-4-\\E4B-IT}}} & \multirow{2}{*}{\texttt{asiam}} & Verb. & \cellcolor{orange!35}0.397 & \cellcolor{orange!65}0.513 & \cellcolor{orange!65}0.641 & \cellcolor{orange!50}0.244 & \cellcolor{orange!50}0.417 & \cellcolor{orange!50}0.539 & \cellcolor{yellow!55}0.878 & \cellcolor{yellow!55}0.010 & \cellcolor{yellow!55}0.103 & \cellcolor{orange!35}0.375 & \cellcolor{orange!20}0.051 & \cellcolor{orange!20}0.219 & \cellcolor{orange!35}0.384 & \cellcolor{yellow!35}0.024 & \cellcolor{yellow!35}0.142 \\*
 &  & Logit & \cellcolor{yellow!35}0.602 & \cellcolor{yellow!35}0.023 & \cellcolor{yellow!35}0.143 & \cellcolor{yellow!55}0.872 & \cellcolor{yellow!35}0.021 & \cellcolor{yellow!55}0.063 & \cellcolor{orange!20}0.500 & \cellcolor{orange!65}0.655 & \cellcolor{orange!65}0.810 & \cellcolor{orange!20}0.562 & \cellcolor{yellow!35}0.044 & \cellcolor{yellow!35}0.205 & \cellcolor{yellow!35}0.665 & \cellcolor{orange!20}0.054 & \cellcolor{yellow!35}0.141 \\*
\cmidrule(lr){2-18}
 & \multirow{2}{*}{\texttt{river}} & Verb. & \cellcolor{orange!50}0.320 & \cellcolor{orange!50}0.356 & \cellcolor{orange!50}0.485 & \cellcolor{orange!65}0.192 & \cellcolor{orange!50}0.389 & \cellcolor{orange!50}0.503 & \cellcolor{orange!35}0.395 & \cellcolor{orange!20}0.088 & \cellcolor{orange!20}0.280 & \cellcolor{orange!65}0.195 & \cellcolor{orange!50}0.316 & \cellcolor{orange!50}0.460 & \cellcolor{orange!50}0.266 & \cellcolor{orange!35}0.255 & \cellcolor{orange!35}0.432 \\*
 &  & Logit & \cellcolor{yellow!35}0.691 & \cellcolor{yellow!35}0.019 & \cellcolor{yellow!35}0.138 & \cellcolor{yellow!35}0.635 & \cellcolor{yellow!35}0.016 & \cellcolor{yellow!35}0.155 & \cellcolor{orange!20}0.503 & \cellcolor{orange!65}0.552 & \cellcolor{orange!65}0.742 & \cellcolor{orange!35}0.485 & \cellcolor{orange!35}0.262 & \cellcolor{orange!50}0.509 & \cellcolor{orange!20}0.547 & \cellcolor{orange!50}0.294 & \cellcolor{orange!50}0.525 \\*
\cmidrule(lr){2-18}
 & \multirow{2}{*}{\texttt{covid}} & Verb. & \cellcolor{orange!35}0.398 & \cellcolor{orange!65}0.550 & \cellcolor{orange!65}0.634 & \cellcolor{orange!50}0.342 & \cellcolor{orange!65}0.590 & \cellcolor{orange!65}0.657 & \cellcolor{yellow!35}0.661 & \cellcolor{yellow!55}0.009 & \cellcolor{yellow!55}0.074 & \cellcolor{orange!50}0.233 & \cellcolor{orange!20}0.081 & \cellcolor{orange!20}0.217 & \cellcolor{orange!50}0.349 & \cellcolor{orange!20}0.057 & \cellcolor{orange!20}0.234 \\*
 &  & Logit & \cellcolor{yellow!35}0.624 & \cellcolor{yellow!55}0.011 & \cellcolor{yellow!55}0.063 & \cellcolor{yellow!35}0.613 & \cellcolor{yellow!55}0.005 & \cellcolor{yellow!55}0.058 & \cellcolor{orange!20}0.500 & \cellcolor{orange!65}0.590 & \cellcolor{orange!65}0.768 & \cellcolor{yellow!35}0.660 & \cellcolor{yellow!35}0.036 & \cellcolor{yellow!35}0.178 & \cellcolor{yellow!35}0.606 & \cellcolor{yellow!35}0.045 & \cellcolor{orange!20}0.225 \\*
\cmidrule(lr){2-18}
 & \multirow{2}{*}{\texttt{coal}} & Verb. & \cellcolor{orange!20}0.541 & \cellcolor{orange!65}0.521 & \cellcolor{orange!50}0.561 & \cellcolor{orange!50}0.245 & \cellcolor{orange!65}0.500 & \cellcolor{orange!50}0.547 & \cellcolor{orange!35}0.405 & \cellcolor{orange!20}0.059 & \cellcolor{yellow!35}0.143 & \cellcolor{orange!65}0.176 & \cellcolor{orange!50}0.298 & \cellcolor{orange!50}0.461 & \cellcolor{orange!65}0.163 & \cellcolor{orange!35}0.260 & \cellcolor{orange!35}0.379 \\*
 &  & Logit & \cellcolor{orange!35}0.499 & \cellcolor{yellow!55}0.001 & \cellcolor{yellow!55}0.026 & \cellcolor{yellow!55}0.774 & \cellcolor{yellow!55}0.002 & \cellcolor{yellow!55}0.044 & \cellcolor{orange!20}0.500 & \cellcolor{orange!65}0.819 & \cellcolor{orange!65}0.905 & \cellcolor{orange!35}0.492 & \cellcolor{orange!35}0.263 & \cellcolor{orange!50}0.510 & \cellcolor{orange!20}0.585 & \cellcolor{orange!20}0.142 & \cellcolor{orange!35}0.381 \\*
\cmidrule(lr){2-18}
 & \multirow{2}{*}{\texttt{hepar2}} & Verb. & \cellcolor{orange!65}0.144 & \cellcolor{orange!65}0.533 & \cellcolor{orange!65}0.583 & \cellcolor{orange!65}0.132 & \cellcolor{orange!65}0.548 & \cellcolor{orange!65}0.598 & \cellcolor{orange!35}0.423 & \cellcolor{yellow!35}0.037 & \cellcolor{yellow!55}0.097 & \cellcolor{orange!65}0.160 & \cellcolor{orange!20}0.130 & \cellcolor{orange!20}0.297 & \cellcolor{orange!50}0.240 & \cellcolor{orange!20}0.108 & \cellcolor{orange!20}0.254 \\*
 &  & Logit & \cellcolor{yellow!55}0.776 & \cellcolor{yellow!55}0.003 & \cellcolor{yellow!55}0.065 & \cellcolor{yellow!35}0.725 & \cellcolor{yellow!55}0.006 & \cellcolor{yellow!55}0.086 & \cellcolor{orange!20}0.500 & \cellcolor{orange!65}0.932 & \cellcolor{orange!65}0.965 & \cellcolor{orange!20}0.571 & \cellcolor{orange!20}0.087 & \cellcolor{orange!20}0.306 & \cellcolor{yellow!35}0.597 & \cellcolor{orange!20}0.105 & \cellcolor{orange!20}0.328 \\*
\cmidrule(lr){2-18}
 & \multirow{2}{*}{\texttt{munin1}} & Verb. & \cellcolor{orange!65}0.136 & \cellcolor{orange!65}0.523 & \cellcolor{orange!50}0.544 & \cellcolor{orange!65}0.104 & \cellcolor{orange!65}0.499 & \cellcolor{orange!50}0.527 & \cellcolor{orange!35}0.382 & \cellcolor{yellow!35}0.040 & \cellcolor{yellow!35}0.134 & \cellcolor{orange!65}0.036 & \cellcolor{orange!65}0.502 & \cellcolor{orange!65}0.575 & \cellcolor{orange!65}0.065 & \cellcolor{orange!50}0.465 & \cellcolor{orange!50}0.551 \\*
 &  & Logit & \cellcolor{yellow!55}0.806 & \cellcolor{yellow!55}0.001 & \cellcolor{yellow!55}0.015 & \cellcolor{yellow!55}0.824 & \cellcolor{yellow!55}0.000 & \cellcolor{yellow!55}0.030 & \cellcolor{orange!35}0.499 & \cellcolor{orange!65}0.854 & \cellcolor{orange!65}0.924 & \cellcolor{orange!35}0.454 & \cellcolor{orange!50}0.416 & \cellcolor{orange!65}0.619 & \cellcolor{orange!35}0.460 & \cellcolor{orange!50}0.409 & \cellcolor{orange!65}0.615 \\
\specialrule{0.10em}{0.10em}{0.10em}
\multirow{12}{*}{\rotatebox[origin=c]{90}{\parbox{1.8cm}{\centering Llama-3.1-8B-\\Instruct}}} & \multirow{2}{*}{\texttt{asiam}} & Verb. & \cellcolor{orange!65}0.220 & \cellcolor{orange!50}0.311 & \cellcolor{orange!50}0.472 & \cellcolor{orange!35}0.435 & \cellcolor{orange!35}0.231 & \cellcolor{orange!35}0.377 & \cellcolor{orange!35}0.473 & \cellcolor{yellow!35}0.023 & \cellcolor{yellow!35}0.155 & \cellcolor{orange!65}0.209 & \cellcolor{orange!50}0.313 & \cellcolor{orange!50}0.482 & \cellcolor{orange!35}0.438 & \cellcolor{yellow!35}0.037 & \cellcolor{yellow!35}0.198 \\*
 &  & Logit & \cellcolor{yellow!55}0.803 & \cellcolor{orange!20}0.059 & \cellcolor{orange!20}0.229 & \cellcolor{yellow!55}0.895 & \cellcolor{yellow!35}0.020 & \cellcolor{yellow!35}0.138 & \cellcolor{yellow!35}0.704 & \cellcolor{yellow!55}0.015 & \cellcolor{yellow!55}0.115 & \cellcolor{orange!20}0.580 & \cellcolor{orange!20}0.068 & \cellcolor{orange!20}0.303 & \cellcolor{yellow!35}0.690 & \cellcolor{orange!20}0.054 & \cellcolor{yellow!35}0.203 \\*
\cmidrule(lr){2-18}
 & \multirow{2}{*}{\texttt{river}} & Verb. & \cellcolor{orange!65}0.226 & \cellcolor{orange!35}0.285 & \cellcolor{orange!50}0.460 & \cellcolor{orange!50}0.343 & \cellcolor{orange!35}0.156 & \cellcolor{orange!35}0.354 & \cellcolor{orange!20}0.512 & \cellcolor{yellow!35}0.017 & \cellcolor{yellow!35}0.192 & \cellcolor{orange!50}0.337 & \cellcolor{orange!35}0.251 & \cellcolor{orange!35}0.458 & \cellcolor{orange!35}0.436 & \cellcolor{orange!20}0.097 & \cellcolor{orange!20}0.293 \\*
 &  & Logit & \cellcolor{orange!20}0.580 & \cellcolor{yellow!35}0.018 & \cellcolor{orange!20}0.256 & \cellcolor{yellow!35}0.723 & \cellcolor{yellow!55}0.004 & \cellcolor{yellow!35}0.158 & \cellcolor{yellow!55}0.785 & \cellcolor{yellow!55}0.014 & \cellcolor{yellow!35}0.162 & \cellcolor{orange!35}0.499 & \cellcolor{orange!35}0.164 & \cellcolor{orange!35}0.394 & \cellcolor{yellow!35}0.651 & \cellcolor{yellow!35}0.032 & \cellcolor{orange!20}0.240 \\*
\cmidrule(lr){2-18}
 & \multirow{2}{*}{\texttt{covid}} & Verb. & \cellcolor{orange!50}0.289 & \cellcolor{orange!50}0.368 & \cellcolor{orange!50}0.522 & \cellcolor{orange!50}0.332 & \cellcolor{orange!50}0.331 & \cellcolor{orange!35}0.459 & \cellcolor{orange!20}0.564 & \cellcolor{orange!20}0.051 & \cellcolor{yellow!35}0.149 & \cellcolor{orange!50}0.295 & \cellcolor{orange!35}0.174 & \cellcolor{orange!35}0.366 & \cellcolor{orange!35}0.468 & \cellcolor{orange!20}0.058 & \cellcolor{orange!20}0.220 \\*
 &  & Logit & \cellcolor{yellow!55}0.777 & \cellcolor{yellow!35}0.031 & \cellcolor{yellow!55}0.095 & \cellcolor{yellow!55}0.777 & \cellcolor{yellow!55}0.011 & \cellcolor{yellow!55}0.069 & \cellcolor{yellow!35}0.678 & \cellcolor{yellow!55}0.005 & \cellcolor{yellow!55}0.065 & \cellcolor{yellow!35}0.677 & \cellcolor{yellow!55}0.002 & \cellcolor{yellow!35}0.193 & \cellcolor{yellow!55}0.816 & \cellcolor{yellow!55}0.005 & \cellcolor{yellow!55}0.101 \\*
\cmidrule(lr){2-18}
 & \multirow{2}{*}{\texttt{coal}} & Verb. & \cellcolor{orange!65}0.127 & \cellcolor{orange!50}0.362 & \cellcolor{orange!50}0.510 & \cellcolor{orange!65}0.195 & \cellcolor{orange!35}0.263 & \cellcolor{orange!35}0.432 & \cellcolor{orange!35}0.449 & \cellcolor{yellow!35}0.028 & \cellcolor{yellow!35}0.205 & \cellcolor{orange!50}0.235 & \cellcolor{orange!35}0.236 & \cellcolor{orange!35}0.427 & \cellcolor{orange!35}0.443 & \cellcolor{orange!20}0.073 & \cellcolor{orange!20}0.271 \\*
 &  & Logit & \cellcolor{yellow!55}0.936 & \cellcolor{yellow!35}0.050 & \cellcolor{orange!20}0.215 & \cellcolor{yellow!55}0.748 & \cellcolor{orange!20}0.082 & \cellcolor{yellow!55}0.114 & \cellcolor{yellow!55}0.835 & \cellcolor{yellow!55}0.012 & \cellcolor{yellow!55}0.097 & \cellcolor{yellow!35}0.671 & \cellcolor{yellow!35}0.025 & \cellcolor{orange!20}0.240 & \cellcolor{yellow!55}0.809 & \cellcolor{yellow!55}0.006 & \cellcolor{yellow!35}0.156 \\*
\cmidrule(lr){2-18}
 & \multirow{2}{*}{\texttt{hepar2}} & Verb. & \cellcolor{orange!65}0.208 & \cellcolor{orange!50}0.340 & \cellcolor{orange!50}0.503 & \cellcolor{orange!50}0.232 & \cellcolor{orange!50}0.322 & \cellcolor{orange!50}0.478 & \cellcolor{orange!20}0.518 & \cellcolor{yellow!35}0.033 & \cellcolor{yellow!35}0.153 & \cellcolor{orange!65}0.210 & \cellcolor{orange!35}0.243 & \cellcolor{orange!35}0.422 & \cellcolor{orange!20}0.502 & \cellcolor{orange!20}0.053 & \cellcolor{orange!20}0.246 \\*
 &  & Logit & \cellcolor{yellow!55}0.809 & \cellcolor{yellow!35}0.026 & \cellcolor{yellow!35}0.151 & \cellcolor{yellow!55}0.866 & \cellcolor{yellow!35}0.032 & \cellcolor{yellow!55}0.094 & \cellcolor{yellow!55}0.864 & \cellcolor{yellow!55}0.003 & \cellcolor{yellow!55}0.066 & \cellcolor{yellow!35}0.613 & \cellcolor{yellow!35}0.046 & \cellcolor{orange!20}0.278 & \cellcolor{yellow!35}0.740 & \cellcolor{yellow!35}0.017 & \cellcolor{yellow!35}0.186 \\*
\cmidrule(lr){2-18}
 & \multirow{2}{*}{\texttt{munin1}} & Verb. & \cellcolor{orange!65}0.123 & \cellcolor{orange!50}0.367 & \cellcolor{orange!50}0.512 & \cellcolor{orange!50}0.232 & \cellcolor{orange!35}0.206 & \cellcolor{orange!35}0.393 & \cellcolor{orange!35}0.464 & \cellcolor{yellow!35}0.028 & \cellcolor{orange!20}0.227 & \cellcolor{orange!65}0.207 & \cellcolor{orange!50}0.390 & \cellcolor{orange!50}0.547 & \cellcolor{orange!35}0.406 & \cellcolor{orange!35}0.192 & \cellcolor{orange!35}0.423 \\*
 &  & Logit & \cellcolor{yellow!35}0.643 & \cellcolor{yellow!35}0.030 & \cellcolor{orange!20}0.257 & \cellcolor{yellow!55}0.874 & \cellcolor{orange!20}0.064 & \cellcolor{yellow!55}0.095 & \cellcolor{yellow!55}0.829 & \cellcolor{yellow!55}0.008 & \cellcolor{yellow!35}0.121 & \cellcolor{orange!35}0.467 & \cellcolor{orange!50}0.303 & \cellcolor{orange!35}0.456 & \cellcolor{yellow!35}0.638 & \cellcolor{orange!35}0.159 & \cellcolor{orange!35}0.333 \\
\specialrule{0.10em}{0.10em}{0.10em}
\multirow{12}{*}{\rotatebox[origin=c]{90}{\parbox{1.8cm}{\centering Ministral-8B-\\Instruct-2410}}} & \multirow{2}{*}{\texttt{asiam}} & Verb. & \cellcolor{orange!20}0.503 & \cellcolor{orange!50}0.387 & \cellcolor{orange!65}0.574 & \cellcolor{orange!35}0.497 & \cellcolor{orange!20}0.131 & \cellcolor{orange!35}0.350 & \cellcolor{orange!20}0.580 & \cellcolor{orange!50}0.392 & \cellcolor{orange!65}0.580 & \cellcolor{orange!20}0.506 & \cellcolor{orange!35}0.175 & \cellcolor{orange!35}0.405 & \cellcolor{orange!20}0.540 & \cellcolor{orange!20}0.147 & \cellcolor{orange!35}0.393 \\*
 &  & Logit & \cellcolor{yellow!35}0.636 & \cellcolor{orange!35}0.242 & \cellcolor{orange!35}0.390 & \cellcolor{orange!20}0.567 & \cellcolor{orange!20}0.152 & \cellcolor{orange!35}0.336 & \cellcolor{yellow!35}0.717 & \cellcolor{orange!35}0.273 & \cellcolor{orange!35}0.417 & \cellcolor{orange!20}0.548 & \cellcolor{orange!35}0.200 & \cellcolor{orange!35}0.375 & \cellcolor{yellow!35}0.663 & \cellcolor{orange!50}0.292 & \cellcolor{orange!35}0.429 \\*
\cmidrule(lr){2-18}
 & \multirow{2}{*}{\texttt{river}} & Verb. & \cellcolor{orange!35}0.391 & \cellcolor{orange!35}0.245 & \cellcolor{orange!50}0.472 & \cellcolor{orange!35}0.420 & \cellcolor{orange!35}0.183 & \cellcolor{orange!35}0.418 & \cellcolor{orange!35}0.458 & \cellcolor{orange!50}0.320 & \cellcolor{orange!50}0.529 & \cellcolor{orange!50}0.306 & \cellcolor{orange!35}0.181 & \cellcolor{orange!35}0.410 & \cellcolor{orange!35}0.410 & \cellcolor{orange!50}0.323 & \cellcolor{orange!50}0.535 \\*
 &  & Logit & \cellcolor{orange!20}0.543 & \cellcolor{orange!35}0.272 & \cellcolor{orange!35}0.376 & \cellcolor{orange!20}0.516 & \cellcolor{orange!20}0.135 & \cellcolor{orange!20}0.285 & \cellcolor{orange!35}0.431 & \cellcolor{orange!50}0.290 & \cellcolor{orange!35}0.403 & \cellcolor{orange!20}0.569 & \cellcolor{orange!35}0.240 & \cellcolor{orange!35}0.365 & \cellcolor{yellow!35}0.716 & \cellcolor{orange!50}0.384 & \cellcolor{orange!50}0.482 \\*
\cmidrule(lr){2-18}
 & \multirow{2}{*}{\texttt{covid}} & Verb. & \cellcolor{orange!35}0.423 & \cellcolor{orange!35}0.209 & \cellcolor{orange!35}0.442 & \cellcolor{orange!35}0.470 & \cellcolor{orange!35}0.162 & \cellcolor{orange!35}0.388 & \cellcolor{orange!20}0.519 & \cellcolor{orange!35}0.168 & \cellcolor{orange!35}0.408 & \cellcolor{orange!50}0.364 & \cellcolor{orange!35}0.225 & \cellcolor{orange!35}0.453 & \cellcolor{orange!35}0.438 & \cellcolor{orange!35}0.238 & \cellcolor{orange!50}0.475 \\*
 &  & Logit & \cellcolor{orange!20}0.565 & \cellcolor{orange!35}0.261 & \cellcolor{orange!35}0.342 & \cellcolor{yellow!35}0.674 & \cellcolor{yellow!35}0.022 & \cellcolor{orange!20}0.233 & \cellcolor{orange!35}0.424 & \cellcolor{yellow!35}0.023 & \cellcolor{orange!20}0.265 & \cellcolor{orange!20}0.584 & \cellcolor{orange!50}0.303 & \cellcolor{orange!35}0.380 & \cellcolor{orange!20}0.503 & \cellcolor{orange!50}0.296 & \cellcolor{orange!35}0.388 \\*
\cmidrule(lr){2-18}
 & \multirow{2}{*}{\texttt{coal}} & Verb. & \cellcolor{orange!35}0.383 & \cellcolor{orange!50}0.293 & \cellcolor{orange!50}0.514 & \cellcolor{orange!50}0.368 & \cellcolor{orange!35}0.247 & \cellcolor{orange!50}0.474 & \cellcolor{orange!20}0.504 & \cellcolor{orange!35}0.280 & \cellcolor{orange!50}0.508 & \cellcolor{orange!50}0.320 & \cellcolor{orange!35}0.267 & \cellcolor{orange!50}0.481 & \cellcolor{orange!35}0.421 & \cellcolor{orange!50}0.337 & \cellcolor{orange!50}0.550 \\*
 &  & Logit & \cellcolor{orange!50}0.292 & \cellcolor{orange!50}0.325 & \cellcolor{orange!35}0.350 & \cellcolor{orange!50}0.284 & \cellcolor{orange!35}0.261 & \cellcolor{orange!20}0.313 & \cellcolor{orange!50}0.306 & \cellcolor{orange!50}0.314 & \cellcolor{orange!35}0.368 & \cellcolor{orange!50}0.308 & \cellcolor{orange!35}0.267 & \cellcolor{orange!35}0.361 & \cellcolor{orange!50}0.323 & \cellcolor{orange!50}0.465 & \cellcolor{orange!50}0.505 \\*
\cmidrule(lr){2-18}
 & \multirow{2}{*}{\texttt{hepar2}} & Verb. & \cellcolor{orange!35}0.378 & \cellcolor{orange!35}0.280 & \cellcolor{orange!50}0.505 & \cellcolor{orange!35}0.412 & \cellcolor{orange!35}0.232 & \cellcolor{orange!50}0.465 & \cellcolor{orange!20}0.555 & \cellcolor{orange!35}0.222 & \cellcolor{orange!50}0.463 & \cellcolor{orange!50}0.371 & \cellcolor{orange!20}0.120 & \cellcolor{orange!35}0.359 & \cellcolor{orange!35}0.462 & \cellcolor{orange!20}0.152 & \cellcolor{orange!35}0.401 \\*
 &  & Logit & \cellcolor{orange!35}0.432 & \cellcolor{orange!50}0.341 & \cellcolor{orange!35}0.381 & \cellcolor{orange!35}0.410 & \cellcolor{orange!35}0.167 & \cellcolor{orange!20}0.315 & \cellcolor{orange!50}0.311 & \cellcolor{orange!35}0.230 & \cellcolor{orange!35}0.341 & \cellcolor{orange!20}0.540 & \cellcolor{yellow!35}0.021 & \cellcolor{orange!20}0.265 & \cellcolor{orange!35}0.473 & \cellcolor{yellow!35}0.046 & \cellcolor{orange!20}0.290 \\*
\cmidrule(lr){2-18}
 & \multirow{2}{*}{\texttt{munin1}} & Verb. & \cellcolor{orange!50}0.368 & \cellcolor{orange!50}0.289 & \cellcolor{orange!50}0.511 & \cellcolor{orange!35}0.375 & \cellcolor{orange!35}0.254 & \cellcolor{orange!50}0.483 & \cellcolor{orange!20}0.516 & \cellcolor{orange!35}0.226 & \cellcolor{orange!50}0.467 & \cellcolor{orange!50}0.327 & \cellcolor{orange!35}0.195 & \cellcolor{orange!35}0.423 & \cellcolor{orange!35}0.438 & \cellcolor{orange!35}0.205 & \cellcolor{orange!35}0.450 \\*
 &  & Logit & \cellcolor{orange!35}0.498 & \cellcolor{orange!50}0.348 & \cellcolor{orange!35}0.357 & \cellcolor{orange!50}0.251 & \cellcolor{orange!50}0.290 & \cellcolor{orange!20}0.323 & \cellcolor{orange!65}0.218 & \cellcolor{orange!35}0.266 & \cellcolor{orange!35}0.332 & \cellcolor{orange!50}0.252 & \cellcolor{orange!35}0.228 & \cellcolor{orange!35}0.335 & \cellcolor{orange!50}0.231 & \cellcolor{orange!35}0.278 & \cellcolor{orange!35}0.371 \\
\specialrule{0.10em}{0.10em}{0.10em}
\multirow{12}{*}{\rotatebox[origin=c]{90}{Phi-4}} & \multirow{2}{*}{\texttt{asiam}} & Verb. & \cellcolor{yellow!35}0.680 & \cellcolor{orange!35}0.179 & \cellcolor{orange!20}0.274 & \cellcolor{yellow!35}0.624 & \cellcolor{orange!20}0.079 & \cellcolor{yellow!35}0.165 & \cellcolor{yellow!55}0.756 & \cellcolor{yellow!55}0.009 & \cellcolor{yellow!35}0.143 & \cellcolor{yellow!35}0.677 & \cellcolor{yellow!35}0.026 & \cellcolor{yellow!35}0.196 & \cellcolor{yellow!35}0.602 & \cellcolor{yellow!35}0.031 & \cellcolor{orange!20}0.233 \\*
 &  & Logit & \cellcolor{orange!20}0.508 & \cellcolor{yellow!35}0.019 & \cellcolor{yellow!35}0.120 & \cellcolor{yellow!55}0.784 & \cellcolor{yellow!35}0.016 & \cellcolor{yellow!55}0.103 & \cellcolor{yellow!55}0.893 & \cellcolor{orange!65}0.655 & \cellcolor{orange!65}0.810 & \cellcolor{yellow!55}0.869 & \cellcolor{orange!20}0.092 & \cellcolor{yellow!35}0.171 & \cellcolor{yellow!35}0.664 & \cellcolor{orange!20}0.079 & \cellcolor{orange!20}0.240 \\*
\cmidrule(lr){2-18}
 & \multirow{2}{*}{\texttt{river}} & Verb. & \cellcolor{yellow!55}0.772 & \cellcolor{yellow!35}0.018 & \cellcolor{yellow!35}0.156 & \cellcolor{yellow!35}0.660 & \cellcolor{yellow!35}0.039 & \cellcolor{yellow!35}0.162 & \cellcolor{yellow!55}0.836 & \cellcolor{orange!20}0.062 & \cellcolor{orange!20}0.210 & \cellcolor{orange!20}0.539 & \cellcolor{yellow!55}0.015 & \cellcolor{orange!20}0.212 & \cellcolor{orange!20}0.556 & \cellcolor{orange!20}0.072 & \cellcolor{orange!20}0.309 \\*
 &  & Logit & \cellcolor{yellow!55}0.771 & \cellcolor{yellow!35}0.024 & \cellcolor{yellow!35}0.152 & \cellcolor{yellow!55}0.783 & \cellcolor{yellow!55}0.012 & \cellcolor{yellow!35}0.142 & \cellcolor{yellow!35}0.627 & \cellcolor{orange!65}0.742 & \cellcolor{orange!65}0.857 & \cellcolor{yellow!55}0.829 & \cellcolor{orange!20}0.059 & \cellcolor{orange!20}0.237 & \cellcolor{orange!50}0.373 & \cellcolor{orange!35}0.237 & \cellcolor{orange!50}0.488 \\*
\cmidrule(lr){2-18}
 & \multirow{2}{*}{\texttt{covid}} & Verb. & \cellcolor{yellow!35}0.671 & \cellcolor{yellow!55}0.012 & \cellcolor{yellow!55}0.077 & \cellcolor{yellow!35}0.699 & \cellcolor{yellow!35}0.020 & \cellcolor{yellow!55}0.083 & \cellcolor{yellow!35}0.715 & \cellcolor{yellow!55}0.002 & \cellcolor{yellow!55}0.062 & \cellcolor{yellow!35}0.627 & \cellcolor{yellow!55}0.004 & \cellcolor{yellow!55}0.110 & \cellcolor{yellow!55}0.760 & \cellcolor{yellow!35}0.021 & \cellcolor{yellow!35}0.134 \\*
 &  & Logit & \cellcolor{yellow!55}0.813 & \cellcolor{yellow!55}0.007 & \cellcolor{yellow!55}0.053 & \cellcolor{yellow!55}0.775 & \cellcolor{yellow!55}0.002 & \cellcolor{yellow!55}0.047 & \cellcolor{yellow!35}0.671 & \cellcolor{orange!65}0.755 & \cellcolor{orange!65}0.869 & \cellcolor{yellow!55}0.829 & \cellcolor{yellow!55}0.005 & \cellcolor{yellow!55}0.086 & \cellcolor{orange!20}0.544 & \cellcolor{yellow!35}0.020 & \cellcolor{yellow!35}0.185 \\*
\cmidrule(lr){2-18}
 & \multirow{2}{*}{\texttt{coal}} & Verb. & \cellcolor{orange!35}0.468 & \cellcolor{orange!65}0.558 & \cellcolor{orange!65}0.627 & \cellcolor{orange!20}0.522 & \cellcolor{orange!35}0.183 & \cellcolor{orange!20}0.308 & \cellcolor{yellow!55}0.849 & \cellcolor{orange!20}0.072 & \cellcolor{orange!20}0.215 & \cellcolor{orange!35}0.483 & \cellcolor{yellow!35}0.035 & \cellcolor{orange!20}0.251 & \cellcolor{yellow!35}0.621 & \cellcolor{orange!20}0.105 & \cellcolor{orange!35}0.337 \\*
 &  & Logit & \cellcolor{yellow!35}0.636 & \cellcolor{yellow!55}0.001 & \cellcolor{yellow!55}0.026 & \cellcolor{yellow!55}0.778 & \cellcolor{yellow!35}0.019 & \cellcolor{yellow!35}0.164 & \cellcolor{orange!20}0.572 & \cellcolor{orange!65}0.913 & \cellcolor{orange!65}0.954 & \cellcolor{yellow!35}0.728 & \cellcolor{orange!20}0.060 & \cellcolor{orange!20}0.265 & \cellcolor{orange!35}0.410 & \cellcolor{orange!35}0.236 & \cellcolor{orange!50}0.484 \\*
\cmidrule(lr){2-18}
 & \multirow{2}{*}{\texttt{hepar2}} & Verb. & \cellcolor{yellow!55}0.825 & \cellcolor{yellow!35}0.026 & \cellcolor{yellow!35}0.119 & \cellcolor{yellow!55}0.810 & \cellcolor{yellow!35}0.041 & \cellcolor{yellow!35}0.143 & \cellcolor{yellow!55}0.888 & \cellcolor{yellow!35}0.050 & \cellcolor{yellow!35}0.153 & \cellcolor{yellow!35}0.626 & \cellcolor{yellow!35}0.035 & \cellcolor{orange!20}0.237 & \cellcolor{yellow!35}0.662 & \cellcolor{orange!20}0.054 & \cellcolor{orange!20}0.260 \\*
 &  & Logit & \cellcolor{yellow!55}0.813 & \cellcolor{yellow!55}0.007 & \cellcolor{yellow!55}0.104 & \cellcolor{yellow!55}0.800 & \cellcolor{yellow!55}0.010 & \cellcolor{yellow!35}0.124 & \cellcolor{orange!20}0.528 & \cellcolor{orange!65}0.913 & \cellcolor{orange!65}0.954 & \cellcolor{yellow!55}0.749 & \cellcolor{orange!20}0.061 & \cellcolor{orange!20}0.262 & \cellcolor{orange!35}0.490 & \cellcolor{orange!20}0.133 & \cellcolor{orange!35}0.377 \\*
\cmidrule(lr){2-18}
 & \multirow{2}{*}{\texttt{munin1}} & Verb. & \cellcolor{orange!35}0.436 & \cellcolor{orange!50}0.370 & \cellcolor{orange!35}0.404 & \cellcolor{yellow!35}0.683 & \cellcolor{orange!20}0.074 & \cellcolor{yellow!35}0.166 & \cellcolor{yellow!55}0.872 & \cellcolor{yellow!35}0.029 & \cellcolor{yellow!35}0.136 & \cellcolor{orange!35}0.480 & \cellcolor{yellow!35}0.020 & \cellcolor{orange!20}0.226 & \cellcolor{orange!20}0.523 & \cellcolor{orange!20}0.125 & \cellcolor{orange!35}0.370 \\*
 &  & Logit & \cellcolor{yellow!55}0.905 & \cellcolor{yellow!55}0.000 & \cellcolor{yellow!55}0.015 & \cellcolor{yellow!55}0.875 & \cellcolor{yellow!55}0.003 & \cellcolor{yellow!55}0.075 & \cellcolor{orange!50}0.272 & \cellcolor{orange!65}0.895 & \cellcolor{orange!65}0.944 & \cellcolor{yellow!35}0.721 & \cellcolor{yellow!35}0.023 & \cellcolor{yellow!35}0.201 & \cellcolor{orange!50}0.332 & \cellcolor{orange!35}0.278 & \cellcolor{orange!50}0.515 \\
\specialrule{0.10em}{0.10em}{0.10em}
\multirow{12}{*}{\rotatebox[origin=c]{90}{\parbox{1.8cm}{\centering Phi-4-Mini-\\Instruct}}} & \multirow{2}{*}{\texttt{asiam}} & Verb. & \cellcolor{orange!50}0.367 & \cellcolor{orange!50}0.410 & \cellcolor{orange!50}0.560 & \cellcolor{orange!35}0.431 & \cellcolor{orange!50}0.343 & \cellcolor{orange!50}0.464 & \cellcolor{yellow!35}0.605 & \cellcolor{orange!20}0.136 & \cellcolor{orange!20}0.242 & \cellcolor{orange!50}0.332 & \cellcolor{orange!35}0.183 & \cellcolor{orange!35}0.359 & \cellcolor{orange!20}0.522 & \cellcolor{orange!20}0.125 & \cellcolor{orange!20}0.271 \\*
 &  & Logit & \cellcolor{yellow!35}0.746 & \cellcolor{yellow!55}0.003 & \cellcolor{yellow!35}0.138 & \cellcolor{yellow!55}0.882 & \cellcolor{yellow!35}0.028 & \cellcolor{yellow!55}0.109 & \cellcolor{yellow!55}0.890 & \cellcolor{yellow!35}0.026 & \cellcolor{yellow!35}0.122 & \cellcolor{yellow!55}0.784 & \cellcolor{yellow!55}0.012 & \cellcolor{yellow!55}0.114 & \cellcolor{yellow!55}0.840 & \cellcolor{yellow!35}0.021 & \cellcolor{yellow!35}0.124 \\*
\cmidrule(lr){2-18}
 & \multirow{2}{*}{\texttt{river}} & Verb. & \cellcolor{orange!50}0.315 & \cellcolor{orange!50}0.386 & \cellcolor{orange!50}0.516 & \cellcolor{orange!50}0.287 & \cellcolor{orange!50}0.424 & \cellcolor{orange!50}0.548 & \cellcolor{orange!35}0.381 & \cellcolor{orange!20}0.121 & \cellcolor{orange!20}0.297 & \cellcolor{orange!65}0.212 & \cellcolor{orange!35}0.184 & \cellcolor{orange!35}0.348 & \cellcolor{orange!50}0.356 & \cellcolor{orange!20}0.058 & \cellcolor{orange!20}0.237 \\*
 &  & Logit & \cellcolor{yellow!35}0.594 & \cellcolor{yellow!55}0.009 & \cellcolor{yellow!55}0.103 & \cellcolor{yellow!35}0.684 & \cellcolor{yellow!55}0.010 & \cellcolor{yellow!55}0.099 & \cellcolor{yellow!35}0.742 & \cellcolor{yellow!55}0.003 & \cellcolor{yellow!55}0.096 & \cellcolor{yellow!55}0.792 & \cellcolor{yellow!55}0.003 & \cellcolor{yellow!35}0.141 & \cellcolor{yellow!55}0.784 & \cellcolor{yellow!55}0.007 & \cellcolor{yellow!35}0.122 \\*
\cmidrule(lr){2-18}
 & \multirow{2}{*}{\texttt{covid}} & Verb. & \cellcolor{orange!50}0.365 & \cellcolor{orange!50}0.399 & \cellcolor{orange!50}0.489 & \cellcolor{orange!50}0.340 & \cellcolor{orange!50}0.432 & \cellcolor{orange!50}0.523 & \cellcolor{orange!35}0.467 & \cellcolor{orange!20}0.077 & \cellcolor{yellow!35}0.180 & \cellcolor{orange!50}0.348 & \cellcolor{orange!35}0.163 & \cellcolor{orange!20}0.283 & \cellcolor{orange!35}0.411 & \cellcolor{orange!20}0.101 & \cellcolor{yellow!35}0.205 \\*
 &  & Logit & \cellcolor{yellow!35}0.703 & \cellcolor{yellow!55}0.002 & \cellcolor{yellow!55}0.059 & \cellcolor{yellow!55}0.816 & \cellcolor{yellow!55}0.003 & \cellcolor{yellow!55}0.057 & \cellcolor{yellow!55}0.838 & \cellcolor{yellow!55}0.003 & \cellcolor{yellow!55}0.058 & \cellcolor{yellow!35}0.718 & \cellcolor{yellow!55}0.005 & \cellcolor{yellow!55}0.063 & \cellcolor{yellow!35}0.682 & \cellcolor{yellow!55}0.003 & \cellcolor{yellow!55}0.062 \\*
\cmidrule(lr){2-18}
 & \multirow{2}{*}{\texttt{coal}} & Verb. & \cellcolor{orange!50}0.289 & \cellcolor{orange!65}0.565 & \cellcolor{orange!65}0.624 & \cellcolor{orange!50}0.243 & \cellcolor{orange!65}0.520 & \cellcolor{orange!65}0.590 & \cellcolor{orange!35}0.435 & \cellcolor{orange!20}0.104 & \cellcolor{yellow!35}0.200 & \cellcolor{orange!65}0.211 & \cellcolor{orange!35}0.200 & \cellcolor{orange!20}0.314 & \cellcolor{orange!50}0.269 & \cellcolor{orange!20}0.093 & \cellcolor{orange!20}0.216 \\*
 &  & Logit & \cellcolor{orange!20}0.550 & \cellcolor{yellow!55}0.001 & \cellcolor{yellow!55}0.026 & \cellcolor{yellow!55}0.790 & \cellcolor{yellow!55}0.004 & \cellcolor{yellow!55}0.030 & \cellcolor{yellow!35}0.705 & \cellcolor{yellow!55}0.001 & \cellcolor{yellow!55}0.027 & \cellcolor{yellow!55}0.765 & \cellcolor{yellow!35}0.017 & \cellcolor{yellow!55}0.048 & \cellcolor{yellow!35}0.696 & \cellcolor{yellow!55}0.009 & \cellcolor{yellow!55}0.035 \\*
\cmidrule(lr){2-18}
 & \multirow{2}{*}{\texttt{hepar2}} & Verb. & \cellcolor{orange!65}0.177 & \cellcolor{orange!65}0.626 & \cellcolor{orange!65}0.692 & \cellcolor{orange!65}0.211 & \cellcolor{orange!65}0.597 & \cellcolor{orange!65}0.658 & \cellcolor{orange!35}0.453 & \cellcolor{orange!20}0.083 & \cellcolor{yellow!35}0.174 & \cellcolor{orange!65}0.176 & \cellcolor{orange!35}0.237 & \cellcolor{orange!35}0.339 & \cellcolor{orange!50}0.342 & \cellcolor{orange!20}0.101 & \cellcolor{orange!20}0.227 \\*
 &  & Logit & \cellcolor{yellow!55}0.852 & \cellcolor{yellow!55}0.002 & \cellcolor{yellow!55}0.036 & \cellcolor{yellow!55}0.845 & \cellcolor{yellow!55}0.001 & \cellcolor{yellow!55}0.031 & \cellcolor{yellow!55}0.805 & \cellcolor{yellow!55}0.002 & \cellcolor{yellow!55}0.028 & \cellcolor{yellow!55}0.863 & \cellcolor{yellow!55}0.009 & \cellcolor{yellow!55}0.060 & \cellcolor{yellow!55}0.856 & \cellcolor{yellow!55}0.008 & \cellcolor{yellow!55}0.050 \\*
\cmidrule(lr){2-18}
 & \multirow{2}{*}{\texttt{munin1}} & Verb. & \cellcolor{orange!65}0.127 & \cellcolor{orange!65}0.573 & \cellcolor{orange!65}0.645 & \cellcolor{orange!65}0.185 & \cellcolor{orange!65}0.482 & \cellcolor{orange!50}0.573 & \cellcolor{orange!35}0.422 & \cellcolor{orange!20}0.067 & \cellcolor{yellow!35}0.190 & \cellcolor{orange!65}0.181 & \cellcolor{orange!20}0.126 & \cellcolor{orange!20}0.242 & \cellcolor{orange!50}0.279 & \cellcolor{orange!20}0.090 & \cellcolor{orange!20}0.218 \\*
 &  & Logit & \cellcolor{yellow!55}0.777 & \cellcolor{yellow!35}0.016 & \cellcolor{yellow!55}0.025 & \cellcolor{yellow!55}0.904 & \cellcolor{yellow!35}0.016 & \cellcolor{yellow!55}0.034 & \cellcolor{yellow!35}0.708 & \cellcolor{yellow!55}0.012 & \cellcolor{yellow!55}0.020 & \cellcolor{yellow!55}0.881 & \cellcolor{yellow!35}0.030 & \cellcolor{yellow!55}0.048 & \cellcolor{yellow!35}0.745 & \cellcolor{yellow!35}0.018 & \cellcolor{yellow!55}0.028 \\
\specialrule{0.10em}{0.10em}{0.10em}
\multirow{12}{*}{\rotatebox[origin=c]{90}{\parbox{1.8cm}{\centering Qwen3-4B-\\Instruct}}} & \multirow{2}{*}{\texttt{asiam}} & Verb. & \cellcolor{orange!65}0.166 & \cellcolor{orange!50}0.300 & \cellcolor{orange!35}0.415 & \cellcolor{orange!50}0.320 & \cellcolor{orange!35}0.226 & \cellcolor{orange!20}0.331 & \cellcolor{orange!50}0.354 & \cellcolor{orange!35}0.185 & \cellcolor{orange!20}0.309 & \cellcolor{orange!50}0.231 & \cellcolor{orange!20}0.126 & \cellcolor{orange!20}0.272 & \cellcolor{orange!50}0.358 & \cellcolor{orange!20}0.119 & \cellcolor{orange!20}0.329 \\*
 &  & Logit & \cellcolor{yellow!35}0.625 & \cellcolor{orange!65}0.514 & \cellcolor{orange!65}0.682 & \cellcolor{orange!20}0.548 & \cellcolor{orange!65}0.529 & \cellcolor{orange!65}0.717 & \cellcolor{yellow!35}0.625 & \cellcolor{orange!35}0.212 & \cellcolor{orange!35}0.431 & \cellcolor{orange!20}0.501 & \cellcolor{orange!20}0.097 & \cellcolor{orange!20}0.308 & \cellcolor{orange!20}0.529 & \cellcolor{orange!35}0.163 & \cellcolor{orange!35}0.402 \\*
\cmidrule(lr){2-18}
 & \multirow{2}{*}{\texttt{river}} & Verb. & \cellcolor{orange!65}0.154 & \cellcolor{orange!35}0.269 & \cellcolor{orange!35}0.370 & \cellcolor{orange!65}0.145 & \cellcolor{orange!50}0.286 & \cellcolor{orange!35}0.398 & \cellcolor{orange!65}0.208 & \cellcolor{orange!20}0.133 & \cellcolor{orange!20}0.299 & \cellcolor{orange!65}0.165 & \cellcolor{orange!20}0.102 & \cellcolor{orange!20}0.289 & \cellcolor{orange!65}0.185 & \cellcolor{orange!35}0.202 & \cellcolor{orange!35}0.362 \\*
 &  & Logit & \cellcolor{orange!20}0.524 & \cellcolor{orange!35}0.272 & \cellcolor{orange!50}0.518 & \cellcolor{orange!20}0.500 & \cellcolor{orange!35}0.281 & \cellcolor{orange!50}0.522 & \cellcolor{orange!20}0.547 & \cellcolor{orange!20}0.100 & \cellcolor{orange!20}0.318 & \cellcolor{orange!20}0.502 & \cellcolor{orange!20}0.143 & \cellcolor{orange!35}0.375 & \cellcolor{orange!35}0.499 & \cellcolor{orange!35}0.174 & \cellcolor{orange!35}0.416 \\*
\cmidrule(lr){2-18}
 & \multirow{2}{*}{\texttt{covid}} & Verb. & \cellcolor{orange!35}0.386 & \cellcolor{orange!50}0.400 & \cellcolor{orange!35}0.448 & \cellcolor{orange!35}0.384 & \cellcolor{orange!50}0.342 & \cellcolor{orange!35}0.390 & \cellcolor{orange!35}0.398 & \cellcolor{orange!20}0.094 & \cellcolor{yellow!35}0.168 & \cellcolor{orange!65}0.194 & \cellcolor{orange!20}0.061 & \cellcolor{yellow!35}0.151 & \cellcolor{orange!50}0.235 & \cellcolor{orange!20}0.061 & \cellcolor{orange!20}0.218 \\*
 &  & Logit & \cellcolor{orange!35}0.494 & \cellcolor{orange!50}0.435 & \cellcolor{orange!65}0.651 & \cellcolor{orange!20}0.531 & \cellcolor{orange!35}0.249 & \cellcolor{orange!50}0.495 & \cellcolor{orange!20}0.592 & \cellcolor{orange!20}0.077 & \cellcolor{orange!20}0.304 & \cellcolor{orange!20}0.573 & \cellcolor{yellow!55}0.014 & \cellcolor{yellow!55}0.111 & \cellcolor{orange!20}0.566 & \cellcolor{orange!20}0.051 & \cellcolor{orange!20}0.222 \\*
\cmidrule(lr){2-18}
 & \multirow{2}{*}{\texttt{coal}} & Verb. & \cellcolor{orange!35}0.425 & \cellcolor{orange!35}0.273 & \cellcolor{orange!20}0.299 & \cellcolor{orange!65}0.101 & \cellcolor{orange!35}0.253 & \cellcolor{orange!20}0.326 & \cellcolor{orange!35}0.417 & \cellcolor{yellow!35}0.032 & \cellcolor{yellow!35}0.190 & \cellcolor{orange!65}0.175 & \cellcolor{orange!20}0.057 & \cellcolor{yellow!35}0.185 & \cellcolor{orange!35}0.419 & \cellcolor{orange!20}0.115 & \cellcolor{orange!20}0.320 \\*
 &  & Logit & \cellcolor{orange!20}0.500 & \cellcolor{orange!65}0.948 & \cellcolor{orange!65}0.974 & \cellcolor{orange!20}0.514 & \cellcolor{orange!35}0.187 & \cellcolor{orange!35}0.433 & \cellcolor{yellow!35}0.609 & \cellcolor{yellow!35}0.038 & \cellcolor{yellow!35}0.204 & \cellcolor{orange!20}0.569 & \cellcolor{yellow!35}0.029 & \cellcolor{yellow!35}0.171 & \cellcolor{orange!20}0.528 & \cellcolor{orange!20}0.147 & \cellcolor{orange!35}0.386 \\*
\cmidrule(lr){2-18}
 & \multirow{2}{*}{\texttt{hepar2}} & Verb. & \cellcolor{orange!65}0.080 & \cellcolor{orange!50}0.383 & \cellcolor{orange!35}0.454 & \cellcolor{orange!65}0.099 & \cellcolor{orange!50}0.286 & \cellcolor{orange!35}0.382 & \cellcolor{orange!50}0.265 & \cellcolor{orange!20}0.126 & \cellcolor{orange!20}0.297 & \cellcolor{orange!65}0.128 & \cellcolor{orange!20}0.134 & \cellcolor{orange!20}0.228 & \cellcolor{orange!65}0.230 & \cellcolor{orange!20}0.118 & \cellcolor{orange!20}0.288 \\*
 &  & Logit & \cellcolor{orange!35}0.454 & \cellcolor{orange!50}0.418 & \cellcolor{orange!65}0.639 & \cellcolor{orange!20}0.512 & \cellcolor{orange!35}0.169 & \cellcolor{orange!35}0.411 & \cellcolor{orange!20}0.524 & \cellcolor{orange!35}0.159 & \cellcolor{orange!35}0.403 & \cellcolor{orange!20}0.531 & \cellcolor{orange!20}0.054 & \cellcolor{orange!20}0.236 & \cellcolor{orange!20}0.526 & \cellcolor{orange!20}0.107 & \cellcolor{orange!20}0.331 \\*
\cmidrule(lr){2-18}
 & \multirow{2}{*}{\texttt{munin1}} & Verb. & \cellcolor{orange!65}0.118 & \cellcolor{orange!50}0.444 & \cellcolor{orange!35}0.452 & \cellcolor{orange!65}0.029 & \cellcolor{orange!50}0.379 & \cellcolor{orange!35}0.408 & \cellcolor{orange!50}0.269 & \cellcolor{orange!20}0.083 & \cellcolor{yellow!35}0.200 & \cellcolor{orange!65}0.085 & \cellcolor{orange!35}0.225 & \cellcolor{orange!20}0.306 & \cellcolor{orange!35}0.403 & \cellcolor{orange!35}0.269 & \cellcolor{orange!50}0.481 \\*
 &  & Logit & \cellcolor{orange!20}0.505 & \cellcolor{orange!65}0.971 & \cellcolor{orange!65}0.985 & \cellcolor{orange!35}0.455 & \cellcolor{orange!65}0.629 & \cellcolor{orange!65}0.783 & \cellcolor{orange!20}0.508 & \cellcolor{orange!35}0.224 & \cellcolor{orange!50}0.471 & \cellcolor{orange!20}0.517 & \cellcolor{orange!20}0.101 & \cellcolor{orange!20}0.322 & \cellcolor{orange!35}0.479 & \cellcolor{orange!50}0.346 & \cellcolor{orange!65}0.583 \\
\specialrule{0.10em}{0.10em}{0.10em}
\pagebreak[4]
\multirow{12}{*}{\rotatebox[origin=c]{90}{\parbox{1.8cm}{\centering Qwen3-8B-\\Instruct}}} & \multirow{2}{*}{\texttt{asiam}} & Verb. & \cellcolor{orange!50}0.346 & \cellcolor{orange!50}0.417 & \cellcolor{orange!50}0.499 & \cellcolor{orange!50}0.247 & \cellcolor{orange!50}0.467 & \cellcolor{orange!50}0.523 & \cellcolor{yellow!35}0.658 & \cellcolor{orange!20}0.086 & \cellcolor{yellow!35}0.171 & \cellcolor{orange!50}0.255 & \cellcolor{orange!35}0.160 & \cellcolor{orange!20}0.324 & \cellcolor{orange!50}0.243 & \cellcolor{orange!20}0.114 & \cellcolor{orange!20}0.266 \\*
 &  & Logit & \cellcolor{orange!20}0.500 & \cellcolor{orange!65}0.655 & \cellcolor{orange!65}0.810 & \cellcolor{orange!20}0.500 & \cellcolor{orange!65}0.655 & \cellcolor{orange!65}0.810 & \cellcolor{orange!20}0.500 & \cellcolor{orange!65}0.655 & \cellcolor{orange!65}0.810 & \cellcolor{orange!20}0.515 & \cellcolor{orange!35}0.206 & \cellcolor{orange!35}0.439 & \cellcolor{orange!35}0.481 & \cellcolor{orange!20}0.068 & \cellcolor{orange!20}0.240 \\*
\cmidrule(lr){2-18}
 & \multirow{2}{*}{\texttt{river}} & Verb. & \cellcolor{orange!50}0.249 & \cellcolor{orange!65}0.481 & \cellcolor{orange!65}0.577 & \cellcolor{orange!65}0.179 & \cellcolor{orange!50}0.436 & \cellcolor{orange!50}0.532 & \cellcolor{orange!50}0.325 & \cellcolor{orange!20}0.076 & \cellcolor{orange!20}0.240 & \cellcolor{orange!65}0.186 & \cellcolor{orange!35}0.232 & \cellcolor{orange!35}0.416 & \cellcolor{orange!65}0.143 & \cellcolor{orange!50}0.316 & \cellcolor{orange!35}0.436 \\*
 &  & Logit & \cellcolor{orange!20}0.500 & \cellcolor{orange!65}0.776 & \cellcolor{orange!65}0.881 & \cellcolor{orange!20}0.500 & \cellcolor{orange!65}0.768 & \cellcolor{orange!65}0.876 & \cellcolor{orange!20}0.500 & \cellcolor{orange!65}0.776 & \cellcolor{orange!65}0.881 & \cellcolor{orange!35}0.438 & \cellcolor{orange!35}0.234 & \cellcolor{orange!50}0.483 & \cellcolor{orange!20}0.513 & \cellcolor{orange!35}0.222 & \cellcolor{orange!50}0.466 \\*
\cmidrule(lr){2-18}
 & \multirow{2}{*}{\texttt{covid}} & Verb. & \cellcolor{orange!50}0.266 & \cellcolor{orange!65}0.611 & \cellcolor{orange!65}0.663 & \cellcolor{orange!65}0.198 & \cellcolor{orange!65}0.669 & \cellcolor{orange!65}0.715 & \cellcolor{orange!20}0.554 & \cellcolor{yellow!55}0.013 & \cellcolor{yellow!55}0.069 & \cellcolor{orange!65}0.165 & \cellcolor{orange!20}0.115 & \cellcolor{orange!20}0.229 & \cellcolor{orange!65}0.128 & \cellcolor{orange!20}0.126 & \cellcolor{orange!20}0.220 \\*
 &  & Logit & \cellcolor{orange!35}0.497 & \cellcolor{orange!65}0.615 & \cellcolor{orange!65}0.782 & \cellcolor{orange!20}0.501 & \cellcolor{orange!65}0.843 & \cellcolor{orange!65}0.918 & \cellcolor{orange!20}0.500 & \cellcolor{orange!65}0.868 & \cellcolor{orange!65}0.932 & \cellcolor{yellow!35}0.604 & \cellcolor{orange!20}0.062 & \cellcolor{orange!20}0.242 & \cellcolor{orange!20}0.574 & \cellcolor{yellow!35}0.049 & \cellcolor{orange!20}0.220 \\*
\cmidrule(lr){2-18}
 & \multirow{2}{*}{\texttt{coal}} & Verb. & \cellcolor{orange!20}0.546 & \cellcolor{orange!65}0.478 & \cellcolor{orange!50}0.503 & \cellcolor{orange!65}0.112 & \cellcolor{orange!65}0.552 & \cellcolor{orange!65}0.576 & \cellcolor{orange!65}0.180 & \cellcolor{orange!20}0.051 & \cellcolor{yellow!35}0.118 & \cellcolor{orange!65}0.069 & \cellcolor{orange!50}0.374 & \cellcolor{orange!35}0.452 & \cellcolor{orange!65}0.041 & \cellcolor{orange!50}0.375 & \cellcolor{orange!35}0.411 \\*
 &  & Logit & \cellcolor{orange!20}0.500 & \cellcolor{orange!65}0.948 & \cellcolor{orange!65}0.974 & \cellcolor{orange!20}0.500 & \cellcolor{orange!65}0.947 & \cellcolor{orange!65}0.973 & \cellcolor{orange!20}0.500 & \cellcolor{orange!65}0.948 & \cellcolor{orange!65}0.974 & \cellcolor{orange!35}0.496 & \cellcolor{orange!50}0.294 & \cellcolor{orange!50}0.538 & \cellcolor{orange!20}0.532 & \cellcolor{orange!35}0.171 & \cellcolor{orange!35}0.415 \\*
\cmidrule(lr){2-18}
 & \multirow{2}{*}{\texttt{hepar2}} & Verb. & \cellcolor{orange!65}0.047 & \cellcolor{orange!65}0.585 & \cellcolor{orange!65}0.614 & \cellcolor{orange!65}0.048 & \cellcolor{orange!65}0.588 & \cellcolor{orange!65}0.624 & \cellcolor{orange!35}0.417 & \cellcolor{yellow!35}0.034 & \cellcolor{yellow!35}0.156 & \cellcolor{orange!65}0.098 & \cellcolor{orange!35}0.250 & \cellcolor{orange!35}0.345 & \cellcolor{orange!65}0.054 & \cellcolor{orange!50}0.294 & \cellcolor{orange!35}0.346 \\*
 &  & Logit & \cellcolor{orange!35}0.498 & \cellcolor{orange!65}0.909 & \cellcolor{orange!65}0.953 & \cellcolor{orange!35}0.497 & \cellcolor{orange!65}0.934 & \cellcolor{orange!65}0.966 & \cellcolor{orange!20}0.500 & \cellcolor{orange!65}0.950 & \cellcolor{orange!65}0.975 & \cellcolor{orange!20}0.522 & \cellcolor{orange!20}0.152 & \cellcolor{orange!35}0.392 & \cellcolor{orange!20}0.524 & \cellcolor{orange!20}0.125 & \cellcolor{orange!35}0.357 \\*
\cmidrule(lr){2-18}
 & \multirow{2}{*}{\texttt{munin1}} & Verb. & \cellcolor{orange!65}0.132 & \cellcolor{orange!65}0.563 & \cellcolor{orange!50}0.571 & \cellcolor{orange!65}0.040 & \cellcolor{orange!65}0.561 & \cellcolor{orange!65}0.576 & \cellcolor{orange!50}0.240 & \cellcolor{yellow!35}0.037 & \cellcolor{yellow!55}0.116 & \cellcolor{orange!65}0.141 & \cellcolor{orange!50}0.381 & \cellcolor{orange!50}0.510 & \cellcolor{orange!65}0.025 & \cellcolor{orange!65}0.595 & \cellcolor{orange!65}0.621 \\*
 &  & Logit & \cellcolor{orange!20}0.500 & \cellcolor{orange!65}0.984 & \cellcolor{orange!65}0.992 & \cellcolor{orange!35}0.498 & \cellcolor{orange!65}0.983 & \cellcolor{orange!65}0.992 & \cellcolor{orange!20}0.500 & \cellcolor{orange!65}0.984 & \cellcolor{orange!65}0.992 & \cellcolor{orange!35}0.442 & \cellcolor{orange!65}0.468 & \cellcolor{orange!65}0.671 & \cellcolor{orange!35}0.421 & \cellcolor{orange!65}0.485 & \cellcolor{orange!65}0.680 \\
\end{xltabular}
\endgroup
\end{landscape}

\begin{table}[H]
\centering
\resizebox{1.1\textwidth}{!}{%
\begin{tabular}{llccclccc}
\toprule
 & \multicolumn{4}{c}{\textbf{Small LLMs}} & \multicolumn{4}{c}{\textbf{Large LLMs}} \\
\cmidrule(lr){2-5} \cmidrule(lr){6-9}
\textbf{Dataset} & \textbf{Model} & \textbf{AUROC $\uparrow$} & \textbf{ECE $\downarrow$} & \textbf{Brier $\downarrow$} & \textbf{Model} & \textbf{AUROC $\uparrow$} & \textbf{ECE $\downarrow$} & \textbf{Brier $\downarrow$} \\
\midrule
\multirow{7}{*}{\rotatebox[origin=c]{90}{\texttt{asiam}}} & Gemma-4-E4B-IT & \textbf{0.845} & 0.005 & \textbf{0.110} & Qwen3-32B-Instruct & 0.747 & 0.002 & 0.077 \\
 & Llama-3.1-8B-Instruct & 0.689 & 0.004 & 0.158 & Gemma-4-31B-IT & 0.774 & \textbf{0.001} & \textbf{0.056} \\
 & Ministral-8B-Instruct-2410 & 0.540 & 0.228 & 0.428 & Qwen2.5-72B-Instruct & 0.822 & 0.006 & 0.065 \\
 & Phi-4 & 0.773 & \textbf{0.002} & 0.118 & Llama-3.3-70B-Instruct & \textbf{0.863} & 0.014 & 0.094 \\
 & Phi-4-Mini-Instruct & 0.779 & 0.010 & 0.138 & Llama-3.1-70B-Instruct & 0.753 & 0.010 & 0.148 \\
 & Qwen3-4B-Instruct & 0.588 & 0.046 & 0.240 &  &  &  &  \\
 & Qwen3-8B-Instruct & 0.774 & 0.018 & 0.151 &  &  &  &  \\
\midrule
\multirow{7}{*}{\rotatebox[origin=c]{90}{\texttt{river}}} & Gemma-4-E4B-IT & 0.616 & 0.027 & 0.234 & Qwen3-32B-Instruct & 0.642 & 0.057 & 0.252 \\
 & Llama-3.1-8B-Instruct & 0.628 & \textbf{0.006} & 0.212 & Gemma-4-31B-IT & 0.527 & 0.071 & 0.267 \\
 & Ministral-8B-Instruct-2410 & 0.403 & 0.169 & 0.392 & Qwen2.5-72B-Instruct & \textbf{0.806} & 0.017 & \textbf{0.161} \\
 & Phi-4 & \textbf{0.698} & 0.023 & 0.197 & Llama-3.3-70B-Instruct & 0.777 & \textbf{0.009} & 0.162 \\
 & Phi-4-Mini-Instruct & 0.647 & 0.012 & \textbf{0.125} & Llama-3.1-70B-Instruct & 0.740 & 0.023 & 0.202 \\
 & Qwen3-4B-Instruct & 0.555 & 0.084 & 0.306 &  &  &  &  \\
 & Qwen3-8B-Instruct & 0.574 & 0.038 & 0.232 &  &  &  &  \\
\midrule
\multirow{7}{*}{\rotatebox[origin=c]{90}{\texttt{covid}}} & Gemma-4-E4B-IT & 0.706 & 0.011 & 0.073 & Qwen3-32B-Instruct & 0.767 & 0.002 & 0.056 \\
 & Llama-3.1-8B-Instruct & \textbf{0.799} & 0.005 & 0.098 & Gemma-4-31B-IT & 0.734 & 0.008 & 0.089 \\
 & Ministral-8B-Instruct-2410 & 0.440 & 0.126 & 0.359 & Qwen2.5-72B-Instruct & 0.729 & 0.004 & \textbf{0.054} \\
 & Phi-4 & 0.782 & \textbf{0.002} & 0.057 & Llama-3.3-70B-Instruct & 0.882 & 0.004 & 0.060 \\
 & Phi-4-Mini-Instruct & 0.681 & 0.004 & 0.071 & Llama-3.1-70B-Instruct & \textbf{0.888} & \textbf{0.001} & 0.066 \\
 & Qwen3-4B-Instruct & 0.784 & 0.004 & \textbf{0.056} &  &  &  &  \\
 & Qwen3-8B-Instruct & 0.788 & 0.007 & 0.066 &  &  &  &  \\
\midrule
\multirow{7}{*}{\rotatebox[origin=c]{90}{\texttt{coal}}} & Gemma-4-E4B-IT & 0.704 & 0.023 & 0.106 & Qwen3-32B-Instruct & 0.818 & 0.022 & 0.133 \\
 & Llama-3.1-8B-Instruct & 0.659 & \textbf{0.001} & 0.176 & Gemma-4-31B-IT & 0.855 & 0.062 & 0.166 \\
 & Ministral-8B-Instruct-2410 & 0.360 & 0.229 & 0.428 & Qwen2.5-72B-Instruct & \textbf{0.857} & 0.003 & \textbf{0.100} \\
 & Phi-4 & 0.766 & 0.018 & 0.174 & Llama-3.3-70B-Instruct & 0.828 & 0.006 & 0.123 \\
 & Phi-4-Mini-Instruct & 0.709 & 0.014 & \textbf{0.049} & Llama-3.1-70B-Instruct & 0.806 & \textbf{0.002} & 0.118 \\
 & Qwen3-4B-Instruct & \textbf{0.817} & 0.002 & 0.106 &  &  &  &  \\
 & Qwen3-8B-Instruct & 0.591 & 0.042 & 0.120 &  &  &  &  \\
\midrule
\multirow{7}{*}{\rotatebox[origin=c]{90}{\texttt{hepar2}}} & Gemma-4-E4B-IT & 0.691 & 0.008 & 0.103 & Qwen3-32B-Instruct & 0.721 & 0.007 & 0.119 \\
 & Llama-3.1-8B-Instruct & 0.709 & \textbf{0.002} & 0.151 & Gemma-4-31B-IT & 0.638 & 0.021 & 0.163 \\
 & Ministral-8B-Instruct-2410 & 0.466 & 0.091 & 0.337 & Qwen2.5-72B-Instruct & 0.775 & 0.004 & \textbf{0.099} \\
 & Phi-4 & 0.683 & 0.010 & 0.159 & Llama-3.3-70B-Instruct & \textbf{0.821} & \textbf{0.001} & 0.108 \\
 & Phi-4-Mini-Instruct & \textbf{0.776} & 0.008 & \textbf{0.052} & Llama-3.1-70B-Instruct & 0.762 & 0.004 & 0.148 \\
 & Qwen3-4B-Instruct & 0.677 & 0.030 & 0.204 &  &  &  &  \\
 & Qwen3-8B-Instruct & 0.629 & 0.018 & 0.168 &  &  &  &  \\
\midrule
\multirow{7}{*}{\rotatebox[origin=c]{90}{\texttt{munin1}}} & Gemma-4-E4B-IT & 0.622 & 0.029 & 0.142 & Qwen3-32B-Instruct & 0.813 & 0.003 & 0.107 \\
 & Llama-3.1-8B-Instruct & 0.551 & 0.021 & 0.262 & Gemma-4-31B-IT & 0.718 & 0.057 & 0.233 \\
 & Ministral-8B-Instruct-2410 & 0.423 & 0.141 & 0.371 & Qwen2.5-72B-Instruct & \textbf{0.827} & 0.002 & 0.125 \\
 & Phi-4 & 0.772 & \textbf{0.001} & 0.120 & Llama-3.3-70B-Instruct & 0.764 & 0.045 & \textbf{0.101} \\
 & Phi-4-Mini-Instruct & \textbf{0.846} & 0.017 & \textbf{0.055} & Llama-3.1-70B-Instruct & 0.699 & \textbf{0.002} & 0.170 \\
 & Qwen3-4B-Instruct & 0.725 & 0.004 & 0.136 &  &  &  &  \\
 & Qwen3-8B-Instruct & 0.511 & 0.055 & 0.148 &  &  &  &  \\
\bottomrule
\end{tabular}%
}
\caption{Cross-prompt agreement calibration results for small (left) and large (right) LLMs across all datasets. This method combines predictions across the five prompt styles for each fixed model and dataset. Bold values indicate the best model within each dataset for each metric (small and large LLMs evaluated independently).}
\label{tab:calib_cross_prompt_combined}
\end{table}

\begin{landscape}
\begingroup
\scriptsize
\setlength{\tabcolsep}{3pt}
\renewcommand{\arraystretch}{1.05}
\begin{xltabular}{\linewidth}{@{}lll *{15}{>{\centering\arraybackslash}X}@{}}
\toprule
\multirow{2}{*}{\textbf{Model}} & \multirow{2}{*}{\textbf{Dataset}} & \multirow{2}{*}{\textbf{Method}} & \multicolumn{3}{c}{\textbf{Name-only}} & \multicolumn{3}{c}{\textbf{Metadata}} & \multicolumn{3}{c}{\textbf{CoT}} & \multicolumn{3}{c}{\textbf{Few-shot}} & \multicolumn{3}{c}{\textbf{Few-shot + CoT}} \\
\cmidrule(lr){4-6} \cmidrule(lr){7-9} \cmidrule(lr){10-12} \cmidrule(lr){13-15} \cmidrule(lr){16-18}
&  &  & AUROC $\uparrow$ & ECE $\downarrow$ & Brier $\downarrow$ & AUROC $\uparrow$ & ECE $\downarrow$ & Brier $\downarrow$ & AUROC $\uparrow$ & ECE $\downarrow$ & Brier $\downarrow$ & AUROC $\uparrow$ & ECE $\downarrow$ & Brier $\downarrow$ & AUROC $\uparrow$ & ECE $\downarrow$ & Brier $\downarrow$ \\
\midrule
\endfirsthead

\multicolumn{18}{l}{\emph{(continued from previous page)}}\\
\toprule
\multirow{2}{*}{\textbf{Model}} & \multirow{2}{*}{\textbf{Dataset}} & \multirow{2}{*}{\textbf{Method}} & \multicolumn{3}{c}{\textbf{Name-only}} & \multicolumn{3}{c}{\textbf{Metadata}} & \multicolumn{3}{c}{\textbf{CoT}} & \multicolumn{3}{c}{\textbf{Few-shot}} & \multicolumn{3}{c}{\textbf{Few-shot + CoT}} \\
\cmidrule(lr){4-6} \cmidrule(lr){7-9} \cmidrule(lr){10-12} \cmidrule(lr){13-15} \cmidrule(lr){16-18}
&  &  & AUROC $\uparrow$ & ECE $\downarrow$ & Brier $\downarrow$ & AUROC $\uparrow$ & ECE $\downarrow$ & Brier $\downarrow$ & AUROC $\uparrow$ & ECE $\downarrow$ & Brier $\downarrow$ & AUROC $\uparrow$ & ECE $\downarrow$ & Brier $\downarrow$ & AUROC $\uparrow$ & ECE $\downarrow$ & Brier $\downarrow$ \\
\midrule
\endhead

\bottomrule
\multicolumn{18}{r}{\emph{(continued on next page)}}\\
\endfoot

\bottomrule
\addlinespace[1.5em]
\caption{Prompt-specific calibration results for large LLMs on \texttt{asiam}, \texttt{river}, \texttt{covid}, \texttt{coal}, \texttt{hepar2}, and \texttt{munin1}. ECE is the 10-bin squared-gap measure defined in Section~\ref{subsec:evaluation_metrics}. Cell colors encode performance percentile within the full table (yellow~=~good, orange~=~bad), with intensity scaled separately per metric direction.}
\label{tab:calib_all_largemodels}\\
\label{tab:calib_prompt_large_models}\\
\endlastfoot

\multirow{12}{*}{\rotatebox[origin=c]{90}{\parbox{1.8cm}{\centering Qwen3-32B-\\Instruct}}} & \multirow{2}{*}{\texttt{asiam}} & Verb. & \cellcolor{orange!65}0.265 & \cellcolor{orange!65}0.525 & \cellcolor{orange!65}0.637 & \cellcolor{orange!65}0.211 & \cellcolor{orange!65}0.516 & \cellcolor{orange!50}0.603 & \cellcolor{orange!50}0.440 & \cellcolor{orange!20}0.036 & \cellcolor{yellow!55}0.097 & \cellcolor{orange!50}0.331 & \cellcolor{orange!20}0.038 & \cellcolor{yellow!35}0.154 & \cellcolor{orange!65}0.306 & \cellcolor{yellow!35}0.029 & \cellcolor{yellow!35}0.162 \\*
 &  & Logit & \cellcolor{yellow!35}0.615 & \cellcolor{orange!50}0.319 & \cellcolor{orange!50}0.531 & \cellcolor{yellow!35}0.676 & \cellcolor{orange!50}0.482 & \cellcolor{orange!65}0.654 & \cellcolor{yellow!55}0.749 & \cellcolor{yellow!35}0.023 & \cellcolor{yellow!35}0.125 & \cellcolor{yellow!35}0.667 & \cellcolor{orange!50}0.177 & \cellcolor{orange!35}0.384 & \cellcolor{yellow!55}0.711 & \cellcolor{orange!20}0.065 & \cellcolor{yellow!35}0.162 \\*
\cmidrule(lr){2-18}
 & \multirow{2}{*}{\texttt{river}} & Verb. & \cellcolor{orange!65}0.142 & \cellcolor{orange!50}0.512 & \cellcolor{orange!65}0.616 & \cellcolor{orange!65}0.161 & \cellcolor{orange!65}0.515 & \cellcolor{orange!50}0.614 & \cellcolor{orange!50}0.354 & \cellcolor{orange!20}0.055 & \cellcolor{orange!20}0.220 & \cellcolor{orange!65}0.213 & \cellcolor{orange!20}0.049 & \cellcolor{orange!20}0.246 & \cellcolor{orange!65}0.235 & \cellcolor{orange!35}0.085 & \cellcolor{orange!35}0.270 \\*
 &  & Logit & \cellcolor{yellow!35}0.675 & \cellcolor{orange!65}0.530 & \cellcolor{orange!65}0.688 & \cellcolor{yellow!55}0.790 & \cellcolor{orange!50}0.378 & \cellcolor{orange!50}0.544 & \cellcolor{orange!20}0.525 & \cellcolor{orange!50}0.218 & \cellcolor{orange!50}0.458 & \cellcolor{orange!20}0.565 & \cellcolor{orange!35}0.101 & \cellcolor{orange!35}0.319 & \cellcolor{orange!20}0.542 & \cellcolor{orange!35}0.080 & \cellcolor{orange!35}0.300 \\*
\cmidrule(lr){2-18}
 & \multirow{2}{*}{\texttt{covid}} & Verb. & \cellcolor{orange!50}0.376 & \cellcolor{orange!65}0.747 & \cellcolor{orange!65}0.806 & \cellcolor{orange!65}0.324 & \cellcolor{orange!65}0.731 & \cellcolor{orange!65}0.783 & \cellcolor{orange!20}0.571 & \cellcolor{orange!20}0.062 & \cellcolor{yellow!35}0.124 & \cellcolor{orange!35}0.500 & \cellcolor{orange!20}0.045 & \cellcolor{yellow!35}0.136 & \cellcolor{orange!50}0.368 & \cellcolor{yellow!35}0.025 & \cellcolor{yellow!35}0.146 \\*
 &  & Logit & \cellcolor{yellow!35}0.607 & \cellcolor{orange!65}0.730 & \cellcolor{orange!65}0.803 & \cellcolor{yellow!55}0.744 & \cellcolor{orange!65}0.633 & \cellcolor{orange!65}0.723 & \cellcolor{yellow!35}0.654 & \cellcolor{orange!35}0.090 & \cellcolor{orange!35}0.276 & \cellcolor{yellow!35}0.614 & \cellcolor{yellow!35}0.012 & \cellcolor{yellow!35}0.123 & \cellcolor{yellow!55}0.710 & \cellcolor{yellow!55}0.008 & \cellcolor{yellow!35}0.104 \\*
\cmidrule(lr){2-18}
 & \multirow{2}{*}{\texttt{coal}} & Verb. & \cellcolor{orange!35}0.503 & \cellcolor{orange!65}0.839 & \cellcolor{orange!65}0.865 & \cellcolor{orange!65}0.055 & \cellcolor{orange!65}0.699 & \cellcolor{orange!65}0.730 & \cellcolor{orange!65}0.154 & \cellcolor{orange!50}0.185 & \cellcolor{orange!35}0.284 & \cellcolor{orange!65}0.059 & \cellcolor{orange!35}0.167 & \cellcolor{orange!35}0.291 & \cellcolor{orange!65}0.113 & \cellcolor{orange!35}0.164 & \cellcolor{orange!35}0.295 \\*
 &  & Logit & \cellcolor{yellow!35}0.607 & \cellcolor{orange!65}0.722 & \cellcolor{orange!65}0.776 & \cellcolor{yellow!35}0.689 & \cellcolor{orange!65}0.630 & \cellcolor{orange!65}0.713 & \cellcolor{yellow!35}0.678 & \cellcolor{orange!20}0.039 & \cellcolor{orange!20}0.195 & \cellcolor{yellow!35}0.602 & \cellcolor{orange!35}0.121 & \cellcolor{orange!35}0.352 & \cellcolor{yellow!35}0.602 & \cellcolor{orange!35}0.080 & \cellcolor{orange!35}0.305 \\*
\cmidrule(lr){2-18}
 & \multirow{2}{*}{\texttt{hepar2}} & Verb. & \cellcolor{orange!65}0.105 & \cellcolor{orange!65}0.601 & \cellcolor{orange!65}0.664 & \cellcolor{orange!65}0.087 & \cellcolor{orange!65}0.610 & \cellcolor{orange!65}0.665 & \cellcolor{orange!50}0.380 & \cellcolor{orange!20}0.040 & \cellcolor{orange!20}0.165 & \cellcolor{orange!65}0.164 & \cellcolor{orange!35}0.080 & \cellcolor{orange!20}0.180 & \cellcolor{orange!65}0.104 & \cellcolor{orange!35}0.107 & \cellcolor{orange!20}0.169 \\*
 &  & Logit & \cellcolor{yellow!35}0.630 & \cellcolor{orange!65}0.675 & \cellcolor{orange!65}0.775 & \cellcolor{yellow!35}0.667 & \cellcolor{orange!65}0.675 & \cellcolor{orange!65}0.778 & \cellcolor{orange!20}0.574 & \cellcolor{orange!35}0.090 & \cellcolor{orange!35}0.303 & \cellcolor{orange!20}0.538 & \cellcolor{yellow!35}0.026 & \cellcolor{orange!20}0.166 & \cellcolor{yellow!35}0.686 & \cellcolor{yellow!35}0.012 & \cellcolor{yellow!35}0.120 \\*
\cmidrule(lr){2-18}
 & \multirow{2}{*}{\texttt{munin1}} & Verb. & \cellcolor{orange!65}0.046 & \cellcolor{orange!65}0.848 & \cellcolor{orange!65}0.859 & \cellcolor{orange!65}0.033 & \cellcolor{orange!65}0.767 & \cellcolor{orange!65}0.796 & \cellcolor{orange!65}0.216 & \cellcolor{orange!35}0.123 & \cellcolor{orange!20}0.208 & \cellcolor{orange!65}0.018 & \cellcolor{orange!50}0.189 & \cellcolor{orange!20}0.248 & \cellcolor{orange!65}0.072 & \cellcolor{orange!35}0.149 & \cellcolor{orange!35}0.280 \\*
 &  & Logit & \cellcolor{yellow!55}0.818 & \cellcolor{orange!65}0.761 & \cellcolor{orange!65}0.807 & \cellcolor{yellow!55}0.704 & \cellcolor{orange!65}0.756 & \cellcolor{orange!65}0.815 & \cellcolor{orange!35}0.513 & \cellcolor{orange!50}0.185 & \cellcolor{orange!50}0.432 & \cellcolor{yellow!35}0.648 & \cellcolor{orange!20}0.032 & \cellcolor{orange!20}0.199 & \cellcolor{yellow!35}0.681 & \cellcolor{orange!20}0.047 & \cellcolor{orange!20}0.244 \\
\specialrule{0.10em}{0.10em}{0.10em}
\multirow{12}{*}{\rotatebox[origin=c]{90}{Gemma-4-31B-IT}} & \multirow{2}{*}{\texttt{asiam}} & Verb. & \cellcolor{orange!20}0.593 & \cellcolor{orange!50}0.241 & \cellcolor{orange!35}0.361 & \cellcolor{yellow!55}0.919 & \cellcolor{orange!20}0.058 & \cellcolor{yellow!55}0.100 & \cellcolor{yellow!55}0.833 & \cellcolor{yellow!55}0.003 & \cellcolor{yellow!55}0.065 & \cellcolor{yellow!55}0.900 & \cellcolor{yellow!55}0.001 & \cellcolor{yellow!55}0.040 & \cellcolor{yellow!55}0.928 & \cellcolor{yellow!35}0.026 & \cellcolor{yellow!55}0.076 \\*
 &  & Logit & \cellcolor{orange!20}0.544 & \cellcolor{orange!65}0.631 & \cellcolor{orange!65}0.782 & \cellcolor{orange!50}0.458 & \cellcolor{orange!50}0.245 & \cellcolor{orange!50}0.471 & \cellcolor{orange!35}0.500 & \cellcolor{orange!65}0.655 & \cellcolor{orange!65}0.810 & \cellcolor{orange!35}0.500 & \cellcolor{yellow!55}0.002 & \cellcolor{yellow!55}0.048 & \cellcolor{orange!35}0.500 & \cellcolor{orange!50}0.354 & \cellcolor{orange!50}0.595 \\*
\cmidrule(lr){2-18}
 & \multirow{2}{*}{\texttt{river}} & Verb. & \cellcolor{orange!50}0.403 & \cellcolor{orange!50}0.447 & \cellcolor{orange!50}0.537 & \cellcolor{yellow!55}0.703 & \cellcolor{orange!35}0.165 & \cellcolor{orange!35}0.319 & \cellcolor{yellow!55}0.833 & \cellcolor{orange!35}0.091 & \cellcolor{orange!35}0.261 & \cellcolor{yellow!55}0.787 & \cellcolor{orange!20}0.052 & \cellcolor{orange!20}0.212 & \cellcolor{yellow!55}0.833 & \cellcolor{orange!35}0.089 & \cellcolor{orange!35}0.261 \\*
 &  & Logit & \cellcolor{orange!50}0.455 & \cellcolor{orange!50}0.287 & \cellcolor{orange!50}0.498 & \cellcolor{orange!20}0.531 & \cellcolor{orange!35}0.138 & \cellcolor{orange!50}0.388 & \cellcolor{orange!35}0.500 & \cellcolor{orange!65}0.776 & \cellcolor{orange!65}0.881 & \cellcolor{orange!35}0.497 & \cellcolor{orange!35}0.079 & \cellcolor{orange!35}0.281 & \cellcolor{orange!35}0.487 & \cellcolor{orange!65}0.674 & \cellcolor{orange!65}0.819 \\*
\cmidrule(lr){2-18}
 & \multirow{2}{*}{\texttt{covid}} & Verb. & \cellcolor{yellow!55}0.709 & \cellcolor{orange!35}0.156 & \cellcolor{orange!20}0.235 & \cellcolor{yellow!55}0.811 & \cellcolor{orange!35}0.076 & \cellcolor{yellow!35}0.146 & \cellcolor{yellow!55}0.901 & \cellcolor{orange!20}0.036 & \cellcolor{yellow!35}0.110 & \cellcolor{yellow!55}0.843 & \cellcolor{yellow!35}0.011 & \cellcolor{yellow!55}0.076 & \cellcolor{yellow!55}0.913 & \cellcolor{yellow!35}0.024 & \cellcolor{yellow!35}0.102 \\*
 &  & Logit & \cellcolor{orange!50}0.410 & \cellcolor{orange!50}0.380 & \cellcolor{orange!50}0.571 & \cellcolor{orange!20}0.588 & \cellcolor{orange!20}0.032 & \cellcolor{orange!20}0.240 & \cellcolor{orange!35}0.500 & \cellcolor{orange!65}0.868 & \cellcolor{orange!65}0.932 & \cellcolor{orange!35}0.504 & \cellcolor{yellow!55}0.008 & \cellcolor{yellow!55}0.085 & \cellcolor{orange!35}0.495 & \cellcolor{orange!50}0.175 & \cellcolor{orange!50}0.414 \\*
\cmidrule(lr){2-18}
 & \multirow{2}{*}{\texttt{coal}} & Verb. & \cellcolor{orange!50}0.453 & \cellcolor{orange!50}0.481 & \cellcolor{orange!50}0.507 & \cellcolor{orange!50}0.392 & \cellcolor{orange!50}0.384 & \cellcolor{orange!50}0.507 & \cellcolor{yellow!55}0.877 & \cellcolor{orange!35}0.137 & \cellcolor{orange!35}0.255 & \cellcolor{yellow!55}0.715 & \cellcolor{yellow!35}0.016 & \cellcolor{orange!20}0.173 & \cellcolor{yellow!55}0.838 & \cellcolor{orange!35}0.117 & \cellcolor{orange!35}0.269 \\*
 &  & Logit & \cellcolor{orange!50}0.432 & \cellcolor{orange!65}0.921 & \cellcolor{orange!65}0.956 & \cellcolor{orange!50}0.356 & \cellcolor{orange!50}0.408 & \cellcolor{orange!50}0.592 & \cellcolor{orange!35}0.500 & \cellcolor{orange!65}0.948 & \cellcolor{orange!65}0.974 & \cellcolor{orange!20}0.526 & \cellcolor{orange!20}0.048 & \cellcolor{orange!20}0.219 & \cellcolor{orange!50}0.480 & \cellcolor{orange!65}0.807 & \cellcolor{orange!65}0.897 \\*
\cmidrule(lr){2-18}
 & \multirow{2}{*}{\texttt{hepar2}} & Verb. & \cellcolor{orange!50}0.372 & \cellcolor{orange!50}0.486 & \cellcolor{orange!50}0.560 & \cellcolor{orange!20}0.558 & \cellcolor{orange!50}0.292 & \cellcolor{orange!50}0.392 & \cellcolor{yellow!55}0.877 & \cellcolor{orange!35}0.073 & \cellcolor{orange!20}0.171 & \cellcolor{yellow!55}0.811 & \cellcolor{yellow!35}0.011 & \cellcolor{yellow!35}0.121 & \cellcolor{yellow!55}0.852 & \cellcolor{orange!20}0.064 & \cellcolor{orange!20}0.177 \\*
 &  & Logit & \cellcolor{orange!65}0.308 & \cellcolor{orange!65}0.849 & \cellcolor{orange!65}0.902 & \cellcolor{orange!50}0.348 & \cellcolor{orange!50}0.396 & \cellcolor{orange!50}0.577 & \cellcolor{orange!35}0.500 & \cellcolor{orange!65}0.950 & \cellcolor{orange!65}0.975 & \cellcolor{orange!20}0.521 & \cellcolor{yellow!35}0.024 & \cellcolor{yellow!35}0.156 & \cellcolor{orange!35}0.492 & \cellcolor{orange!65}0.645 & \cellcolor{orange!65}0.803 \\*
\cmidrule(lr){2-18}
 & \multirow{2}{*}{\texttt{munin1}} & Verb. & \cellcolor{orange!65}0.013 & \cellcolor{orange!65}0.845 & \cellcolor{orange!65}0.855 & \cellcolor{orange!65}0.257 & \cellcolor{orange!65}0.555 & \cellcolor{orange!65}0.652 & \cellcolor{yellow!55}0.871 & \cellcolor{orange!35}0.151 & \cellcolor{orange!35}0.287 & \cellcolor{orange!20}0.552 & \cellcolor{orange!20}0.031 & \cellcolor{orange!20}0.240 & \cellcolor{yellow!35}0.696 & \cellcolor{orange!35}0.094 & \cellcolor{orange!35}0.304 \\*
 &  & Logit & \cellcolor{orange!50}0.436 & \cellcolor{orange!65}0.969 & \cellcolor{orange!65}0.979 & \cellcolor{orange!50}0.463 & \cellcolor{orange!50}0.216 & \cellcolor{orange!50}0.465 & \cellcolor{orange!35}0.500 & \cellcolor{orange!65}0.984 & \cellcolor{orange!65}0.992 & \cellcolor{orange!20}0.554 & \cellcolor{orange!35}0.081 & \cellcolor{orange!35}0.288 & \cellcolor{orange!35}0.487 & \cellcolor{orange!50}0.280 & \cellcolor{orange!50}0.528 \\
\specialrule{0.10em}{0.10em}{0.10em}
\multirow{12}{*}{\rotatebox[origin=c]{90}{\parbox{1.8cm}{\centering Qwen2.5-72B-\\Instruct}}} & \multirow{2}{*}{\texttt{asiam}} & Verb. & \cellcolor{orange!35}0.516 & \cellcolor{yellow!55}0.002 & \cellcolor{yellow!35}0.107 & \cellcolor{yellow!35}0.632 & \cellcolor{orange!35}0.101 & \cellcolor{yellow!35}0.141 & \cellcolor{yellow!55}0.722 & \cellcolor{yellow!55}0.003 & \cellcolor{yellow!35}0.113 & \cellcolor{yellow!35}0.688 & \cellcolor{yellow!55}0.005 & \cellcolor{yellow!35}0.116 & \cellcolor{orange!50}0.368 & \cellcolor{orange!20}0.034 & \cellcolor{yellow!35}0.107 \\*
 &  & Logit & \cellcolor{orange!20}0.558 & \cellcolor{yellow!35}0.016 & \cellcolor{yellow!35}0.135 & \cellcolor{orange!35}0.500 & \cellcolor{yellow!55}0.002 & \cellcolor{yellow!55}0.048 & \cellcolor{orange!35}0.500 & \cellcolor{orange!65}0.655 & \cellcolor{orange!65}0.810 & \cellcolor{orange!20}0.571 & \cellcolor{orange!20}0.031 & \cellcolor{yellow!35}0.153 & \cellcolor{orange!50}0.408 & \cellcolor{orange!20}0.029 & \cellcolor{yellow!35}0.113 \\*
\cmidrule(lr){2-18}
 & \multirow{2}{*}{\texttt{river}} & Verb. & \cellcolor{yellow!35}0.662 & \cellcolor{yellow!55}0.001 & \cellcolor{yellow!35}0.124 & \cellcolor{yellow!35}0.670 & \cellcolor{yellow!35}0.012 & \cellcolor{yellow!35}0.136 & \cellcolor{yellow!35}0.614 & \cellcolor{orange!20}0.036 & \cellcolor{orange!35}0.254 & \cellcolor{orange!50}0.368 & \cellcolor{orange!20}0.038 & \cellcolor{orange!20}0.224 & \cellcolor{orange!50}0.360 & \cellcolor{orange!20}0.068 & \cellcolor{orange!35}0.299 \\*
 &  & Logit & \cellcolor{orange!20}0.538 & \cellcolor{yellow!35}0.021 & \cellcolor{yellow!35}0.133 & \cellcolor{orange!20}0.560 & \cellcolor{yellow!35}0.023 & \cellcolor{yellow!35}0.141 & \cellcolor{orange!35}0.500 & \cellcolor{orange!65}0.776 & \cellcolor{orange!65}0.881 & \cellcolor{orange!35}0.503 & \cellcolor{orange!20}0.069 & \cellcolor{orange!35}0.257 & \cellcolor{orange!35}0.506 & \cellcolor{orange!35}0.141 & \cellcolor{orange!35}0.379 \\*
\cmidrule(lr){2-18}
 & \multirow{2}{*}{\texttt{covid}} & Verb. & \cellcolor{yellow!55}0.714 & \cellcolor{yellow!35}0.022 & \cellcolor{yellow!55}0.066 & \cellcolor{yellow!55}0.765 & \cellcolor{yellow!35}0.012 & \cellcolor{yellow!55}0.059 & \cellcolor{yellow!35}0.664 & \cellcolor{yellow!55}0.004 & \cellcolor{yellow!55}0.057 & \cellcolor{yellow!55}0.776 & \cellcolor{yellow!55}0.002 & \cellcolor{yellow!55}0.090 & \cellcolor{yellow!35}0.699 & \cellcolor{yellow!55}0.003 & \cellcolor{yellow!35}0.120 \\*
 &  & Logit & \cellcolor{orange!20}0.526 & \cellcolor{yellow!55}0.004 & \cellcolor{yellow!55}0.049 & \cellcolor{orange!20}0.550 & \cellcolor{yellow!55}0.004 & \cellcolor{yellow!55}0.049 & \cellcolor{orange!35}0.500 & \cellcolor{orange!65}0.868 & \cellcolor{orange!65}0.932 & \cellcolor{orange!35}0.520 & \cellcolor{yellow!35}0.011 & \cellcolor{yellow!55}0.097 & \cellcolor{yellow!35}0.609 & \cellcolor{yellow!35}0.016 & \cellcolor{yellow!35}0.133 \\*
\cmidrule(lr){2-18}
 & \multirow{2}{*}{\texttt{coal}} & Verb. & \cellcolor{orange!50}0.433 & \cellcolor{orange!20}0.070 & \cellcolor{yellow!55}0.096 & \cellcolor{orange!50}0.434 & \cellcolor{yellow!35}0.026 & \cellcolor{yellow!55}0.095 & \cellcolor{yellow!35}0.600 & \cellcolor{yellow!55}0.003 & \cellcolor{yellow!35}0.154 & \cellcolor{orange!50}0.349 & \cellcolor{orange!20}0.033 & \cellcolor{orange!20}0.198 & \cellcolor{orange!65}0.318 & \cellcolor{orange!20}0.062 & \cellcolor{orange!35}0.280 \\*
 &  & Logit & \cellcolor{orange!35}0.500 & \cellcolor{yellow!55}0.001 & \cellcolor{yellow!55}0.026 & \cellcolor{orange!20}0.536 & \cellcolor{yellow!55}0.003 & \cellcolor{yellow!55}0.054 & \cellcolor{orange!35}0.500 & \cellcolor{orange!65}0.948 & \cellcolor{orange!65}0.974 & \cellcolor{orange!20}0.557 & \cellcolor{orange!20}0.048 & \cellcolor{orange!20}0.222 & \cellcolor{orange!20}0.557 & \cellcolor{orange!35}0.100 & \cellcolor{orange!35}0.325 \\*
\cmidrule(lr){2-18}
 & \multirow{2}{*}{\texttt{hepar2}} & Verb. & \cellcolor{orange!35}0.507 & \cellcolor{yellow!35}0.015 & \cellcolor{yellow!55}0.075 & \cellcolor{orange!50}0.474 & \cellcolor{yellow!35}0.025 & \cellcolor{yellow!55}0.095 & \cellcolor{yellow!35}0.649 & \cellcolor{yellow!55}0.001 & \cellcolor{yellow!35}0.125 & \cellcolor{yellow!35}0.620 & \cellcolor{yellow!55}0.004 & \cellcolor{orange!20}0.164 & \cellcolor{orange!35}0.493 & \cellcolor{yellow!55}0.007 & \cellcolor{orange!20}0.193 \\*
 &  & Logit & \cellcolor{orange!20}0.551 & \cellcolor{yellow!55}0.003 & \cellcolor{yellow!55}0.053 & \cellcolor{orange!20}0.533 & \cellcolor{yellow!55}0.004 & \cellcolor{yellow!55}0.064 & \cellcolor{orange!35}0.500 & \cellcolor{orange!65}0.950 & \cellcolor{orange!65}0.975 & \cellcolor{orange!20}0.576 & \cellcolor{orange!20}0.038 & \cellcolor{orange!20}0.199 & \cellcolor{orange!20}0.579 & \cellcolor{orange!20}0.045 & \cellcolor{orange!20}0.221 \\*
\cmidrule(lr){2-18}
 & \multirow{2}{*}{\texttt{munin1}} & Verb. & \cellcolor{yellow!35}0.649 & \cellcolor{orange!35}0.158 & \cellcolor{orange!20}0.201 & \cellcolor{orange!65}0.282 & \cellcolor{orange!35}0.070 & \cellcolor{yellow!35}0.127 & \cellcolor{orange!20}0.550 & \cellcolor{yellow!35}0.013 & \cellcolor{orange!20}0.212 & \cellcolor{orange!65}0.266 & \cellcolor{orange!20}0.068 & \cellcolor{orange!20}0.229 & \cellcolor{orange!65}0.197 & \cellcolor{orange!35}0.162 & \cellcolor{orange!35}0.367 \\*
 &  & Logit & \cellcolor{orange!35}0.500 & \cellcolor{yellow!55}0.000 & \cellcolor{yellow!55}0.008 & \cellcolor{yellow!35}0.617 & \cellcolor{yellow!55}0.001 & \cellcolor{yellow!55}0.034 & \cellcolor{orange!35}0.500 & \cellcolor{orange!65}0.984 & \cellcolor{orange!65}0.992 & \cellcolor{orange!20}0.559 & \cellcolor{orange!20}0.054 & \cellcolor{orange!20}0.233 & \cellcolor{orange!35}0.511 & \cellcolor{orange!50}0.215 & \cellcolor{orange!50}0.464 \\
\specialrule{0.10em}{0.10em}{0.10em}
\pagebreak[4]
\multirow{12}{*}{\rotatebox[origin=c]{90}{\parbox{1.8cm}{\centering Llama-3.3-70B-\\Instruct}}} & \multirow{2}{*}{\texttt{asiam}} & Verb. & \cellcolor{orange!50}0.331 & \cellcolor{orange!50}0.395 & \cellcolor{orange!50}0.507 & \cellcolor{orange!65}0.145 & \cellcolor{orange!65}0.543 & \cellcolor{orange!50}0.613 & \cellcolor{yellow!35}0.696 & \cellcolor{yellow!55}0.006 & \cellcolor{yellow!35}0.131 & \cellcolor{orange!35}0.497 & \cellcolor{yellow!55}0.009 & \cellcolor{orange!20}0.175 & \cellcolor{orange!50}0.367 & \cellcolor{yellow!35}0.029 & \cellcolor{orange!20}0.191 \\*
 &  & Logit & \cellcolor{orange!20}0.525 & \cellcolor{yellow!35}0.029 & \cellcolor{yellow!35}0.131 & \cellcolor{yellow!35}0.598 & \cellcolor{yellow!35}0.021 & \cellcolor{yellow!55}0.076 & \cellcolor{orange!35}0.500 & \cellcolor{yellow!35}0.028 & \cellcolor{orange!20}0.167 & \cellcolor{yellow!35}0.608 & \cellcolor{orange!20}0.046 & \cellcolor{orange!20}0.208 & \cellcolor{orange!20}0.556 & \cellcolor{orange!20}0.044 & \cellcolor{orange!20}0.198 \\*
\cmidrule(lr){2-18}
 & \multirow{2}{*}{\texttt{river}} & Verb. & \cellcolor{orange!65}0.311 & \cellcolor{orange!50}0.369 & \cellcolor{orange!50}0.476 & \cellcolor{orange!65}0.248 & \cellcolor{orange!50}0.467 & \cellcolor{orange!50}0.570 & \cellcolor{orange!20}0.553 & \cellcolor{orange!20}0.048 & \cellcolor{orange!35}0.281 & \cellcolor{orange!65}0.318 & \cellcolor{orange!20}0.048 & \cellcolor{orange!20}0.231 & \cellcolor{orange!50}0.347 & \cellcolor{orange!35}0.117 & \cellcolor{orange!35}0.339 \\*
 &  & Logit & \cellcolor{orange!20}0.562 & \cellcolor{yellow!35}0.016 & \cellcolor{yellow!35}0.126 & \cellcolor{yellow!35}0.616 & \cellcolor{yellow!35}0.024 & \cellcolor{yellow!35}0.124 & \cellcolor{orange!35}0.520 & \cellcolor{orange!35}0.159 & \cellcolor{orange!50}0.395 & \cellcolor{orange!20}0.561 & \cellcolor{orange!20}0.060 & \cellcolor{orange!35}0.259 & \cellcolor{orange!20}0.524 & \cellcolor{orange!35}0.169 & \cellcolor{orange!50}0.404 \\*
\cmidrule(lr){2-18}
 & \multirow{2}{*}{\texttt{covid}} & Verb. & \cellcolor{orange!65}0.304 & \cellcolor{orange!65}0.607 & \cellcolor{orange!65}0.668 & \cellcolor{orange!50}0.407 & \cellcolor{orange!65}0.753 & \cellcolor{orange!65}0.818 & \cellcolor{yellow!55}0.799 & \cellcolor{yellow!35}0.009 & \cellcolor{yellow!55}0.088 & \cellcolor{orange!35}0.509 & \cellcolor{yellow!35}0.019 & \cellcolor{yellow!35}0.116 & \cellcolor{orange!35}0.506 & \cellcolor{yellow!55}0.006 & \cellcolor{yellow!35}0.150 \\*
 &  & Logit & \cellcolor{yellow!35}0.613 & \cellcolor{yellow!55}0.008 & \cellcolor{yellow!55}0.051 & \cellcolor{orange!35}0.519 & \cellcolor{yellow!55}0.004 & \cellcolor{yellow!55}0.050 & \cellcolor{orange!20}0.540 & \cellcolor{yellow!35}0.012 & \cellcolor{yellow!35}0.102 & \cellcolor{yellow!35}0.674 & \cellcolor{yellow!35}0.012 & \cellcolor{yellow!55}0.095 & \cellcolor{orange!20}0.542 & \cellcolor{orange!20}0.031 & \cellcolor{orange!20}0.172 \\*
\cmidrule(lr){2-18}
 & \multirow{2}{*}{\texttt{coal}} & Verb. & \cellcolor{orange!50}0.470 & \cellcolor{orange!65}0.939 & \cellcolor{orange!65}0.967 & \cellcolor{orange!50}0.340 & \cellcolor{orange!50}0.487 & \cellcolor{orange!50}0.513 & \cellcolor{orange!20}0.580 & \cellcolor{orange!35}0.090 & \cellcolor{orange!35}0.322 & \cellcolor{orange!50}0.368 & \cellcolor{yellow!35}0.012 & \cellcolor{orange!20}0.173 & \cellcolor{orange!65}0.301 & \cellcolor{orange!50}0.226 & \cellcolor{orange!50}0.423 \\*
 &  & Logit & \cellcolor{orange!35}0.500 & \cellcolor{yellow!55}0.001 & \cellcolor{yellow!55}0.026 & \cellcolor{orange!20}0.525 & \cellcolor{yellow!55}0.001 & \cellcolor{yellow!55}0.026 & \cellcolor{orange!35}0.506 & \cellcolor{orange!50}0.207 & \cellcolor{orange!50}0.456 & \cellcolor{yellow!35}0.657 & \cellcolor{yellow!35}0.019 & \cellcolor{orange!20}0.163 & \cellcolor{orange!35}0.501 & \cellcolor{orange!50}0.266 & \cellcolor{orange!50}0.514 \\*
\cmidrule(lr){2-18}
 & \multirow{2}{*}{\texttt{hepar2}} & Verb. & \cellcolor{orange!65}0.164 & \cellcolor{orange!65}0.514 & \cellcolor{orange!50}0.562 & \cellcolor{orange!65}0.140 & \cellcolor{orange!65}0.607 & \cellcolor{orange!65}0.645 & \cellcolor{yellow!55}0.748 & \cellcolor{orange!20}0.047 & \cellcolor{orange!20}0.215 & \cellcolor{orange!50}0.433 & \cellcolor{yellow!35}0.012 & \cellcolor{yellow!35}0.157 & \cellcolor{orange!20}0.523 & \cellcolor{orange!20}0.065 & \cellcolor{orange!35}0.296 \\*
 &  & Logit & \cellcolor{yellow!35}0.629 & \cellcolor{yellow!55}0.003 & \cellcolor{yellow!55}0.051 & \cellcolor{yellow!35}0.657 & \cellcolor{yellow!55}0.002 & \cellcolor{yellow!55}0.039 & \cellcolor{orange!20}0.521 & \cellcolor{orange!35}0.087 & \cellcolor{orange!35}0.296 & \cellcolor{yellow!35}0.599 & \cellcolor{yellow!35}0.024 & \cellcolor{orange!20}0.163 & \cellcolor{orange!35}0.514 & \cellcolor{orange!35}0.135 & \cellcolor{orange!35}0.369 \\*
\cmidrule(lr){2-18}
 & \multirow{2}{*}{\texttt{munin1}} & Verb. & \cellcolor{orange!50}0.339 & \cellcolor{orange!65}0.688 & \cellcolor{orange!65}0.697 & \cellcolor{orange!50}0.417 & \cellcolor{orange!50}0.464 & \cellcolor{orange!50}0.475 & \cellcolor{orange!35}0.507 & \cellcolor{orange!35}0.088 & \cellcolor{orange!35}0.337 & \cellcolor{orange!65}0.076 & \cellcolor{orange!35}0.075 & \cellcolor{yellow!55}0.089 & \cellcolor{orange!65}0.150 & \cellcolor{orange!50}0.490 & \cellcolor{orange!50}0.594 \\*
 &  & Logit & \cellcolor{orange!35}0.502 & \cellcolor{yellow!55}0.000 & \cellcolor{yellow!55}0.008 & \cellcolor{orange!20}0.522 & \cellcolor{yellow!55}0.000 & \cellcolor{yellow!55}0.008 & \cellcolor{orange!35}0.508 & \cellcolor{orange!50}0.218 & \cellcolor{orange!50}0.467 & \cellcolor{yellow!55}0.795 & \cellcolor{yellow!55}0.001 & \cellcolor{yellow!55}0.042 & \cellcolor{orange!50}0.467 & \cellcolor{orange!50}0.502 & \cellcolor{orange!65}0.700 \\
\specialrule{0.10em}{0.10em}{0.10em}
\multirow{12}{*}{\rotatebox[origin=c]{90}{\parbox{1.8cm}{\centering Llama-3.1-70B-\\Instruct}}} & \multirow{2}{*}{\texttt{asiam}} & Verb. & \cellcolor{orange!65}0.317 & \cellcolor{orange!50}0.289 & \cellcolor{orange!50}0.448 & \cellcolor{orange!65}0.273 & \cellcolor{orange!50}0.441 & \cellcolor{orange!50}0.519 & \cellcolor{yellow!35}0.637 & \cellcolor{orange!20}0.040 & \cellcolor{orange!20}0.229 & \cellcolor{yellow!35}0.686 & \cellcolor{yellow!35}0.019 & \cellcolor{orange!20}0.190 & \cellcolor{orange!50}0.406 & \cellcolor{yellow!35}0.016 & \cellcolor{yellow!35}0.151 \\*
 &  & Logit & \cellcolor{yellow!55}0.776 & \cellcolor{yellow!35}0.026 & \cellcolor{orange!20}0.169 & \cellcolor{yellow!55}0.734 & \cellcolor{yellow!35}0.026 & \cellcolor{yellow!35}0.109 & \cellcolor{orange!20}0.587 & \cellcolor{orange!20}0.045 & \cellcolor{orange!20}0.223 & \cellcolor{yellow!35}0.702 & \cellcolor{orange!20}0.042 & \cellcolor{orange!20}0.201 & \cellcolor{yellow!35}0.634 & \cellcolor{yellow!35}0.021 & \cellcolor{yellow!35}0.130 \\*
\cmidrule(lr){2-18}
 & \multirow{2}{*}{\texttt{river}} & Verb. & \cellcolor{orange!65}0.278 & \cellcolor{orange!50}0.361 & \cellcolor{orange!50}0.502 & \cellcolor{orange!50}0.356 & \cellcolor{orange!50}0.322 & \cellcolor{orange!50}0.457 & \cellcolor{orange!50}0.464 & \cellcolor{orange!20}0.060 & \cellcolor{orange!35}0.285 & \cellcolor{orange!50}0.364 & \cellcolor{orange!20}0.057 & \cellcolor{orange!35}0.253 & \cellcolor{orange!50}0.452 & \cellcolor{orange!35}0.144 & \cellcolor{orange!50}0.385 \\*
 &  & Logit & \cellcolor{yellow!55}0.760 & \cellcolor{yellow!55}0.007 & \cellcolor{yellow!35}0.124 & \cellcolor{yellow!55}0.714 & \cellcolor{yellow!55}0.006 & \cellcolor{yellow!35}0.107 & \cellcolor{orange!50}0.477 & \cellcolor{orange!35}0.125 & \cellcolor{orange!35}0.357 & \cellcolor{yellow!55}0.705 & \cellcolor{orange!20}0.050 & \cellcolor{orange!20}0.231 & \cellcolor{orange!35}0.520 & \cellcolor{orange!35}0.142 & \cellcolor{orange!35}0.383 \\*
\cmidrule(lr){2-18}
 & \multirow{2}{*}{\texttt{covid}} & Verb. & \cellcolor{orange!65}0.191 & \cellcolor{orange!65}0.635 & \cellcolor{orange!65}0.698 & \cellcolor{orange!65}0.287 & \cellcolor{orange!65}0.602 & \cellcolor{orange!65}0.655 & \cellcolor{orange!20}0.576 & \cellcolor{yellow!55}0.002 & \cellcolor{yellow!55}0.092 & \cellcolor{orange!35}0.516 & \cellcolor{yellow!35}0.012 & \cellcolor{yellow!35}0.155 & \cellcolor{orange!20}0.524 & \cellcolor{yellow!55}0.006 & \cellcolor{yellow!35}0.150 \\*
 &  & Logit & \cellcolor{yellow!55}0.865 & \cellcolor{yellow!55}0.001 & \cellcolor{yellow!55}0.041 & \cellcolor{yellow!55}0.812 & \cellcolor{yellow!55}0.001 & \cellcolor{yellow!55}0.041 & \cellcolor{yellow!55}0.883 & \cellcolor{yellow!35}0.009 & \cellcolor{yellow!55}0.067 & \cellcolor{yellow!55}0.817 & \cellcolor{yellow!35}0.011 & \cellcolor{yellow!35}0.106 & \cellcolor{yellow!35}0.694 & \cellcolor{yellow!35}0.010 & \cellcolor{yellow!35}0.139 \\*
\cmidrule(lr){2-18}
 & \multirow{2}{*}{\texttt{coal}} & Verb. & \cellcolor{orange!50}0.446 & \cellcolor{orange!50}0.384 & \cellcolor{orange!50}0.418 & \cellcolor{orange!65}0.236 & \cellcolor{orange!50}0.396 & \cellcolor{orange!50}0.453 & \cellcolor{orange!35}0.486 & \cellcolor{orange!35}0.087 & \cellcolor{orange!35}0.325 & \cellcolor{orange!50}0.368 & \cellcolor{orange!20}0.033 & \cellcolor{orange!20}0.178 & \cellcolor{orange!50}0.325 & \cellcolor{orange!35}0.165 & \cellcolor{orange!35}0.378 \\*
 &  & Logit & \cellcolor{orange!35}0.501 & \cellcolor{yellow!55}0.000 & \cellcolor{yellow!55}0.027 & \cellcolor{yellow!55}0.713 & \cellcolor{yellow!55}0.004 & \cellcolor{yellow!55}0.028 & \cellcolor{orange!20}0.555 & \cellcolor{orange!35}0.128 & \cellcolor{orange!35}0.373 & \cellcolor{yellow!55}0.829 & \cellcolor{yellow!55}0.001 & \cellcolor{yellow!35}0.105 & \cellcolor{orange!20}0.562 & \cellcolor{orange!35}0.112 & \cellcolor{orange!35}0.357 \\*
\cmidrule(lr){2-18}
 & \multirow{2}{*}{\texttt{hepar2}} & Verb. & \cellcolor{orange!65}0.167 & \cellcolor{orange!50}0.452 & \cellcolor{orange!50}0.538 & \cellcolor{orange!65}0.235 & \cellcolor{orange!50}0.386 & \cellcolor{orange!50}0.461 & \cellcolor{yellow!35}0.610 & \cellcolor{orange!20}0.036 & \cellcolor{orange!35}0.256 & \cellcolor{orange!35}0.495 & \cellcolor{yellow!35}0.023 & \cellcolor{orange!20}0.219 & \cellcolor{orange!35}0.511 & \cellcolor{orange!35}0.078 & \cellcolor{orange!35}0.314 \\*
 &  & Logit & \cellcolor{yellow!55}0.803 & \cellcolor{yellow!55}0.001 & \cellcolor{yellow!55}0.064 & \cellcolor{yellow!55}0.868 & \cellcolor{yellow!55}0.001 & \cellcolor{yellow!55}0.045 & \cellcolor{yellow!35}0.604 & \cellcolor{orange!20}0.063 & \cellcolor{orange!35}0.283 & \cellcolor{yellow!35}0.700 & \cellcolor{yellow!35}0.024 & \cellcolor{orange!20}0.197 & \cellcolor{yellow!35}0.609 & \cellcolor{orange!35}0.089 & \cellcolor{orange!35}0.320 \\*
\cmidrule(lr){2-18}
 & \multirow{2}{*}{\texttt{munin1}} & Verb. & \cellcolor{orange!65}0.112 & \cellcolor{orange!50}0.428 & \cellcolor{orange!50}0.469 & \cellcolor{orange!65}0.201 & \cellcolor{orange!50}0.366 & \cellcolor{orange!50}0.403 & \cellcolor{orange!50}0.386 & \cellcolor{orange!50}0.225 & \cellcolor{orange!50}0.438 & \cellcolor{orange!65}0.201 & \cellcolor{orange!35}0.108 & \cellcolor{orange!35}0.260 & \cellcolor{orange!65}0.221 & \cellcolor{orange!50}0.467 & \cellcolor{orange!50}0.612 \\*
 &  & Logit & \cellcolor{yellow!55}0.869 & \cellcolor{yellow!55}0.002 & \cellcolor{yellow!55}0.011 & \cellcolor{yellow!55}0.786 & \cellcolor{yellow!55}0.003 & \cellcolor{yellow!55}0.013 & \cellcolor{orange!50}0.446 & \cellcolor{orange!50}0.316 & \cellcolor{orange!50}0.532 & \cellcolor{yellow!55}0.842 & \cellcolor{yellow!35}0.012 & \cellcolor{yellow!55}0.092 & \cellcolor{orange!50}0.356 & \cellcolor{orange!50}0.497 & \cellcolor{orange!65}0.644 \\
\end{xltabular}
\endgroup
\end{landscape}

\begin{table}[H]
\centering
\begin{tabular}{llcccccc}
\toprule
 &  & \multicolumn{3}{c}{\textbf{Small LLMs}} & \multicolumn{3}{c}{\textbf{Large LLMs}} \\
\cmidrule(lr){3-5} \cmidrule(lr){6-8}
\textbf{Dataset} & \textbf{Prompt} & \textbf{AUROC $\uparrow$} & \textbf{ECE $\downarrow$} & \textbf{Brier $\downarrow$} & \textbf{AUROC $\uparrow$} & \textbf{ECE $\downarrow$} & \textbf{Brier $\downarrow$} \\
\midrule
\multirow{5}{*}{\rotatebox[origin=c]{90}{\texttt{asiam}}} & Name-only & 0.772 & \textbf{0.008} & 0.135 & 0.632 & 0.022 & 0.114 \\
 & Metadata & 0.711 & 0.014 & \textbf{0.105} & 0.656 & 0.012 & \textbf{0.054} \\
 & CoT & 0.655 & 0.024 & 0.131 & 0.829 & \textbf{0.008} & 0.099 \\
 & Few-shot & 0.671 & 0.012 & 0.207 & \textbf{0.871} & 0.018 & 0.095 \\
 & Few-shot + CoT & \textbf{0.833} & 0.019 & 0.158 & 0.523 & 0.016 & 0.100 \\
\midrule
\multirow{5}{*}{\rotatebox[origin=c]{90}{\texttt{river}}} & Name-only & \textbf{0.722} & 0.003 & \textbf{0.147} & 0.665 & \textbf{0.009} & 0.136 \\
 & Metadata & 0.711 & \textbf{0.001} & 0.150 & \textbf{0.726} & 0.010 & \textbf{0.124} \\
 & CoT & 0.654 & 0.009 & 0.193 & 0.631 & 0.086 & 0.292 \\
 & Few-shot & 0.478 & 0.051 & 0.291 & 0.708 & 0.036 & 0.216 \\
 & Few-shot + CoT & 0.566 & 0.046 & 0.286 & 0.526 & 0.102 & 0.338 \\
\midrule
\multirow{5}{*}{\rotatebox[origin=c]{90}{\texttt{covid}}} & Name-only & 0.620 & 0.015 & 0.068 & 0.774 & 0.005 & 0.045 \\
 & Metadata & 0.729 & 0.010 & \textbf{0.056} & 0.794 & \textbf{0.001} & \textbf{0.043} \\
 & CoT & 0.748 & 0.011 & 0.057 & \textbf{0.903} & 0.005 & 0.052 \\
 & Few-shot & \textbf{0.759} & \textbf{0.005} & 0.111 & 0.828 & 0.005 & 0.080 \\
 & Few-shot + CoT & 0.733 & 0.005 & 0.123 & 0.691 & 0.007 & 0.113 \\
\midrule
\multirow{5}{*}{\rotatebox[origin=c]{90}{\texttt{coal}}} & Name-only & 0.532 & 0.032 & \textbf{0.058} & 0.513 & \textbf{0.001} & \textbf{0.027} \\
 & Metadata & \textbf{0.802} & 0.014 & 0.080 & \textbf{0.868} & 0.002 & 0.062 \\
 & CoT & 0.768 & \textbf{0.011} & 0.102 & 0.594 & 0.029 & 0.228 \\
 & Few-shot & 0.629 & 0.014 & 0.235 & 0.699 & 0.010 & 0.159 \\
 & Few-shot + CoT & 0.650 & 0.027 & 0.249 & 0.554 & 0.077 & 0.315 \\
\midrule
\multirow{5}{*}{\rotatebox[origin=c]{90}{\texttt{hepar2}}} & Name-only & 0.738 & 0.008 & \textbf{0.099} & 0.743 & 0.003 & 0.071 \\
 & Metadata & 0.757 & \textbf{0.005} & 0.102 & \textbf{0.790} & \textbf{0.002} & \textbf{0.064} \\
 & CoT & \textbf{0.800} & 0.006 & 0.099 & 0.689 & 0.009 & 0.160 \\
 & Few-shot & 0.687 & 0.006 & 0.189 & 0.695 & 0.007 & 0.137 \\
 & Few-shot + CoT & 0.701 & 0.006 & 0.186 & 0.691 & 0.019 & 0.193 \\
\midrule
\multirow{5}{*}{\rotatebox[origin=c]{90}{\texttt{munin1}}} & Name-only & 0.725 & 0.037 & \textbf{0.054} & 0.790 & 0.011 & \textbf{0.025} \\
 & Metadata & \textbf{0.808} & 0.020 & 0.069 & \textbf{0.878} & 0.009 & 0.056 \\
 & CoT & 0.798 & \textbf{0.011} & 0.093 & 0.627 & 0.030 & 0.248 \\
 & Few-shot & 0.523 & 0.044 & 0.293 & 0.760 & \textbf{0.001} & 0.122 \\
 & Few-shot + CoT & 0.492 & 0.108 & 0.349 & 0.467 & 0.151 & 0.396 \\
\bottomrule
\end{tabular}
\caption{Cross-model agreement calibration results for small (left) and large (right) LLMs across all datasets. This method combines predictions across all models in each group for each fixed prompt style and dataset. Bold values indicate the best prompt within each dataset for each metric (small and large LLMs evaluated independently).}
\label{tab:calib_cross_model_combined}
\end{table}

\clearpage
\subsection{Statistical Comparison of Confidence Methods}
\label{app:calibration_significance}

Table~\ref{tab:wilcoxon_calibration} reports pairwise comparisons among the
four confidence methods. Tests are performed separately within each model
group and metric using the six paired dataset-level scores.

\begin{table}[H]
\centering
\resizebox{\textwidth}{!}{%
\begin{tabular}{llrrrlrrr}
\toprule
 & & \multicolumn{3}{c}{\textbf{Small LLMs}} & & \multicolumn{3}{c}{\textbf{Large LLMs}} \\
\cmidrule(lr){3-5} \cmidrule(lr){7-9}
\textbf{Metric} & \textbf{Comparison} & \textbf{$\widetilde{\Delta}$} & \textbf{Raw $p$} & \textbf{Holm $p$} & \textbf{Comparison} & \textbf{$\widetilde{\Delta}$} & \textbf{Raw $p$} & \textbf{Holm $p$} \\
\midrule
\multirow{6}{*}{ECE}
 & CM vs. CP    & -0.022 & 0.0625 & 0.1875 & CM vs. CP    & 0.007  & 0.0312 & 0.1875 \\
 & CM vs. Logit & -0.196 & 0.0312 & 0.1875 & CM vs. Logit & -0.189 & 0.0312 & 0.1875 \\
 & CM vs. Verb. & -0.207 & 0.0312 & 0.1875 & CM vs. Verb. & -0.179 & 0.0312 & 0.1875 \\
 & CP vs. Logit & -0.173 & 0.0312 & 0.1875 & CP vs. Logit & -0.196 & 0.0312 & 0.1875 \\
 & CP vs. Verb. & -0.191 & 0.0312 & 0.1875 & CP vs. Verb. & -0.179 & 0.0312 & 0.1875 \\
 & Logit vs. Verb. & -0.001 & 0.6875 & 0.6875 & Logit vs. Verb. & 0.019 & 0.4375 & 0.4375 \\
\midrule
\multirow{6}{*}{Brier}
 & CM vs. CP    & -0.029 & 0.0312 & 0.1875 & CM vs. CP    & 0.008  & 0.0938 & 0.1875 \\
 & CM vs. Logit & -0.200 & 0.0312 & 0.1875 & CM vs. Logit & -0.216 & 0.0312 & 0.1875 \\
 & CM vs. Verb. & -0.229 & 0.0312 & 0.1875 & CM vs. Verb. & -0.189 & 0.0312 & 0.1875 \\
 & CP vs. Logit & -0.170 & 0.0312 & 0.1875 & CP vs. Logit & -0.219 & 0.0312 & 0.1875 \\
 & CP vs. Verb. & -0.199 & 0.0312 & 0.1875 & CP vs. Verb. & -0.188 & 0.0312 & 0.1875 \\
 & Logit vs. Verb. & -0.011 & 0.4375 & 0.4375 & Logit vs. Verb. & 0.032 & 0.1562 & 0.1875 \\
\midrule
\multirow{6}{*}{AUROC}
 & CM vs. CP    & 0.026 & 0.0312 & 0.1875 & CM vs. CP    & -0.054 & 0.0312 & 0.1875 \\
 & CM vs. Logit & 0.089 & 0.0312 & 0.1875 & CM vs. Logit & 0.118  & 0.0312 & 0.1875 \\
 & CM vs. Verb. & 0.321 & 0.0312 & 0.1875 & CM vs. Verb. & 0.226  & 0.0312 & 0.1875 \\
 & CP vs. Logit & 0.059 & 0.0625 & 0.1875 & CP vs. Logit & 0.181  & 0.0312 & 0.1875 \\
 & CP vs. Verb. & 0.309 & 0.0312 & 0.1875 & CP vs. Verb. & 0.286  & 0.0312 & 0.1875 \\
 & Logit vs. Verb. & 0.246 & 0.0312 & 0.1875 & Logit vs. Verb. & 0.131 & 0.0312 & 0.1875 \\
\bottomrule
\end{tabular}%
}
\caption{Pairwise two-sided Wilcoxon signed-rank tests over six paired dataset-level scores, small (left) and large (right) LLMs evaluated separately. $\widetilde{\Delta}$ is the median difference (first method minus second). Holm correction is applied across the six pairwise comparisons within each model group and metric. CM~=~cross-model agreement, CP~=~cross-prompt agreement, Verb.~=~verbalized confidence. With $N=6$ paired observations, the smallest attainable two-sided raw $p$-value is $2/2^6 \approx 0.031$; no adjusted (Holm) $p$-value falls below 0.05.}
\label{tab:wilcoxon_calibration}
\end{table}

\clearpage
\subsection{Overconfident False Positives on Non-Edges}
\label{app:overconfident_false_positives}

\begin{table}[H]
\centering
\footnotesize
\begin{minipage}[t]{0.48\textwidth}
\centering
\textbf{(a) Confidence buckets among false positives.}\\[0.35em]
\resizebox{\linewidth}{!}{%
\begin{tabular}{lrrrr}
\toprule
Non-edge type &
\makecell{False-positive\\predictions} &
\makecell{Conf.\\$<50$} &
\makecell{Conf.\\$50$--$79$} &
\makecell{Conf.\\$\geq 80$} \\
\midrule
Reversed direct & 10,635 & 0.8\% & 14.7\% & 84.6\% \\
Indirect        & 64,503 & 0.6\% & 18.6\% & 80.8\% \\
Other           & 636,333 & 0.9\% & 20.7\% & 78.4\% \\
\bottomrule
\end{tabular}%
}
\end{minipage}
\hfill
\begin{minipage}[t]{0.48\textwidth}
\centering
\textbf{(b) High-confidence false positives by prompt.}\\[0.35em]
\resizebox{\linewidth}{!}{%
\begin{tabular}{lrrr}
\toprule
Prompt &
\makecell{Reversed direct\\non-edge} &
\makecell{Indirect\\non-edge} &
\makecell{Other\\non-edge} \\
\midrule
Few-shot + CoT & 46.4\% & 55.2\% & 42.5\% \\
Few-shot       & 36.6\% & 35.8\% & 25.0\% \\
CoT            & 30.3\% & 35.2\% & 21.5\% \\
Metadata       & 21.8\% & 21.7\% & 12.1\% \\
Name-only      & 16.9\% & 13.6\% & 9.5\% \\
\bottomrule
\end{tabular}%
}
\end{minipage}
\caption{Overconfident false-positive summaries on reference-graph non-edges. Panel (a) reports confidence distributions using false positives with a parseable verbal confidence score (99.9\% of all false positives); its counts therefore differ slightly from the all-inclusive counts in Table~\ref{tab:error_behavior_nonedge_type}. Panel (b) reports high-confidence false-positive rates over valid non-edge queries by prompt and non-edge type.}
\label{tab:overconfident_fp_side_by_side}
\label{tab:false_positive_confidence_buckets}
\label{tab:prompt_high_conf_fp}
\end{table}

\begin{table}[H]
\centering
\resizebox{\textwidth}{!}{%
\begin{tabular}{llrrr}
\toprule
Non-edge type &
Model &
\makecell{False-positive\\rate} &
\makecell{High-conf. false-\\positive rate} &
\makecell{False positives\\with conf. $\geq 80$} \\
\midrule
Reversed direct & Ministral-8B-Instruct-2410 & 58.6\% & 45.0\% & 76.8\% \\
Indirect & Gemma-4-31B-IT & 56.3\% & 52.2\% & 92.7\% \\
Other & Ministral-8B-Instruct-2410 & 56.4\% & 41.0\% & 72.7\% \\
\bottomrule
\end{tabular}%
}
\caption{
Worst model-level high-confidence false-positive behavior by non-edge type. For
each non-edge category, we report the model with the largest high-confidence
false-positive rate, along with its overall false-positive rate and the share
of its false positives that have verbal confidence of at least 80.
}
\label{tab:model_worst_case_high_conf_fp}
\end{table}

\clearpage
\onecolumn
\subsection{Additional Classification Figures}
\label{app:additional_classification_figures}


\begin{figure}[H]
    \centering
    \includegraphics[width=0.92\textwidth,keepaspectratio]{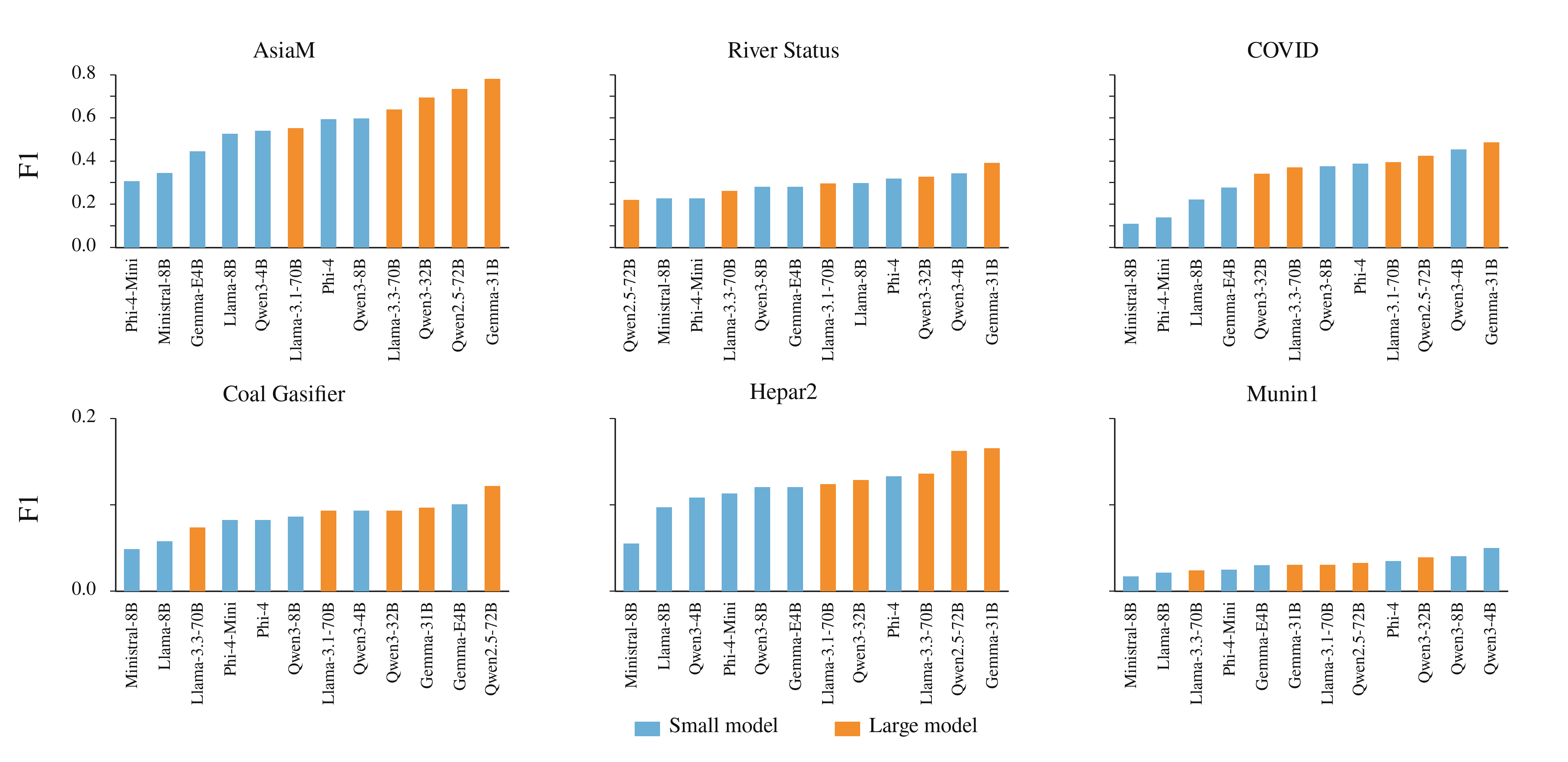}
    \caption{Model classification performance ranking across datasets. Performance is averaged over five prompt styles.}
    \label{fig:avg_f1_six_datasets}
\end{figure}

\begin{figure}[H]
    \centering
    \includegraphics[width=0.82\textwidth,keepaspectratio]{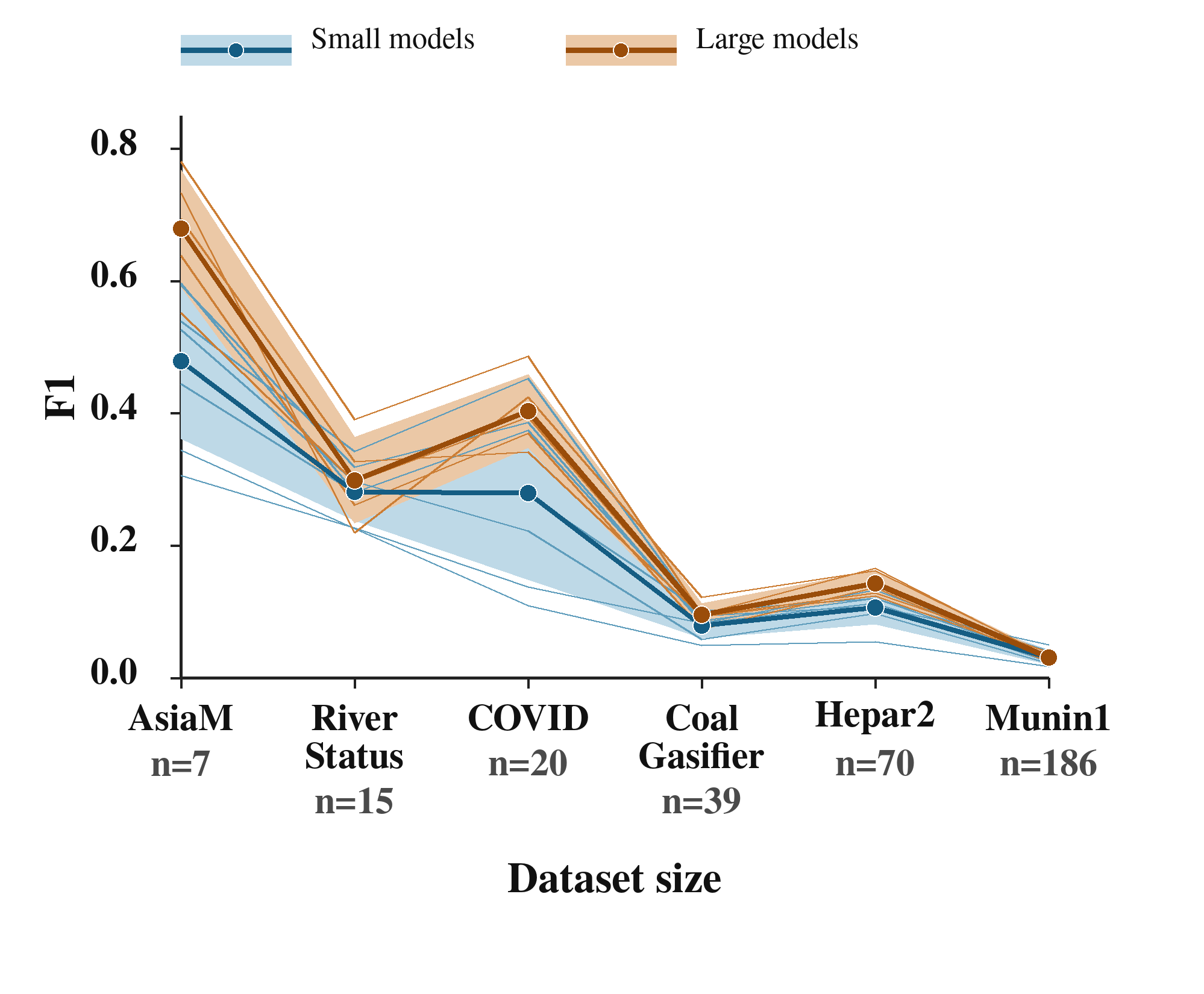}
    \caption{Model classification performance as a function of dataset size. Datasets are ordered by increasing number of variables. Points show mean F1 averaged over prompt styles, and shaded bands show variation across models within each group.}
    \label{fig:dataset_size_avg_f1_by_model}
\end{figure}

\clearpage
\subsection{Reliability Diagram Examples}
\label{app:reliability_diagram_examples}

\begin{figure}[H]
    \centering
    \captionsetup{font=scriptsize, skip=2pt}

    \begin{minipage}[t]{\textwidth}
        \centering
        \includegraphics[width=0.82\textwidth,keepaspectratio]{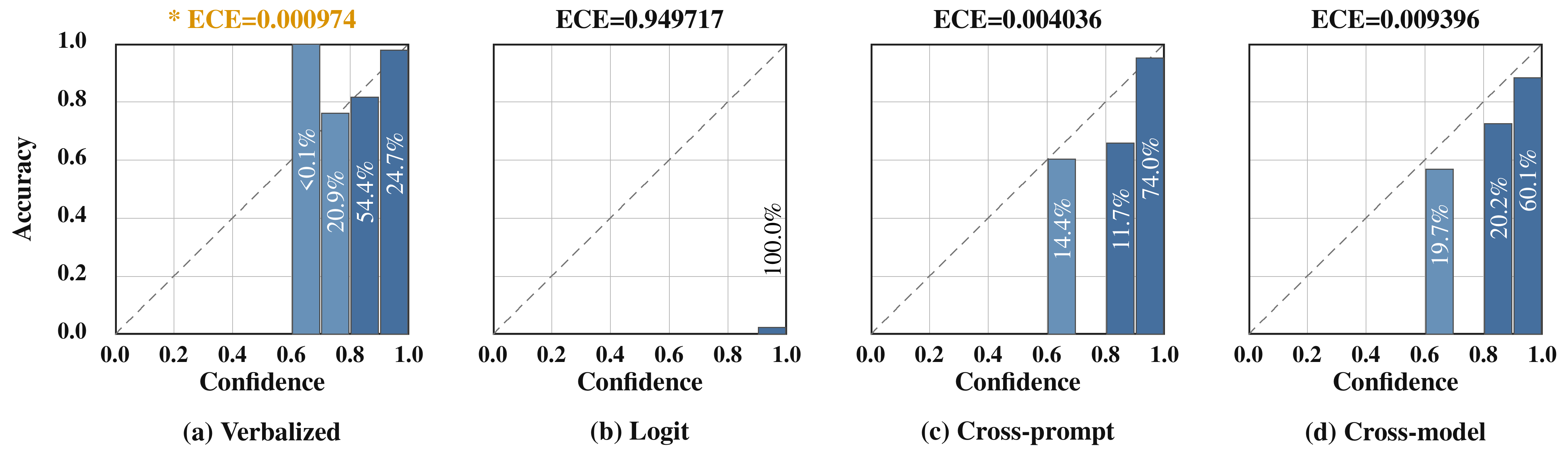}
        \vspace{0.1em}
        \caption{Reliability diagrams for Qwen2.5-72B-Instruct on Hepar2 with CoT prompting across four confidence sources: (a) verbalized, (b) logit-based, (c) cross-prompt, and (d) cross-model. Each panel reports ECE; $\star$ marks the lowest-ECE source, here verbalized confidence.}
        \label{fig:verbalized_best_reliability}
    \end{minipage}

    \vspace{0.35em}

    \begin{minipage}[t]{\textwidth}
        \centering
        \includegraphics[width=0.82\textwidth,keepaspectratio]{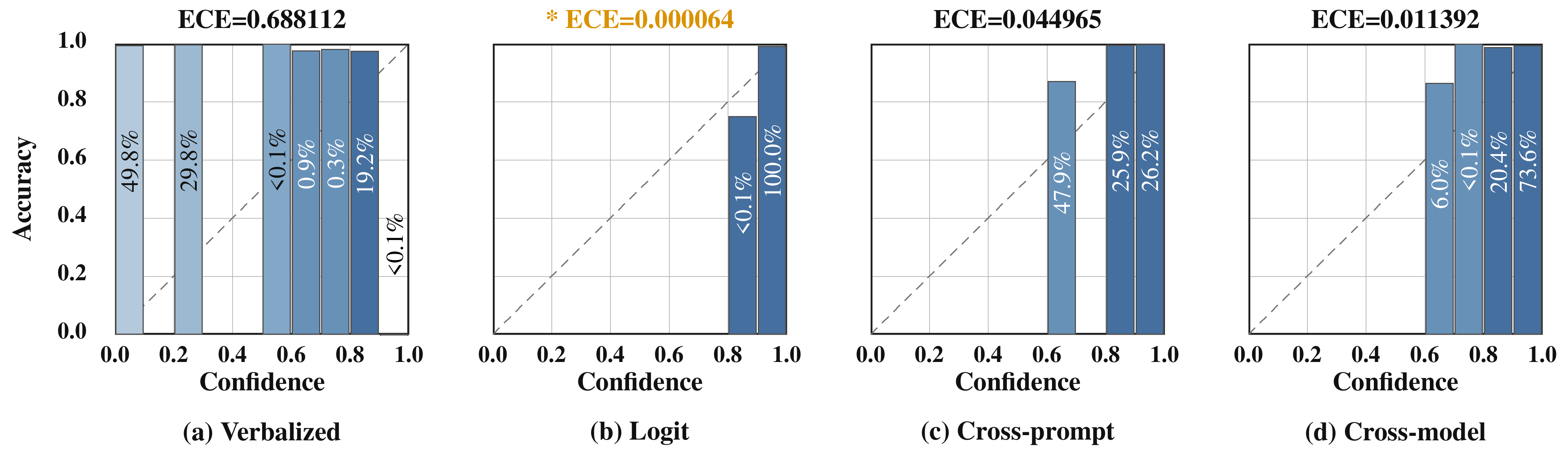}
        \vspace{0.1em}
        \caption{Reliability diagrams for Llama-3.3-70B-Instruct on Munin1 with name-only prompting across four confidence sources: (a) verbalized, (b) logit-based, (c) cross-prompt, and (d) cross-model. Each panel reports ECE; $\star$ marks the lowest-ECE source, here logit-based confidence.}
        \label{fig:logit_best_reliability}
    \end{minipage}

    \vspace{0.35em}

    \begin{minipage}[t]{0.48\textwidth}
        \centering
        \includegraphics[width=\linewidth,keepaspectratio]{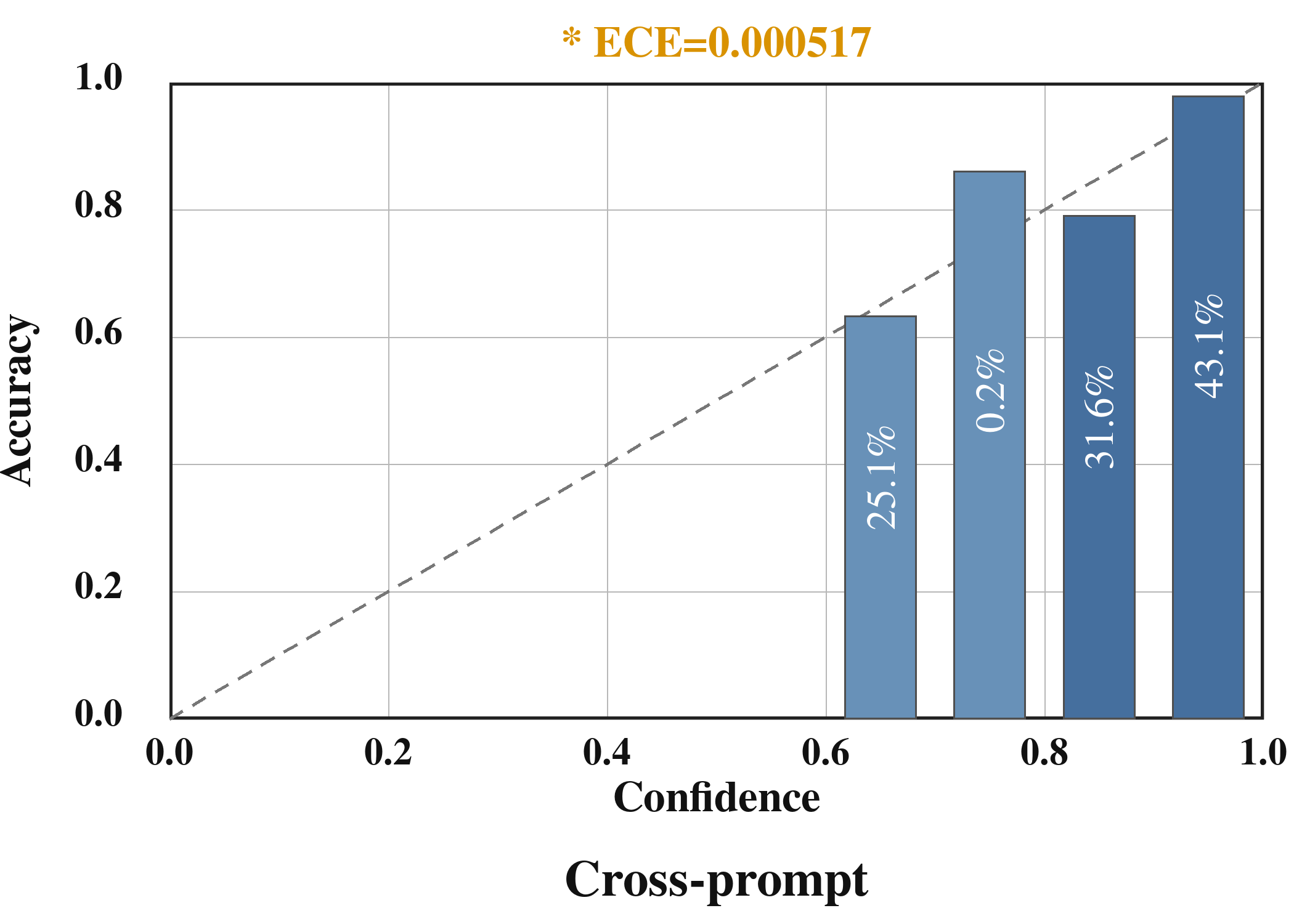}
        \vspace{-0.7em}
        \caption{Reliability diagram for cross-prompt agreement on Munin1 with Phi-4.}
        \label{fig:cross_prompt_best_reliability}
    \end{minipage}
    \hfill
    \begin{minipage}[t]{0.48\textwidth}
        \centering
        \includegraphics[width=\linewidth,keepaspectratio]{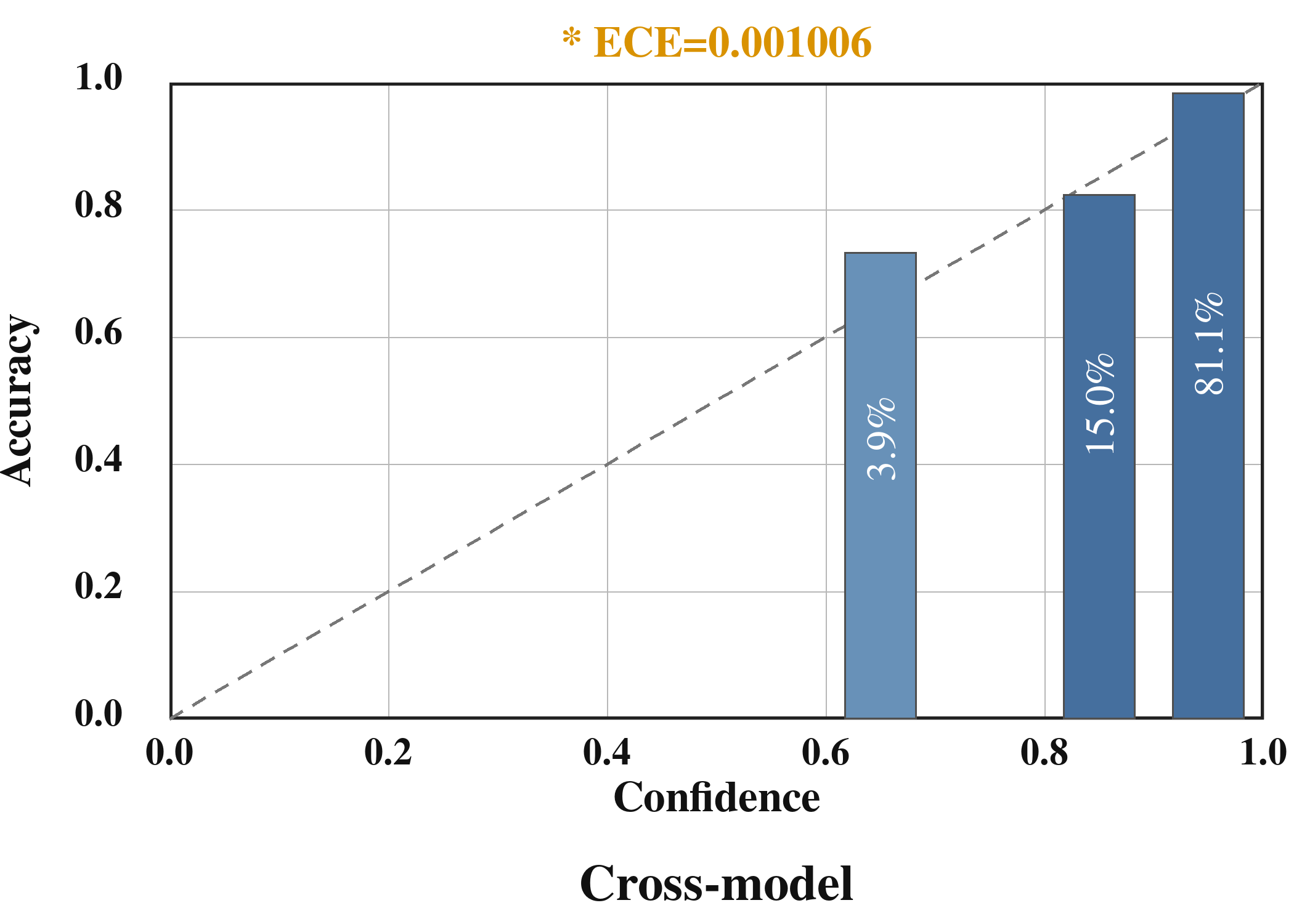}
        \vspace{-0.7em}
        \caption{Reliability diagram for cross-model agreement on COVID under metadata prompting.}
        \label{fig:cross_model_best_reliability}
    \end{minipage}
\end{figure}

\clearpage
\twocolumn

\subsection{Additional Confidence Distributions}
\label{app:additional_confidence_distributions}

\begingroup
\captionsetup{font=scriptsize,skip=2pt}

\begin{figure}[H]
    \centering
    \includegraphics[width=0.82\columnwidth,keepaspectratio]{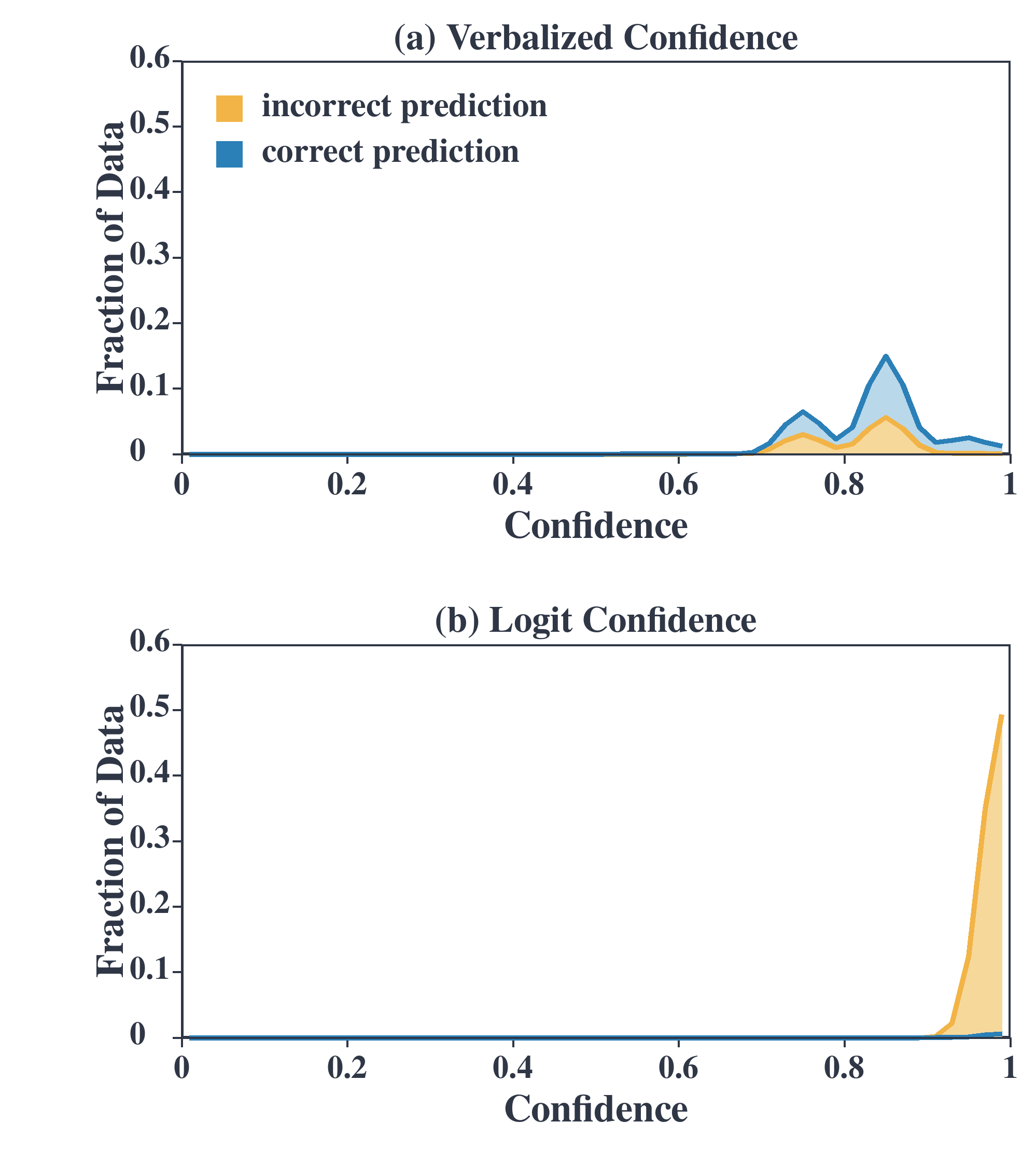}
    \vspace{-0.5em}
    \caption{Confidence distributions for Qwen2.5-72B-Instruct under the Chain-of-Thought prompt. Predictions are pooled across all six benchmark datasets. Blue curves denote correctly classified variable pairs, while orange curves denote misclassified pairs. Panel (a) shows verbalized confidence, and panel (b) shows logit-based confidence. Both confidence sources concentrate in the high-confidence region, with logit-based confidence collapsing almost entirely near 1.0 for both correct and incorrect predictions, indicating strong overconfidence and limited separation between reliable and unreliable predictions.}
    \label{fig:qwen25_72b_cot_confidence_distribution}
\end{figure}

\begin{figure}[H]
    \centering
    \includegraphics[width=0.82\columnwidth,keepaspectratio]{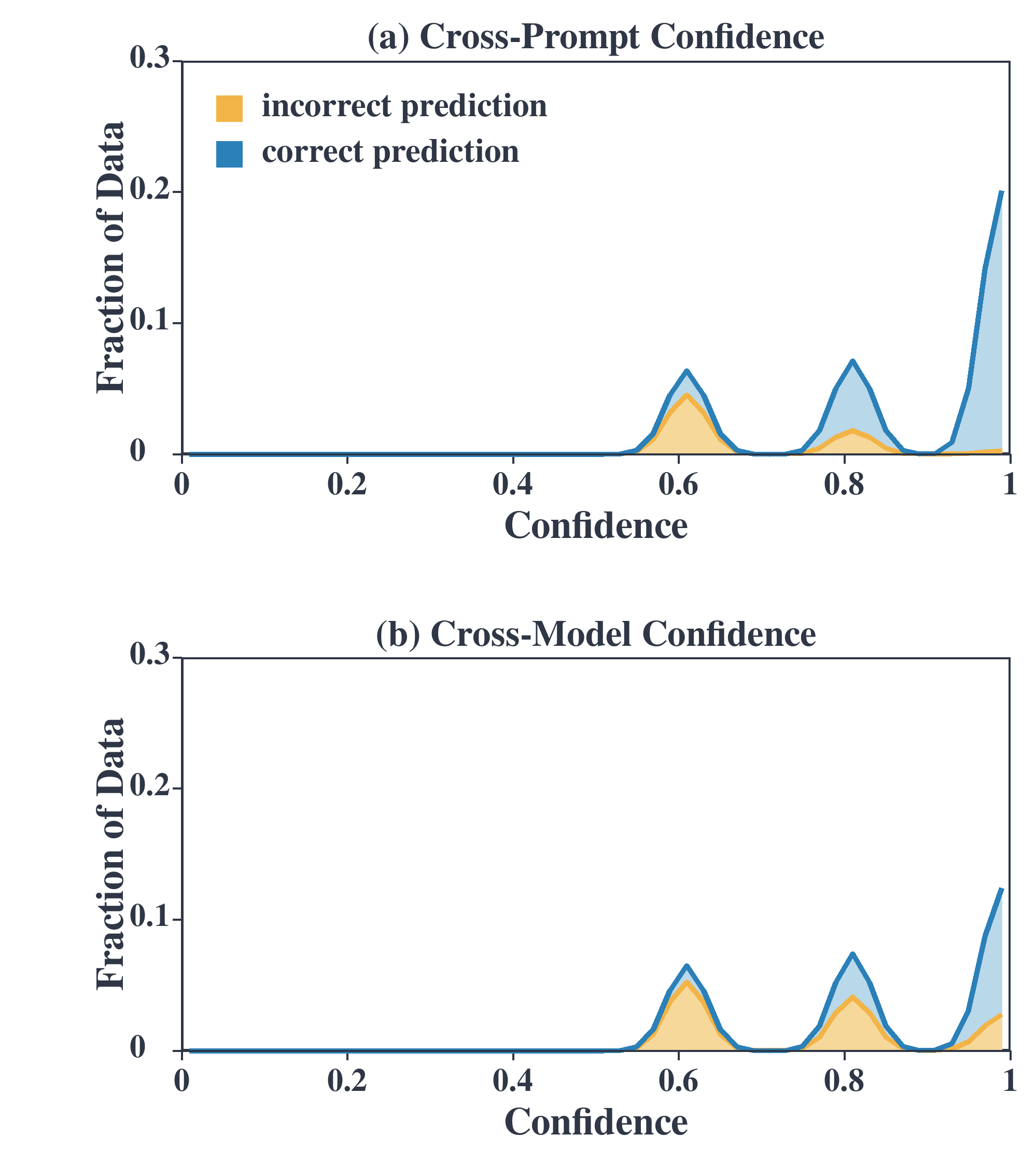}
    \vspace{-0.5em}
    \caption{Agreement-based confidence distributions for Qwen2.5-72B-Instruct and the large-model ensemble. Panel (a) shows cross-prompt agreement for Qwen2.5-72B-Instruct, while panel (b) shows cross-model agreement under the Chain-of-Thought prompt. Correct predictions concentrate more strongly near full agreement, indicating that agreement-based confidence is more informative than raw confidence for identifying reliable predictions.}
    \label{fig:qwen_cot_agreement_confidence}
\end{figure}

\begin{figure}[H]
    \centering
    \includegraphics[width=0.82\columnwidth,keepaspectratio]{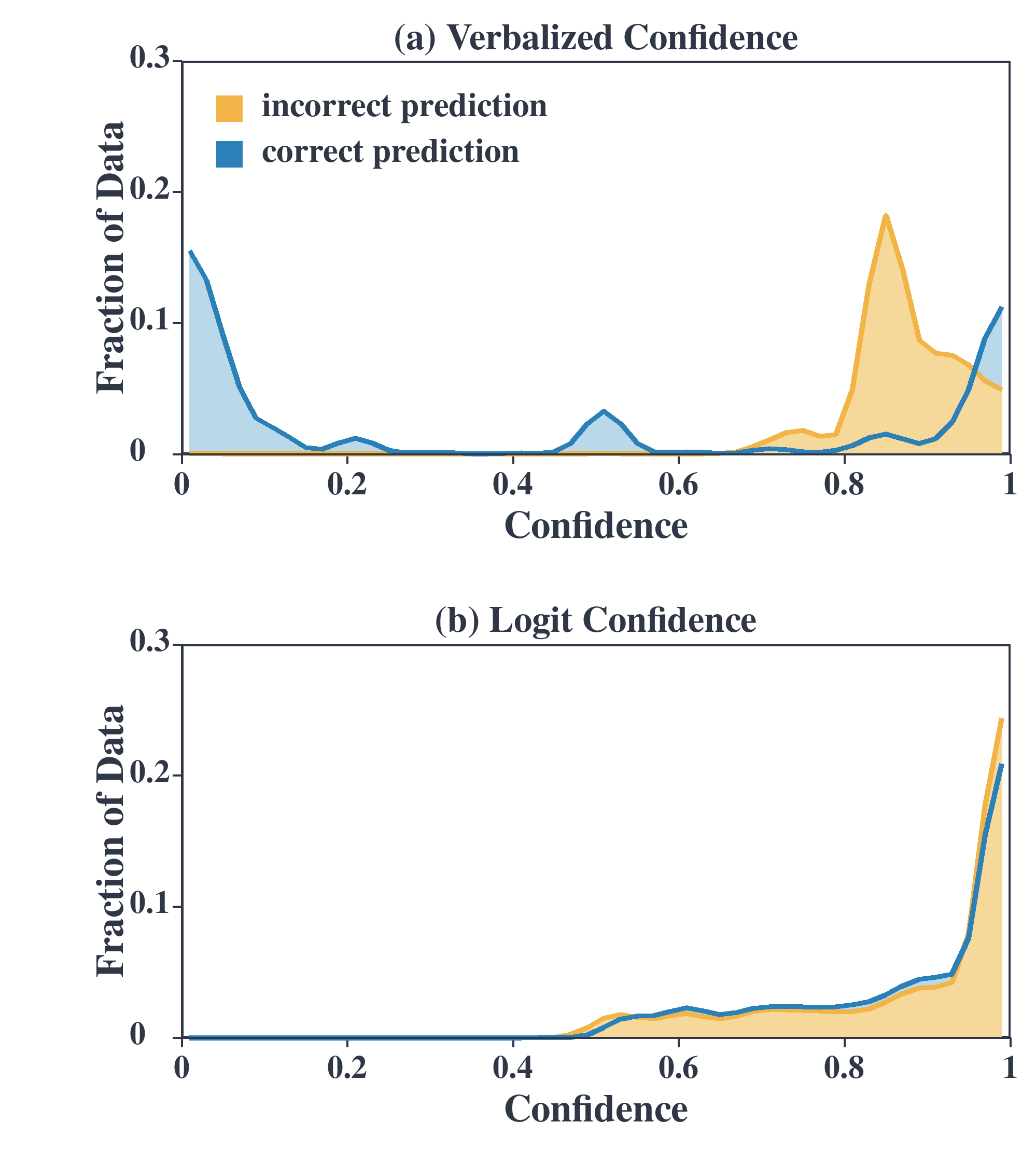}
    \vspace{-0.5em}
    \caption{Confidence distributions for Gemma-4-31B-IT using the Metadata prompt.}
    \label{fig:gemma_31b_metadata_confidence_distribution}
\end{figure}

\endgroup

\clearpage
\onecolumn
\subsection{Three-Way Label-Space Ablation Results}
\label{app:three_way_ablation}

\begin{table}[H]
\centering
\scriptsize
\setlength{\tabcolsep}{3.2pt}
\renewcommand{\arraystretch}{1.08}
\resizebox{\textwidth}{!}{%
\begin{tabular}{llcccccccc}
\toprule
\textbf{Model} & \textbf{Dataset}
& \textbf{P$\uparrow$}
& \textbf{R$\uparrow$}
& \textbf{F1$\uparrow$}
& \textbf{nSHD$\downarrow$}
& \makecell{\textbf{Non-edge}\\\textbf{FP Rate$\downarrow$}}
& \makecell{\textbf{Orientation}\\\textbf{Error$\downarrow$}}
& \makecell{\textbf{$A\!\rightarrow\!B$}\\\textbf{Recall$\uparrow$}}
& \makecell{\textbf{$B\!\rightarrow\!A$}\\\textbf{Recall$\uparrow$}} \\
\midrule
\multirow{4}{*}{Qwen3-4B}
& River Status & 0.247 & 0.760 & 0.373 & 0.305 & 72.5\% & 0.0\%  & 0.760 & -- \\
& COVID        & 0.167 & 0.500 & 0.250 & 0.197 & 39.6\% & 18.8\% & 0.812 & 0.000 \\
& Hepar2       & 0.067 & 0.813 & 0.124 & 0.291 & 60.6\% & 5.7\%  & 0.909 & 0.000 \\
& Munin1       & 0.016 & 0.637 & 0.031 & 0.311 & 62.9\% & 22.7\% & 0.879 & 0.000 \\
\midrule
\multirow{4}{*}{Phi-4}
& River Status & 0.287 & 0.920 & 0.438 & 0.281 & 71.3\% & 0.0\%  & 0.920 & -- \\
& COVID        & 0.471 & 0.615 & 0.533 & 0.071 & 11.0\% & 5.9\%  & 0.812 & 0.300 \\
& Hepar2       & 0.102 & 0.675 & 0.177 & 0.159 & 32.0\% & 8.8\%  & 0.718 & 0.308 \\
& Munin1       & 0.035 & 0.297 & 0.062 & 0.070 & 13.2\% & 17.3\% & 0.404 & 0.013 \\
\midrule
\multirow{4}{*}{Gemma-4-31B}
& River Status & 0.293 & 0.880 & 0.440 & 0.267 & 66.2\% & 0.0\%  & 0.880 & -- \\
& COVID        & 0.358 & 0.731 & 0.481 & 0.108 & 20.7\% & 0.0\%  & 0.812 & 0.600 \\
& Hepar2       & 0.079 & 0.724 & 0.142 & 0.220 & 45.5\% & 15.2\% & 0.764 & 0.385 \\
& Munin1       & 0.018 & 0.520 & 0.035 & 0.224 & 45.2\% & 30.4\% & 0.556 & 0.427 \\
\midrule
\multirow{4}{*}{Llama-3.3-70B}
& River Status & 0.278 & 1.000 & 0.435 & 0.310 & 81.2\% & 0.0\%  & 1.000 & -- \\
& COVID        & 0.242 & 0.615 & 0.348 & 0.153 & 30.5\% & 11.1\% & 0.875 & 0.200 \\
& Hepar2       & 0.075 & 0.846 & 0.138 & 0.267 & 55.6\% & 5.5\%  & 0.900 & 0.385 \\
& Munin1       & 0.013 & 0.718 & 0.025 & 0.448 & 91.0\% & 23.1\% & 0.990 & 0.000 \\
\bottomrule
\end{tabular}%
}
\caption{Dataset-level three-way classification results under Metadata prompting. Non-edge FP rate is the fraction of true no-edge pairs assigned either directed label. Orientation error is the fraction of detected true-edge pairs assigned the reversed direction. A dash indicates that the dataset contains no $B\!\rightarrow\!A$ instances under the fixed unordered-pair ordering.}
\label{tab:three_way_ablation_detailed}
\end{table}

\clearpage
\subsection{Benchmark Familiarity Audit}
\label{app:data_contamination}

\begin{table}[H]
\centering
\resizebox{1.1\textwidth}{!}{%
\begin{tabular}{llccccclcccccc}
\toprule
 & \multicolumn{6}{c}{\textbf{Small LLMs}} & \multicolumn{6}{c}{\textbf{Large LLMs}} \\
\cmidrule(lr){2-7} \cmidrule(lr){8-13}
\textbf{Dataset} & \textbf{Model} & \textbf{Nodes} & \textbf{Matches} & \textbf{Recall} & \textbf{Dev.} & \textbf{Risk} & \textbf{Model} & \textbf{Nodes} & \textbf{Matches} & \textbf{Recall} & \textbf{Dev.} & \textbf{Risk} \\
\midrule
\multirow{7}{*}{\rotatebox[origin=c]{90}{\texttt{asiam}}} & Gemma-4-E4B-IT & 10/7 & 2 & 0.286 & 0.429 & No & Qwen3-32B-Instruct & 7/7 & 6 & 0.857 & 0.000 & \cellcolor{bestshade}\textbf{Yes} \\
 & Llama-3.1-8B-Instruct & 8/7 & 1 & 0.143 & 0.143 & No & Gemma-4-31B-IT & 7/7 & 7 & 1.000 & 0.000 & \cellcolor{bestshade}\textbf{Yes} \\
 & Ministral-8B-Instruct-2410 & 1/7 & 1 & 0.143 & 0.857 & No & Qwen2.5-72B-Instruct & 8/7 & 7 & 1.000 & 0.143 & \cellcolor{bestshade}\textbf{Yes} \\
 & Phi-4 & 8/7 & 6 & 0.857 & 0.143 & \cellcolor{bestshade}\textbf{Yes} & Llama-3.3-70B-Instruct & 8/7 & 7 & 1.000 & 0.143 & \cellcolor{bestshade}\textbf{Yes} \\
 & Phi-4-Mini-Instruct & 5/7 & 3 & 0.429 & 0.286 & No & Llama-3.1-70B-Instruct & 10/7 & 7 & 1.000 & 0.429 & No \\
 & Qwen3-4B-Instruct & 2/7 & 1 & 0.143 & 0.714 & No &  &  &  &  &  &  \\
 & Qwen3-8B-Instruct & 9/7 & 6 & 0.857 & 0.286 & No &  &  &  &  &  &  \\
\midrule
\multirow{7}{*}{\rotatebox[origin=c]{90}{\texttt{river}}} & Gemma-4-E4B-IT & 16/15 & 6 & 0.400 & 0.067 & No & Qwen3-32B-Instruct & 9/15 & 8 & 0.533 & 0.400 & No \\
 & Llama-3.1-8B-Instruct & 11/15 & 3 & 0.200 & 0.267 & No & Gemma-4-31B-IT & 10/15 & 9 & 0.600 & 0.333 & No \\
 & Ministral-8B-Instruct-2410 & 9/15 & 7 & 0.467 & 0.400 & No & Qwen2.5-72B-Instruct & 8/15 & 6 & 0.400 & 0.467 & No \\
 & Phi-4 & 8/15 & 2 & 0.133 & 0.467 & No & Llama-3.3-70B-Instruct & 9/15 & 7 & 0.467 & 0.400 & No \\
 & Phi-4-Mini-Instruct & 9/15 & 7 & 0.467 & 0.400 & No & Llama-3.1-70B-Instruct & 13/15 & 3 & 0.200 & 0.133 & No \\
 & Qwen3-4B-Instruct & 10/15 & 1 & 0.067 & 0.333 & No &  &  &  &  &  &  \\
 & Qwen3-8B-Instruct & 3/15 & 1 & 0.067 & 0.800 & No &  &  &  &  &  &  \\
\midrule
\multirow{7}{*}{\rotatebox[origin=c]{90}{\texttt{covid}}} & Gemma-4-E4B-IT & 12/20 & 6 & 0.300 & 0.400 & No & Qwen3-32B-Instruct & 15/20 & 8 & 0.400 & 0.250 & No \\
 & Llama-3.1-8B-Instruct & 14/20 & 1 & 0.050 & 0.300 & No & Gemma-4-31B-IT & 6/20 & 3 & 0.150 & 0.700 & No \\
 & Ministral-8B-Instruct-2410 & 13/20 & 7 & 0.350 & 0.350 & No & Qwen2.5-72B-Instruct & 12/20 & 2 & 0.100 & 0.400 & No \\
 & Phi-4 & 13/20 & 2 & 0.100 & 0.350 & No & Llama-3.3-70B-Instruct & 15/20 & 2 & 0.100 & 0.250 & No \\
 & Phi-4-Mini-Instruct & 6/20 & 5 & 0.250 & 0.700 & No & Llama-3.1-70B-Instruct & 13/20 & 11 & 0.550 & 0.350 & No \\
 & Qwen3-4B-Instruct & 16/20 & 1 & 0.050 & 0.200 & No &  &  &  &  &  &  \\
 & Qwen3-8B-Instruct & 3/20 & 0 & 0.000 & 0.850 & No &  &  &  &  &  &  \\
\midrule
\multirow{7}{*}{\rotatebox[origin=c]{90}{\texttt{coal}}} & Gemma-4-E4B-IT & 8/39 & 7 & 0.179 & 0.795 & No & Qwen3-32B-Instruct & 1/39 & 0 & 0.000 & 0.974 & No \\
 & Llama-3.1-8B-Instruct & 10/39 & 7 & 0.179 & 0.744 & No & Gemma-4-31B-IT & 9/39 & 0 & 0.000 & 0.769 & No \\
 & Ministral-8B-Instruct-2410 & 11/39 & 10 & 0.256 & 0.718 & No & Qwen2.5-72B-Instruct & 10/39 & 7 & 0.179 & 0.744 & No \\
 & Phi-4 & 7/39 & 6 & 0.154 & 0.821 & No & Llama-3.3-70B-Instruct & 19/39 & 10 & 0.256 & 0.513 & No \\
 & Phi-4-Mini-Instruct & 6/39 & 6 & 0.154 & 0.846 & No & Llama-3.1-70B-Instruct & 8/39 & 7 & 0.179 & 0.795 & No \\
 & Qwen3-4B-Instruct & 15/39 & 8 & 0.205 & 0.615 & No &  &  &  &  &  &  \\
 & Qwen3-8B-Instruct & 3/39 & 0 & 0.000 & 0.923 & No &  &  &  &  &  &  \\
\midrule
\multirow{7}{*}{\rotatebox[origin=c]{90}{\texttt{hepar2}}} & Gemma-4-E4B-IT & 8/70 & 8 & 0.114 & 0.886 & No & Qwen3-32B-Instruct & 14/70 & 11 & 0.157 & 0.800 & No \\
 & Llama-3.1-8B-Instruct & 15/70 & 5 & 0.071 & 0.786 & No & Gemma-4-31B-IT & 13/70 & 12 & 0.171 & 0.814 & No \\
 & Ministral-8B-Instruct-2410 & 1/70 & 0 & 0.000 & 0.986 & No & Qwen2.5-72B-Instruct & 13/70 & 12 & 0.171 & 0.814 & No \\
 & Phi-4 & 13/70 & 13 & 0.186 & 0.814 & No & Llama-3.3-70B-Instruct & 14/70 & 11 & 0.157 & 0.800 & No \\
 & Phi-4-Mini-Instruct & 12/70 & 9 & 0.129 & 0.829 & No & Llama-3.1-70B-Instruct & 12/70 & 11 & 0.157 & 0.829 & No \\
 & Qwen3-4B-Instruct & 13/70 & 12 & 0.171 & 0.814 & No &  &  &  &  &  &  \\
 & Qwen3-8B-Instruct & 3/70 & 2 & 0.029 & 0.957 & No &  &  &  &  &  &  \\
\midrule
\multirow{7}{*}{\rotatebox[origin=c]{90}{\texttt{munin1}}} & Gemma-4-E4B-IT & 13/186 & 8 & 0.043 & 0.930 & No & Qwen3-32B-Instruct & 7/186 & 7 & 0.038 & 0.962 & No \\
 & Llama-3.1-8B-Instruct & 14/186 & 8 & 0.043 & 0.925 & No & Gemma-4-31B-IT & 12/186 & 12 & 0.065 & 0.935 & No \\
 & Ministral-8B-Instruct-2410 & 9/186 & 9 & 0.048 & 0.952 & No & Qwen2.5-72B-Instruct & 15/186 & 14 & 0.075 & 0.919 & No \\
 & Phi-4 & 7/186 & 7 & 0.038 & 0.962 & No & Llama-3.3-70B-Instruct & 14/186 & 13 & 0.070 & 0.925 & No \\
 & Phi-4-Mini-Instruct & 3/186 & 3 & 0.016 & 0.984 & No & Llama-3.1-70B-Instruct & 21/186 & 17 & 0.091 & 0.887 & No \\
 & Qwen3-4B-Instruct & 15/186 & 12 & 0.065 & 0.919 & No &  &  &  &  &  &  \\
 & Qwen3-8B-Instruct & 2/186 & 1 & 0.005 & 0.989 & No &  &  &  &  &  &  \\
\bottomrule
\end{tabular}%
}
\caption{Full node-level contamination results for all model--dataset pairs, small (left) and large (right) LLMs evaluated separately. Nodes reports generated/true counts, and Matches reports semantically aligned nodes. High-risk pairs (node-count deviation below 15\% and node recall above 0.85) are highlighted.}
\label{tab:full_node_contamination}
\end{table}

\clearpage
\subsection{Post-hoc Temperature Calibration}
\label{app:posthoc_temperature_calibration}

\begin{figure}[H]
    \centering
    \includegraphics[width=\columnwidth,keepaspectratio]{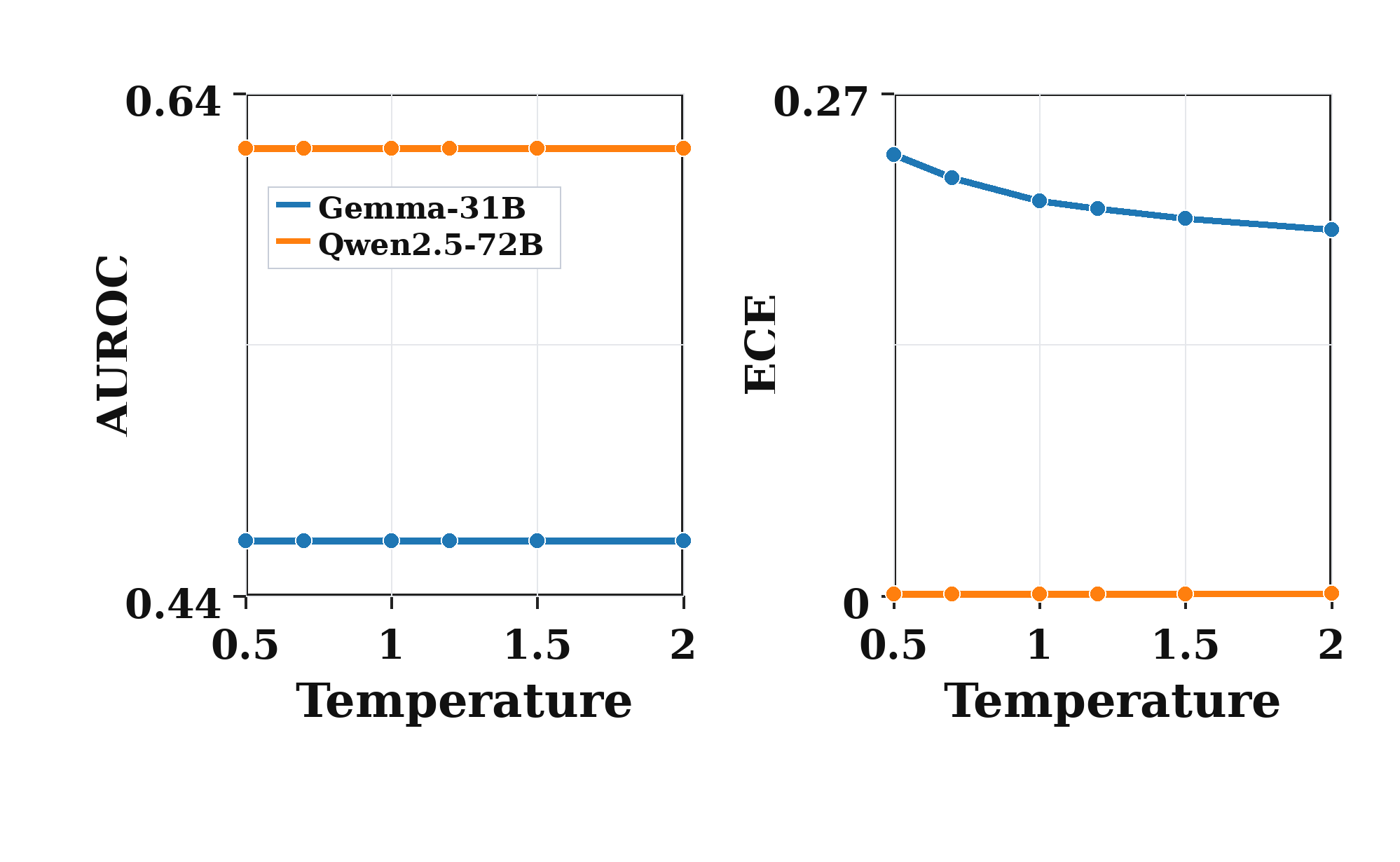}
    \caption{Post-hoc temperature sensitivity of logit-based calibration on Munin1.}
    \label{fig:temperature_sensitivity_munin1}
\end{figure}

\begin{figure}[H]
    \centering
    \includegraphics[width=\columnwidth,keepaspectratio]{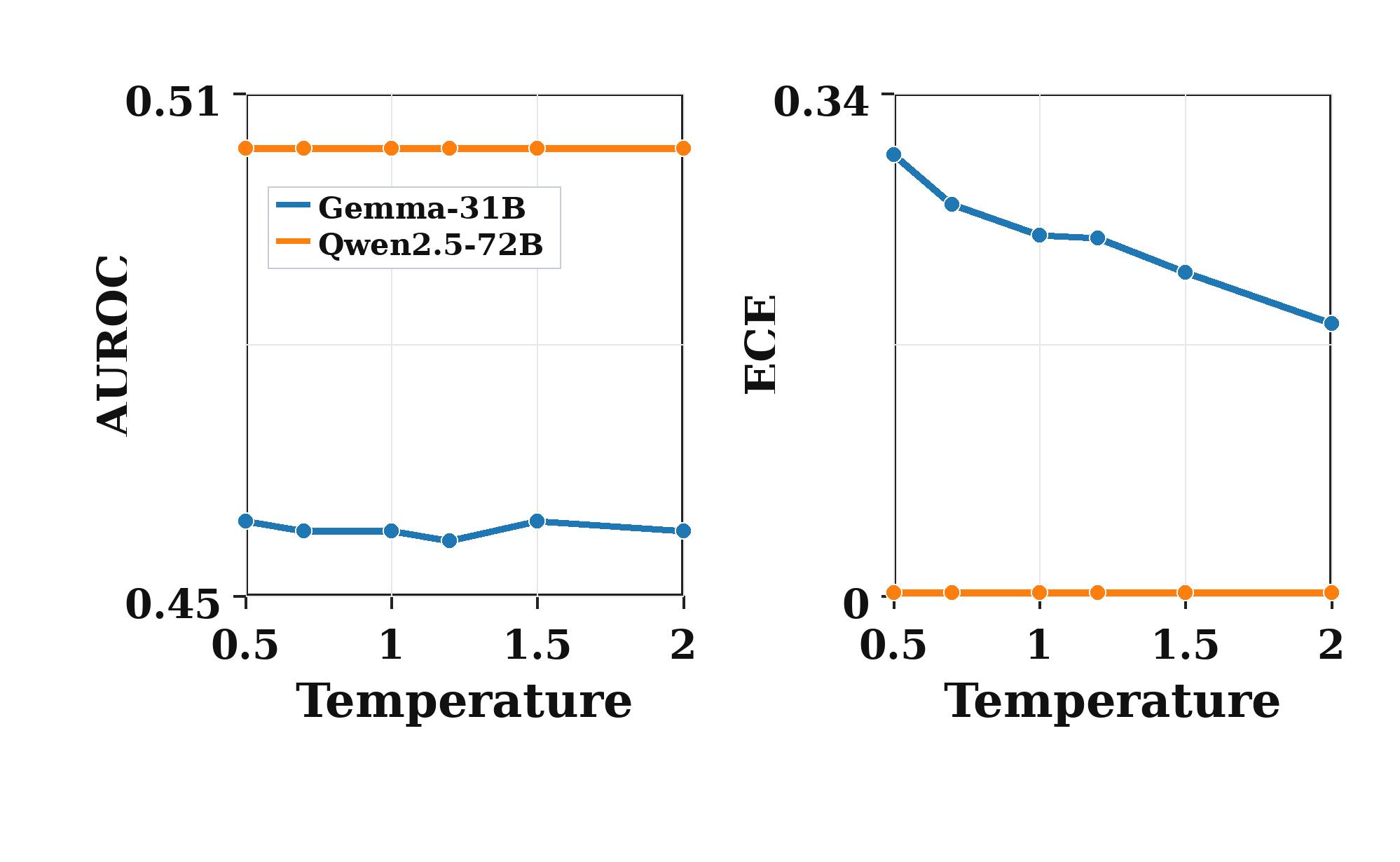}
    \caption{Post-hoc temperature sensitivity of logit-based calibration on AsiaM. AUROC is nearly unchanged because temperature scaling preserves the logit ranking, while Gemma-31B's ECE decreases at higher temperatures, indicating overconfident Yes/No logits. Qwen2.5-72B remains stable, suggesting better calibrated logit confidence.}
    \label{fig:temperature_sensitivity_asiam}
\end{figure}

\begin{figure}[H]
    \centering
    \includegraphics[width=\columnwidth,keepaspectratio]{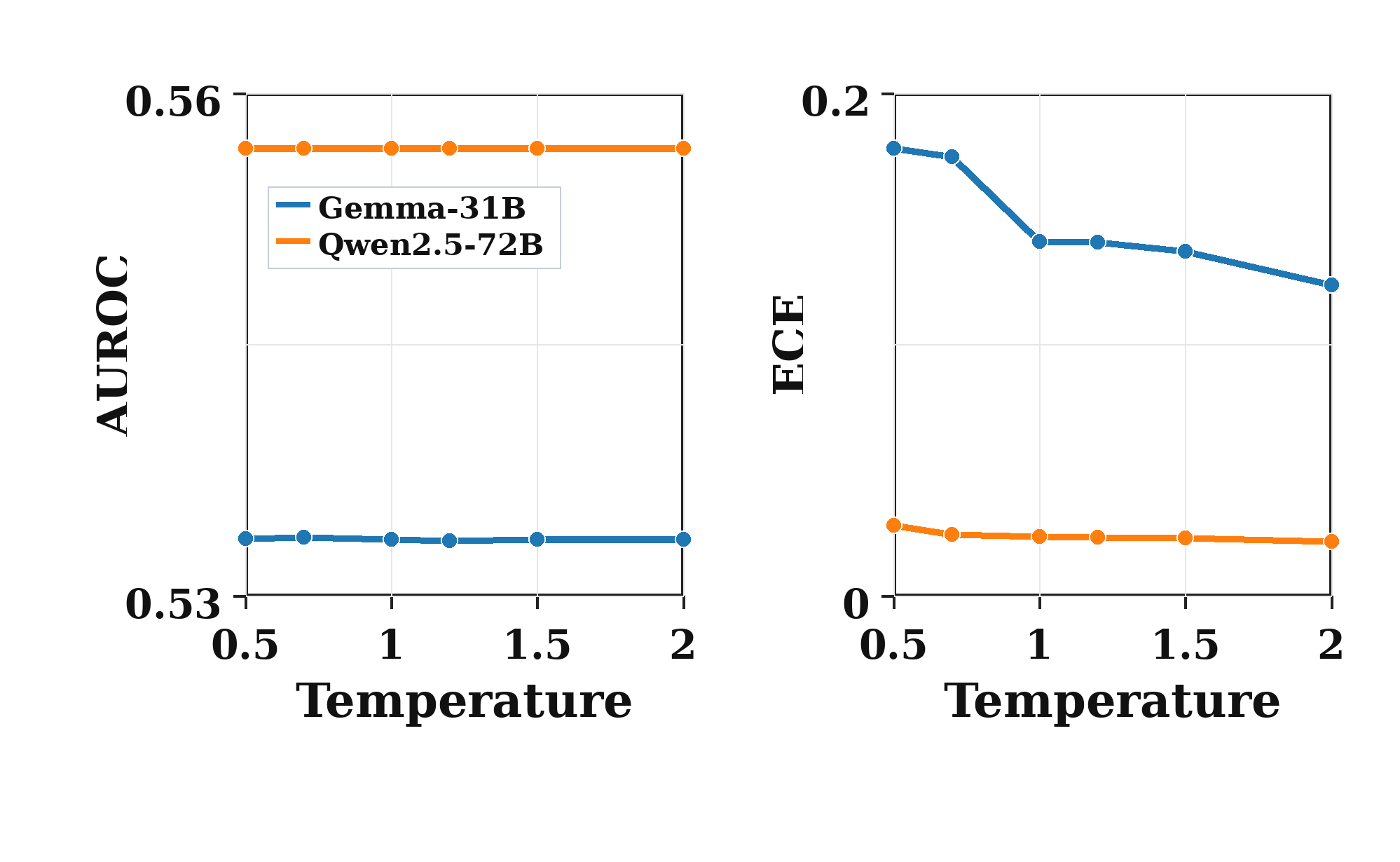}
    \caption{Post-hoc temperature sensitivity of logit-based calibration on River Status. The main effect of temperature appears in ECE rather than AUROC, showing that temperature changes confidence sharpness rather than discrimination. Gemma-31B benefits from softer logits, while Qwen2.5-72B is comparatively insensitive.}
    \label{fig:temperature_sensitivity_river_status}
\end{figure}

\begin{figure}[H]
    \centering
    \includegraphics[width=\columnwidth,keepaspectratio]{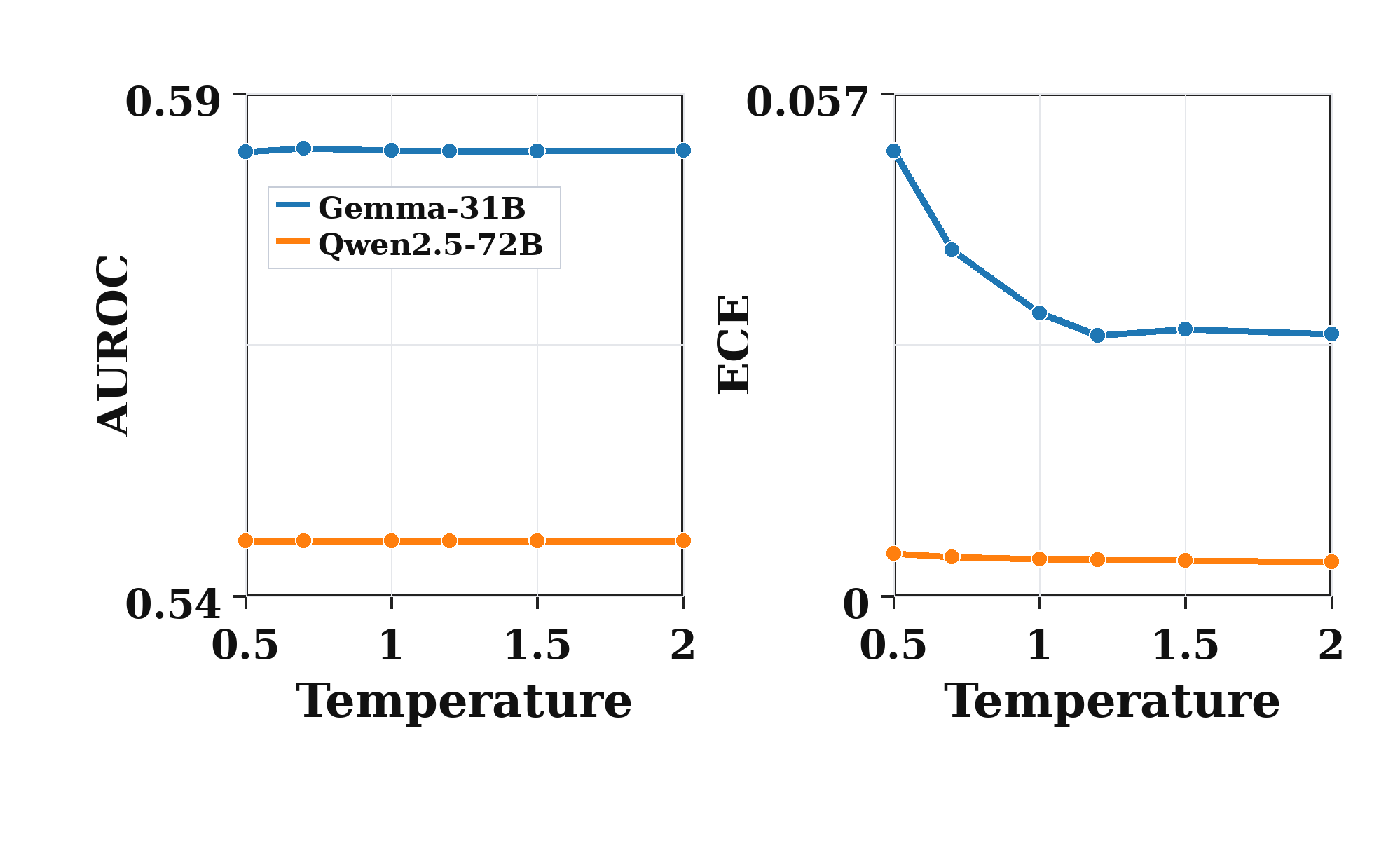}
    \caption{Post-hoc temperature sensitivity of logit-based calibration on COVID. AUROC remains stable across temperatures, confirming that prediction ranking is preserved. ECE varies modestly, with Gemma-31B showing some calibration benefit from temperature adjustment and Qwen2.5-72B remaining consistently well calibrated.}
    \label{fig:temperature_sensitivity_covid}
\end{figure}

\begin{figure}[H]
    \centering
    \includegraphics[width=\columnwidth,keepaspectratio]{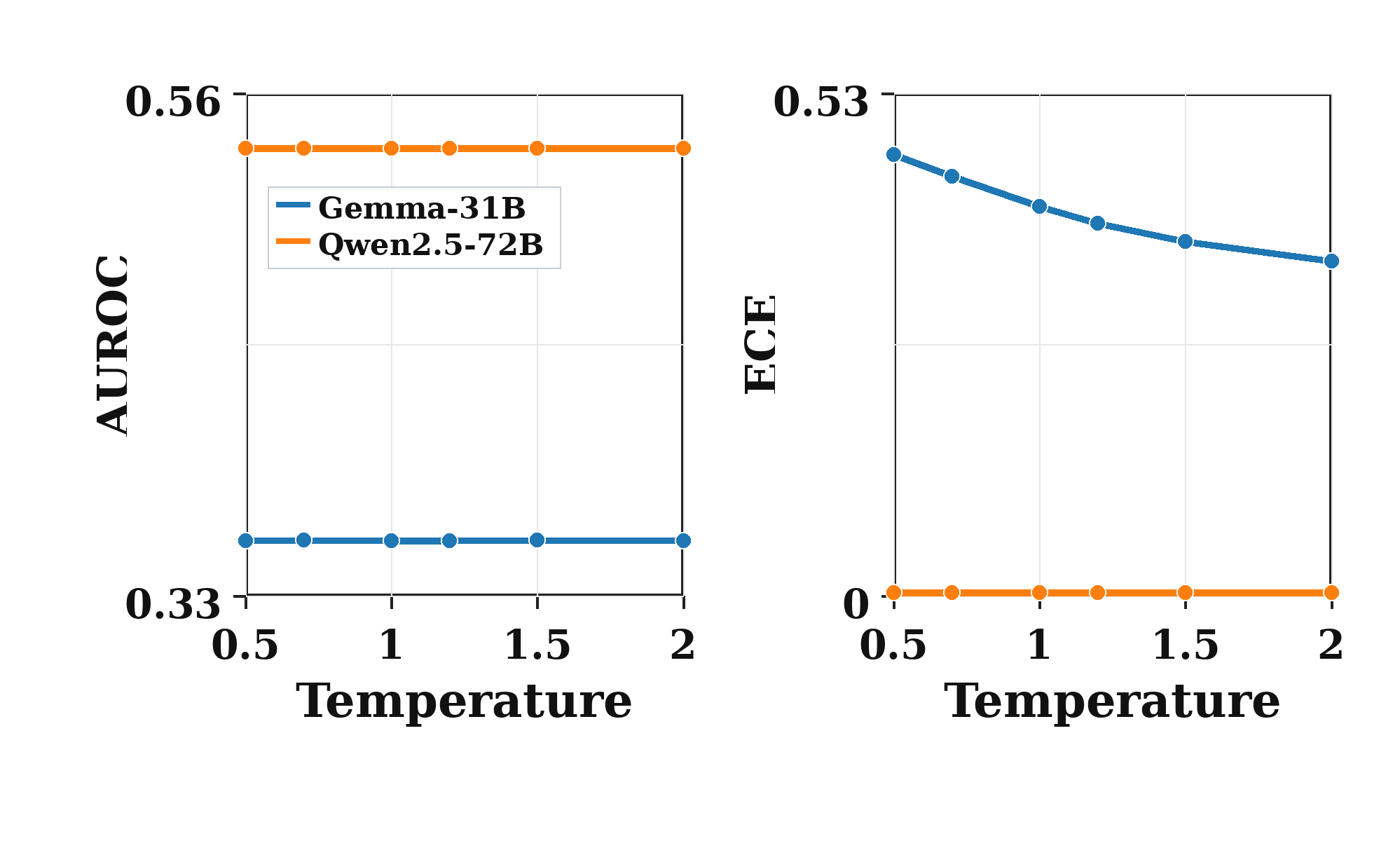}
    \caption{Post-hoc temperature sensitivity of logit-based calibration on Coal Gasifier. Gemma-31B shows a clear ECE reduction as temperature increases, suggesting its saved Yes/No logits are too sharp for reliable confidence estimates. Qwen2.5-72B has low and stable ECE, indicating stronger post-hoc calibration.}
    \label{fig:temperature_sensitivity_coal_gasifier_risk}
\end{figure}

\begin{figure}[H]
    \centering
    \includegraphics[width=\columnwidth,keepaspectratio]{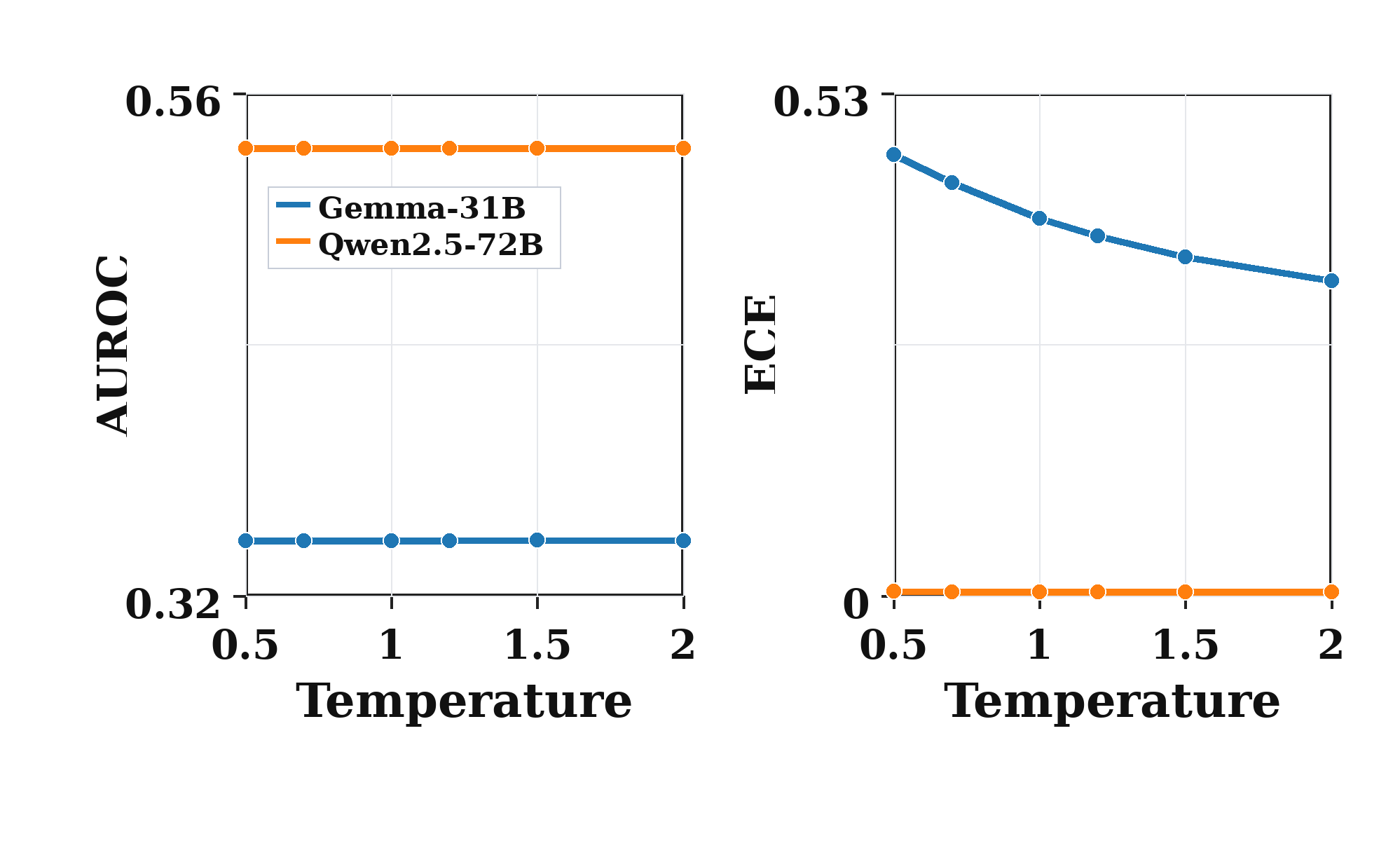}
    \caption{Post-hoc temperature sensitivity of logit-based calibration on Hepar2. Gemma-31B calibration improves substantially under higher-temperature softening, while AUROC remains nearly constant. This supports the project's finding that logit confidence can be recalibrated without changing causal-edge predictions.}
    \label{fig:temperature_sensitivity_hepar2}
\end{figure}

\end{document}